\documentclass[twoside,twocolumn,9pt]{article}
\usepackage{extsizes}
\usepackage[super,sort&compress,comma]{natbib} 
\usepackage[version=3]{mhchem}
\usepackage[left=1.5cm, right=1.5cm, top=1.785cm, bottom=2.0cm]{geometry}
\usepackage{balance}
\usepackage{mathptmx}
\usepackage{sectsty}
\usepackage{graphicx} 
\usepackage{lastpage}
\usepackage[format=plain,justification=justified,singlelinecheck=false,font={stretch=1.125,small,sf},labelfont=bf,labelsep=space]{caption}
\usepackage{float}
\usepackage{fancyhdr}
\usepackage{fnpos}
\usepackage[english]{babel}
\addto{\captionsenglish}{%
  
}
\usepackage{array}
\usepackage{droidsans}
\usepackage{charter}
\usepackage[T1]{fontenc}
\usepackage[usenames,dvipsnames]{xcolor}
\usepackage{setspace}
\usepackage[compact]{titlesec}
\usepackage{hyperref}

\usepackage{acronym}
\usepackage{algorithm}
\usepackage{algpseudocode}
\usepackage{amsmath}
\usepackage{amssymb}
\usepackage{booktabs}
\usepackage{graphicx}
\usepackage{lipsum}
\usepackage{makecell}
\usepackage{multirow}
\usepackage{placeins}
\usepackage{subcaption}
\usepackage{threeparttable}
\usepackage{wrapfig}

\usepackage{epstopdf}

\definecolor{cream}{RGB}{222,217,201}

\begin{document}

\pagestyle{fancy}
\thispagestyle{plain}
\fancypagestyle{plain}{\renewcommand{\headrulewidth}{0pt}}

\makeFNbottom
\makeatletter
\renewcommand\LARGE{\@setfontsize\LARGE{15pt}{17}}
\renewcommand\Large{\@setfontsize\Large{12pt}{14}}
\renewcommand\large{\@setfontsize\large{10pt}{12}}
\renewcommand\footnotesize{\@setfontsize\footnotesize{7pt}{10}}
\makeatother

\renewcommand{\thefootnote}{\fnsymbol{footnote}}
\renewcommand\footnoterule{\vspace*{1pt}%
\color{cream}\hrule width 3.5in height 0.4pt \color{black}\vspace*{5pt}} 
\setcounter{secnumdepth}{5}

\makeatletter 
\renewcommand\@biblabel[1]{#1}            
\renewcommand\@makefntext[1]%
{\noindent\makebox[0pt][r]{\@thefnmark\,}#1}
\makeatother 
\renewcommand{\figurename}{\small{Fig.}~}
\sectionfont{\sffamily\Large}
\subsectionfont{\normalsize}
\subsubsectionfont{\bf}
\setstretch{1.125} 
\setlength{\skip\footins}{0.8cm}
\setlength{\footnotesep}{0.25cm}
\setlength{\jot}{10pt}
\titlespacing*{\section}{0pt}{4pt}{4pt}
\titlespacing*{\subsection}{0pt}{15pt}{1pt}

\fancyfoot{}
\fancyfoot[LO,RE]{\vspace{-7.1pt}\includegraphics[height=9pt]{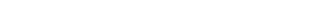}}
\fancyfoot[CO]{\vspace{-7.1pt}\hspace{13.2cm}\includegraphics{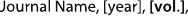}}
\fancyfoot[CE]{\vspace{-7.2pt}\hspace{-14.2cm}\includegraphics{head_foot/RF}}
\fancyfoot[RO]{\footnotesize{\sffamily{1--\pageref{LastPage} ~\textbar  \hspace{2pt}\thepage}}}
\fancyfoot[LE]{\footnotesize{\sffamily{\thepage~\textbar\hspace{3.45cm} 1--\pageref{LastPage}}}}
\fancyhead{}
\renewcommand{\headrulewidth}{0pt} 
\renewcommand{\footrulewidth}{0pt}
\setlength{\arrayrulewidth}{1pt}
\setlength{\columnsep}{6.5mm}
\setlength\bibsep{1pt}

\makeatletter 
\newlength{\figrulesep} 
\setlength{\figrulesep}{0.5\textfloatsep} 

\newcommand{\topfigrule}{\vspace*{-1pt}%
\noindent{\color{cream}\rule[-\figrulesep]{\columnwidth}{1.5pt}} }

\newcommand{\botfigrule}{\vspace*{-2pt}%
\noindent{\color{cream}\rule[\figrulesep]{\columnwidth}{1.5pt}} }

\newcommand{\dblfigrule}{\vspace*{-1pt}%
\noindent{\color{cream}\rule[-\figrulesep]{\textwidth}{1.5pt}} }

\makeatother

\twocolumn[
  \begin{@twocolumnfalse}
{\includegraphics[height=30pt]{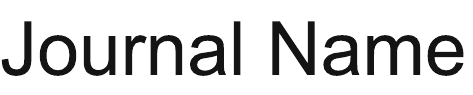}\hfill\raisebox{0pt}[0pt][0pt]{\includegraphics[height=55pt]{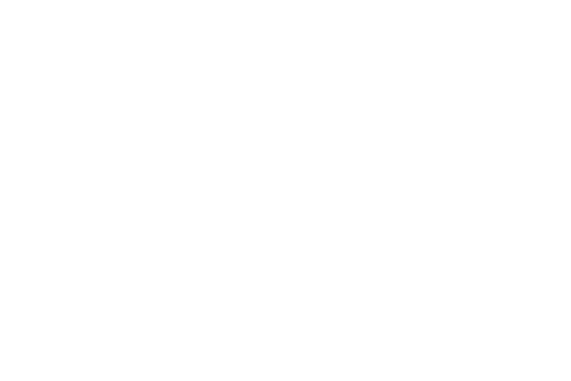}}\\[1ex]
\includegraphics[width=18.5cm]{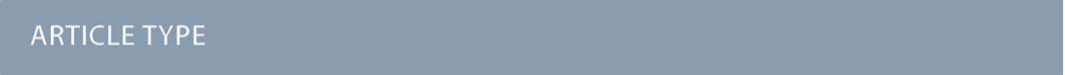}}\par
\vspace{1em}
\sffamily
\begin{tabular}{m{4.5cm} p{13.5cm} }

\includegraphics{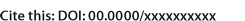} & \noindent\LARGE{\textbf{NEAT-POCKET: Pocket-Conditioned Autoregressive 3D Molecular Generation with a Neighborhood-Guided Set Transformer}} \\
\vspace{0.3cm} & \vspace{0.3cm} \\

& \noindent\large{Roxane Axel Jacob,$^{\ddag\,a\,b\,c}$ Daniel Rose,$^{\ddag\,a\,b\,c}$ Thierry Langer,$^{a\,b}$ and Johannes Kirchmair$^{\ast\,a\,b}$}\\

\includegraphics{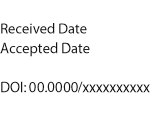} & \noindent\normalsize{AI-driven de novo molecular design offers a promising route to accelerate early-stage drug discovery by generating novel ligands directly within target protein binding pockets. We present NEAT-POCKET, a pocket-conditioned extension of the autoregressive NEAT model for 3D molecular generation. NEAT-POCKET generates molecules atom by atom in protein pocket environments while preserving atom permutation invariance and explicitly modeling hydrogen atoms. Benchmarks on the CrossDocked and SPINDR datasets show that NEAT-POCKET achieves competitive structure-based generation performance while sampling substantially faster than existing baselines. Beyond full-molecule generation, NEAT-POCKET naturally enables pocket-conditioned fragment completion, a task directly relevant to lead optimization and scaffold elaboration. These results position NEAT-POCKET as a fast, flexible, and practical framework for structure-based drug design.}\\

\end{tabular}

 \end{@twocolumnfalse} \vspace{0.6cm}

  ]

\renewcommand*\rmdefault{bch}\normalfont\upshape
\rmfamily
\section*{}
\vspace{-1cm}

\footnotetext{$\ddag$~These authors contributed equally to this work.}
\footnotetext{$\ast$\,Correspondence to: johannes.kirchmair@univie.ac.at}
\footnotetext{$^{a}$\,Department of Pharmaceutical Sciences, Faculty of Life Sciences, University of Vienna, Josef-Holaubek-Platz 2, 1090 Vienna, Austria.}
\footnotetext{$^{b}$\,Christian Doppler Laboratory for Molecular Informatics in the Biosciences, Department of Pharmaceutical Sciences, Faculty of Life Sciences, Josef-Holaubek-Platz 2, 1090 Vienna, Austria.}
\footnotetext{$^{c}$\,Vienna Doctoral School of Pharmaceutical, Nutritional and Sport Sciences (PhaNuSpo), Josef-Holaubek-Platz 2, 1090 Vienna, Austria.}

\acrodef{ai}[AI]{artificial intelligence}
\acrodef{cfg}[CFG]{classifier-free guidance}
\acrodef{gnn}[GNN]{graph neural network}
\acrodefplural{gnn}[GNNs]{graph neural networks}
\acrodef{hba}[HBA]{hydrogen bond acceptor}
\acrodefplural{hba}[HBAs]{hydrogen bond acceptors}
\acrodef{hbd}[HBD]{hydrogen bond donor}
\acrodefplural{hbd}[HBDs]{hydrogen bond donors}
\acrodef{ilp}[ILP]{integer linear programming}
\acrodef{mad}[MAD]{median absolute deviation}
\acrodef{mcmc}[MCMC]{Markov chain Monte Carlo}
\acrodef{ml}[ML]{machine learning}
\acrodef{mlp}[MLP]{multi-layer perceptron}
\acrodef{ode}[ODE]{ordinary differential equation}
\acrodef{qed}[QED]{quantitative estimate of drug-likeness}
\acrodef{rmsd}[RMSD]{root mean square deviation}
\acrodef{ro5}[RO5]{rule of five}
\acrodef{sa}[SA]{synthetic accessibility}
\acrodef{se}[SE]{strain energy}
\acrodef{uff}[UFF]{universal force field}
\acrodef{vdw}[vdW]{van der Waals} 



\section{Introduction}\label{sec:introduction}
The early stages of drug discovery often involve virtual screening of large compound or fragment libraries to identify molecules with potentially strong binding affinity to a desired therapeutic target.\cite{Warr2022VirtualScreening} This process relies on techniques like molecular docking or pharmacophore searches, which can become computationally expensive for ultra-large libraries, and whose outcome is inherently constrained by the size and diversity of the libraries being screened. Generative \ac{ai} offers a promising alternative by enabling the \textit{de novo} design of ligands tailored to a specific target.\cite{Oezcelik2025Generative}

Generative \ac{ml} models can produce molecular structures in various formats, including one-dimensional strings,\cite{GomezBombarelli2018CVAE, Kusner2017GVAE} two-dimensional graphs,\cite{Simonovsky2018GraphVAE, DeCao2022MolGAN} and three-dimensional point clouds.\cite{Gebauer2019GSchnNet, Hoogeboom2022EDM, Luo2021, Luo2022GSphereNet, Xu2023GeoLDM, Daigavane2024Symphony, Irwin2025SemlaFlow, Dunn2025FlowMol, Peng2022Pocket2Mol, Guan2023TargetDiff, Schneuing2024DiffSBDD, Schneuing2025DrugFlow, Cremer2026Flowr} Since many molecular properties and processes, including protein-ligand binding, depend on the 3D conformation of a molecule, generating molecules in three dimensions is particularly valuable.

In this work, we focus on generating 3D small molecules directly within specified protein pockets. The goal is to produce a diverse set of molecules that both fit the pocket geometrically, avoiding steric clashes with pocket atoms, and form favorable intermolecular interactions. Most state-of-the-art approaches formulate this task as transport-based generation, in which a diffusion or flow-matching model learns to transform samples from a simple tractable prior distribution, usually a Gaussian, into samples from the complex distribution of valid 3D molecular structures.\cite{Guan2023TargetDiff, Schneuing2024DiffSBDD, Schneuing2025DrugFlow, Cremer2026Flowr} Diffusion models \cite{Ho2020DDPM, Song2021ScoreMatching} achieve this by progressively corrupting molecular structures with noise during training and then learning to reverse this process at inference time.\cite{Hoogeboom2022EDM, Guan2023TargetDiff, Schneuing2024DiffSBDD} Flow-matching models \cite{Lipman2023FlowMatching, Albergo2025StochasticInterpolants} learn a time-dependent velocity field that transports samples from the tractable source distribution to the target molecular distribution.\cite{Irwin2025SemlaFlow, Dunn2025FlowMol, Schneuing2025DrugFlow, Cremer2026Flowr}

In transport-based approaches to 3D molecule generation, E(3)- or SE(3)-equivariant \acp{gnn} are commonly used as encoder backbones to capture interatomic dependencies and to parameterize the denoising, score, or velocity fields that drive generation.\cite{Guan2023TargetDiff, Schneuing2024DiffSBDD, Schneuing2025DrugFlow, Irwin2025SemlaFlow, Dunn2025FlowMol, Cremer2026Flowr} By respecting the natural symmetries of 3D molecular structures, these networks ensure that the learned fields transform consistently under global translations and rotations.\cite{thomas2018tensorfieldnetworks, satorras2021n} This combination of iterative transport and equivariant message passing has therefore become a standard design choice for 3D molecular generation.

However, this approach is computationally expensive.  The primary bottleneck is the repeated evaluation of the equivariant architecture during sampling: diffusion samplers perform denoising or scoring at each noise level, while flow-matching samplers evaluate the velocity field at each integration step. When these fields are parameterized by an equivariant \ac{gnn} over the ligand--pocket system, each evaluation requires geometric message passing over many atom pairs. Thus, before a single molecule is produced, every atom may be processed by the full equivariant backbone dozens or hundreds of times. This repeated global evaluation is a major source of sampling latency.\cite{Li2024AutoregressiveImageGenerationWithoutVQ}

In our previous work,\cite{rose2026neat} we explored an alternative design for unconditional 3D molecular generation that avoids iterative global transport with an equivariant \ac{gnn} backbone. The NEAT (\underline{N}eighborhood-Guided, \underline{E}fficient, \underline{A}utoregressive Set \underline{T}ransformer) model constructs molecules autoregressively, adding one atom at a time. At each generation step, NEAT encodes the current partial molecule once using a set transformer. A linear classifier then predicts the atom type, and a lightweight flow-matching head predicts its 3D coordinates. Importantly, this coordinate-generation head does not require re-running a global equivariant \ac{gnn} at every denoising or integration step. Consequently, generating an $n$-atom molecule requires only one context encoding per added atom, rather than repeatedly refining the entire molecular structure through many expensive equivariant backbone evaluations. This shift from iterative global transport to autoregressive local construction explains NEAT's substantial speed advantage while maintaining competitive unconditional generation quality. As an additional benefit, the autoregressive formulation supports molecule completion from prefixes, \textit{i.e.}, partial molecules, which is common in lead optimization scenarios. Diffusion and flow-matching models are not suitable for this task without modifications to the training or inference procedures.

In this work, we extend NEAT to condition molecular generation on protein pockets. Our main contributions are as follows:

\begin{itemize}

    \item We propose NEAT-POCKET, an autoregressive model for generating 3D molecules directly within protein-binding pockets. NEAT-POCKET achieves performance competitive with state-of-the-art diffusion- and flow-based approaches, while maintaining a substantial inference-time speed-up.
    
    \item We establish a comprehensive benchmark of existing pocket-conditioned 3D molecular generation methods using a unified evaluation pipeline. This enables consistent comparison across models in terms of molecular validity, target compatibility, similarity to the training data, and sampling efficiency.

    \item We show that NEAT-POCKET can generate molecules from arbitrary molecular prefixes without architectural changes, retraining, or task-specific modifications. This capability naturally supports fragment growing, scaffold decoration, and lead optimization workflows.
\end{itemize}

Overall, NEAT-POCKET provides a fast, flexible, and competitive framework for pocket-conditioned 3D molecular generation, making it a promising tool for structure-based hit generation and lead optimization.

\section{Related works}\label{sec:related_works}
\subsection{Pocket-conditioned 3D molecular generation}

The first protein-conditioned generative model for 3D molecular generation was introduced by \citet{Ragoza2022}, who used a variational autoencoder for pocket-conditioned sampling of atom density grids. \citet{Luo2021} combined atom density prediction with \ac{mcmc} sampling, introducing a novel approach by employing a \ac{gnn} as the encoder backbone. With Pocket2Mol, \citet{Peng2022Pocket2Mol} proposed an E(3)-equivariant network architecture and replaced \ac{mcmc} sampling with a four-step autoregressive approach for 3D molecule generation. At each autoregressive step, their method predicts the frontier atoms of the molecular fragment, samples a focus atom, determines the relative position of the next atom with respect to the focus atom, and predicts the atom type along with the bond types connecting the new atom to existing atoms. The same research group later explored diffusion as an alternative to autoregression, reporting improved performance and faster sampling times with their TargetDiff model.\cite{Guan2023TargetDiff} At the same time, \citet{Schneuing2024DiffSBDD} proposed the DiffSBDD model,\cite{Schneuing2024DiffSBDD} where they combined an SE(3)-equivariant architecture with a diffusion objective. \citet{Schneuing2025DrugFlow} further developed this approach by replacing the diffusion objective with a flow-matching objective, thereby introducing DrugFlow.
\citet{Cremer2026Flowr} highlighted limitations in the training data quality of prior works and introduced a novel dataset derived from protein-ligand cocrystal structures (SPINDR). Alongside this dataset, they proposed the FLOWR model, a pocket-conditioned variant of the SemlaFlow model.\cite{Irwin2025SemlaFlow} Unlike earlier approaches that modeled hydrogen atoms implicitly, FLOWR offers two variants: hydrogen-implicit and hydrogen-explicit. \citet{Cremer2026Flowr} emphasized that explicit hydrogen modeling is a more challenging task and encouraged the research community to focus on this direction in future studies.

\subsection{Fragment-based drug design}

Most pocket-conditioned 3D molecular generators construct ligands in empty binding pockets. While this setting is useful for hit identification, it is less suited to lead optimization, where substructures of an existing ligand are often preserved while other regions are modified to improve pharmacological or pharmacodynamic properties. Autoregressive models naturally accommodate this setting, since conditioning on a fixed prefix aligns with their sequential inference procedure. In contrast, diffusion- and flow-matching-based approaches require additional mechanisms to impose partial molecular constraints.

One approach is to train on the joint distribution of scaffolds and protein pockets.\cite{Xie2024DiffDec, Le2025PilotSBDD}
However, this strategy produces task-specific models whose applicability is constrained by the scaffold or fragment patterns seen in the training data. Moreover, in hydrogen-implicit settings, the model must also infer attachment or anchor points for fragment growth, adding another layer of complexity. Alternatively, inpainting can be applied without retraining, as in DiffSBDD,\cite{Schneuing2024DiffSBDD} by fixing the reverse trajectory of prefix atoms using a precomputed forward trajectory. 
This strategy is more flexible, but increases sampling cost through resampling and, in the absence of explicit hydrogens or learned anchor points, offers limited control over where newly generated atoms attach to the retained fragment.

FLOWR \cite{Cremer2026Flowr} addresses fragment conditioning by including both \textit{de novo} and fragment-conditioned examples during training, effectively incorporating inpainting into the learning procedure. Its hydrogen-explicit variant further enables finer control over attachment sites, making FLOWR the closest existing competitor to NEAT-POCKET for this task.

\section{Model design}\label{sec:model_design}
\begin{figure*}[htbp]
    \centering
    \includegraphics[width=1.\textwidth]{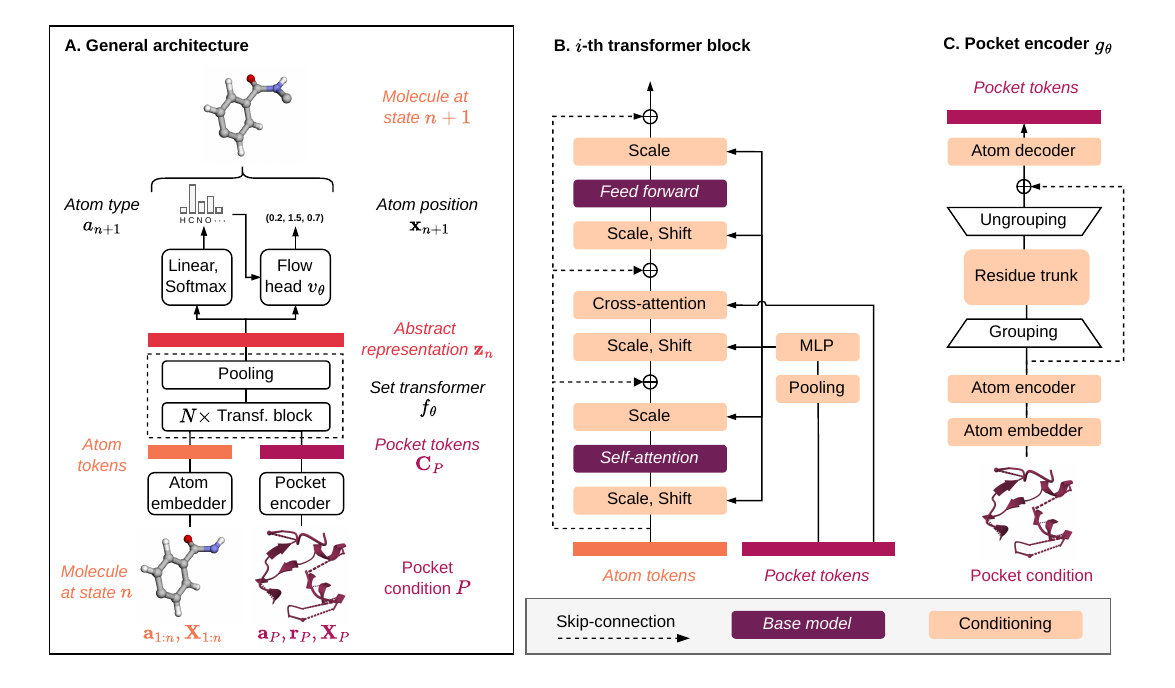}
    \caption{Overview of the architecture of NEAT-POCKET.}
    \label{fig:overview}
\end{figure*}

\subsection{Overview}

Training generative models typically requires substantial amounts of data and computational resources. However, high-quality molecular datasets are limited in availability. Directly training a conditional generative model from scratch is an inefficient use of training data, as the model must first learn fundamental principles of chemistry, such as bond lengths and angles, before it can effectively generate molecules that satisfy specific user-defined constraints. To address this issue, this work adopts a staged training approach, a strategy widely employed in transformer-based text-to-image generation models.\cite{chen2024pixart}
In these models, the training process is divided into two phases. First, the model is trained on large-scale unlabeled image datasets to capture high-level semantics and learn the unconditional data distribution. Subsequently, the model is fine-tuned using conditional data to adapt it to specific tasks. This two-stage training paradigm has proven effective in leveraging limited conditional data while ensuring robust model performance.

In our work, we employ the unconditional NEAT model as the starting point for fine-tuning. NEAT is pre-trained on the GEOM-Drugs dataset,\cite{Axelrod2022GEOM} as described in our previous work \cite{rose2026neat} (details on the datasets used are provided in Section \ref{exp:data}). Building on this foundation, we introduce NEAT-POCKET, a conditional variant of NEAT, which is fine-tuned on the CrossDocked dataset \cite{Francoeur2020CrossDocked} to generate pocket-specific molecular structures \textit{via} the following architectural modifications and training strategies:
\begin{itemize}
    \item The pocket information is encoded with a transformer model. We follow the work of SimpleFold \cite{wang2025simplefold} and adopt a fine-coarse-fine encoding scheme.
    \item Cross-attention blocks are inserted after each self-attention block for fine-grained control on an atom-level. 
    \item Adaptive layer normalization is used to insert coarse-grained information on a global pocket-level.
    \item A \ac{cfg} factor and a clash penalty are used for controlling the strength of the conditioning signal.  
\end{itemize}

\subsection{Data representation}
We use the same molecular representation as in the NEAT base model. Molecules are modeled as undirected graphs $M = (V, E, \mathbf{a}, \mathbf{X})$, with node indices $V = [N]$ representing atoms, edge indices $E \subseteq \{\{i,j\}: 1 \le i < j \le N\}$ representing undirected bonds, $\mathbf{a} \in \mathcal{A}^N$ denoting atom types from a vocabulary $\mathcal{A}$ that includes hydrogen atoms, heavy atoms, a start token, and a stop token, and zero-centered Cartesian coordinates $\mathbf{X} \in \mathbb{R}^{N \times 3}$. 
Protein pockets are modeled as point clouds without edge information. We denote a pocket as $P = (V_p, \mathbf{a}_p, \mathbf{r}_p, \mathbf{X}_p)$, with pocket atom indices $V_p$, pocket atom types $\mathbf{a}_p$, residue types $\mathbf{r}_p$ (\textit{i.e.}, amino acid), and pocket atom positions $\mathbf{X}_p$. 

\subsection{NEAT base model}
NEAT is an autoregressive transformer model for unconditional 3D molecular generation. A key challenge in adapting the standard transformer architecture for this task lies in ensuring permutation invariance of the atom set. This requirement precludes the use of sequence embeddings and causal attention mechanisms, which are key components of standard transformer architectures. To address this limitation, NEAT introduced neighborhood guidance, a novel training paradigm tailored for 3D molecular generation. In short, neighborhood guidance works as follows: during training, connected subgraphs of molecular graphs from the training dataset are dynamically sampled (referred to as the \textit{source set}). The transformer model $f_\theta$ maps the source set $(\mathbf{a}_{1:n}, \mathbf{X}_{1:n})$ to an abstract representation $\mathbf{z}_{n}$ that encapsulates a next-token distribution corresponding to its one-hop neighboring nodes (referred to as the \textit{target set}). This design eliminates any dependence on canonical atom orderings, ensuring permutation invariance. During inference, the learned representation of the source set $\mathbf{z}_{n}$ is used to generate the next atom token, including its type $a_{n+1}$ and 3D position $\mathbf{x}_{n+1}$. The atom type is inferred from the categorical distribution $p(a_{n+1} \mid \mathbf{z}_{n})$ using a linear–softmax head, while the 3D position is sampled from $p(\mathbf{x}_{n+1} \mid a_{n+1}, \mathbf{z}_{n})$ via a conditional flow-matching head implemented as a multi-layer perceptron, modulated with an adaptive layer normalization. 
Further details regarding the training procedure and architecture of the NEAT base model can be found in \citet{rose2026neat}.

\subsection{Pocket encoder model}
The pocket encoder $g_\theta$ maps the all-atom pocket representation $P$ to the atom-level pocket tokens $\mathbf{C}_P$.
We conceptually follow the structure of the SimpleFold encoder \citet{wang2025simplefold}, which employs a \textit{fine-coarse-fine} encoding scheme to balance computational efficiency with resolution. This hierarchical approach is particularly suited for encoding protein pockets, where both local atom-level information and global residue-level context are important. The input to the pocket encoder is at the atom level, with each atom represented by the summation of three components: an atom type embedding, a residue type embedding, and the Fourier encoding of the atom's Cartesian coordinates. To capture local details, a shallow atom-level transformer block is employed, restricting attention to atoms within the same residue. The resulting atom-level encodings are then aggregated via residue-based additive pooling, yielding a residue-level representation. Next, a deeper residue-level transformer trunk with bidirectional self-attention is applied to capture global structural details of the protein pocket. Following this, an ungrouping operation is performed, where residue-level representations are broadcast back to the atom level and residually added to the original atom-level encodings prior to the grouping operation. A second atom-level transformer block, identical in structure to the first, refines the atom-level pocket encodings, completing the hierarchical encoding process. These final atom-level pocket encodings $\mathbf{C}_P$ serve as input to subsequent cross-attention blocks and adaptive layer normalization.

\subsection{Cross-attention}
The architecture of NEAT is inspired by the autoregressive image generation model proposed by \citet{Li2024AutoregressiveImageGenerationWithoutVQ}. In subsequent works, \citet{Li2024AutoregressiveImageGenerationWithoutVQ} extended their model to text-to-image generation, introducing the FLUID framework,\cite{ICLR2025_f8e7248f} which leverages cross-attention to guide the generation process using language prompts. Similarly, we employ cross-attention to integrate pocket-specific information into NEAT's molecule encoder, enabling pocket-conditioned molecular generation. To ensure compatibility with the pretrained NEAT base model, we implemented a zero-initialization strategy for the weights of the final projection layer within the cross-attention block. At initialization, the cross-attention mechanism adds a null vector to the base model's representations, leaving the pretrained model's output unaltered. 
As training progresses, the cross-attention mechanism gradually activates, modulating the learned representations in response to the pocket input. This approach allows us to seamlessly transition from the unconditional NEAT base model to a pocket-conditioned variant, without catastrophic forgetting of chemical knowledge. The pocket-conditioned transformer model, \( f_\theta(\mathbf{a}_{1:n}, \mathbf{X}_{1:n}, \mathbf{C}_P)\), produces the abstract representation \( \mathbf{z}_{n,c} \), analogous to the output of the unconditional base model. This representation can be directly passed to the subsequent layers, allowing all subsequent model components to remain unchanged.

\subsection{Adaptive layer normalization}
Since the introduction of the diffusion transformer (DiT) architecture,\cite{peebles2023dit} adaptive layer normalization (AdaLN) has become a standard mechanism for injecting conditional information into transformer models at a global level. AdaLN modulates the output of layer normalization using learned scale and shift operations derived from the conditional input. In the conventional implementation, these scale and shift vectors are computed at each transformer block by mapping the conditional input through an MLP. However, this approach significantly increases the number of model parameters. To address this, \citet{chen2024pixart} proposed the AdaLN-single layer, which computes the global scale and shift vectors only once and adjusts them at the block level using learned bias vectors. This modification was shown to maintain the performance of standard AdaLN layers while substantially reducing the parameter count. Inspired by this approach, we adopt AdaLN-single layers to inject global pocket information into NEAT's transformer-based encoder. In our implementation, atom-level pocket encodings $\mathbf{C}_P$ are aggregated via additive pooling to produce a pocket-level encoding $\mathbf{c}_P$. This global encoding is then mapped through an MLP to compute the initial scale and shift vectors for AdaLN. To ensure compatibility with the pretrained NEAT base model, we initialize the scale vectors to ones and the shift vectors to zeros, thereby preserving the base model's output at the start of training.

\subsection{Classifier-free guidance}
\ac{cfg} has become a standard training paradigm for conditional generative models.\cite{ho2022CFG}
The core idea involves randomly dropping the conditional input for 10–20\% of the training samples and instead passing a null condition to the model. This approach enables learning both conditional and unconditional generation with a unified model, allowing for the production of samples in either mode during inference. At inference time, the conditional model's learned vector field can be amplified by subtracting the vector field of the unconditional model from it. This operation effectively pushes generated samples away from general regions of the data distribution and draws them closer to regions specific to the conditional input. In our implementation, when conditional information is dropped during training, we pass a learnable unconditional embedding to the cross-attention layer in place of the atom-level pocket embeddings, ensuring the model can still process the input in the absence of conditional information. The same unconditional embedding is also used as input to the AdaLN-single layers, ensuring smooth integration of the \ac{cfg} mechanism into the model architecture.
We denote the output of the transformer without conditioning signal as $ f_\theta(\mathbf{a}_{1:n}, \mathbf{X}_{1:n}, \cdot)=\mathbf{z}_{n,u}$. The next atom position $\mathbf{x}_{n+1}$ is sampled from the distribution $p(\mathbf{x}_{n+1} \mid a_{n+1}, \mathbf{z}_{n})$ through integration of the learned velocity field $\mathbf{v}_\theta (\mathbf{x}_t, t, a_{n+1}, \mathbf{z}_n)$, where $\mathbf{x}_t$ is the next atom's position at time $t$ during integration, and $\mathbf{v}_\theta$ is implemented through NEAT's flow head (details of the integration process can be found in \citet{rose2026neat}). We modulate the learned vector field with \ac{cfg} as follows:
\begin{equation}
    \mathbf{v}_\theta (\mathbf{x}_t, t, a_{n+1}, \mathbf{z}_n) = (1 + \omega) \mathbf{v}_\theta (\mathbf{x}_t, t, a_{n+1}, \mathbf{z}_{n, c}) - \omega \mathbf{v}_\theta (\mathbf{x}_t, t, a_{n+1}, \mathbf{z}_{n, u})
\end{equation}
Analogously, we apply \ac{cfg} to the logits of the atom type classification head.
The \ac{cfg} factor $\omega$ is used to control the degree of adherence to the conditioning signal, offering a trade-off between exploration of the broader data distribution and exploitation of the conditional information.

\subsection{Clash penalty}
To prevent steric clashes with the protein backbone, we introduce a clash penalty term into the training objective. This term penalizes configurations in which ligand atoms overlap with protein backbone atoms, thereby encouraging the generation of physically plausible molecules. During training, the instantaneous velocity of each atom in the target set is computed. Using the interpolated positions of the target set atoms and their instantaneous velocities, we project the target atoms to their final positions. Based on these projected positions, we calculate pairwise distances between the target-set atoms and all atoms in the protein backbone. Additionally, we compute the sum of van der Waals (vdW) radii for each pair of target set atoms and pocket atoms. To quantify clashes, we subtract the pairwise distances from the pairwise sum of vdW radii, clip negative values to zero (indicating no clash), and square the contributions to amplify the penalty for more severe clashes. The resulting clash scores are summed along the pocket dimension to yield atom-level penalty scores. Finally, the overall clash penalty is computed as the mean of the atom-level penalty scores across the batch.

\subsection{Start token}
The inference process of the unconditional NEAT base model begins by randomly sampling an atom type from the atom type distribution of the training data and positioning it at the zero-center. This is possible because the first token is independent of any preceding tokens, eliminating the need for an initial transformer pass. However, in the conditional case, inference starts with the context provided by the protein pocket. To address this, we retrained the unconditional NEAT base model to include a start token that signals the beginning of the inference process. The start token is a learned embedding that is concatenated to the source set tensor. During training, when the model encounters a source set containing only the start token, all atoms of the training molecule are assigned to the target set, since any atom can serve as a valid starting point for autoregressive generation. In the pocket-conditioned setting, the start token plays a critical role in initializing the generation process. It allows the model to pass an empty source set through the pocket-conditioned encoder, producing an abstract representation that captures the positional distribution of the initial atom within the context of the protein pocket.

\subsection{Parameter freezing}
The set transformer is initialized with the pretrained weights of the NEAT base model, whereas the pocket encoder is initialized randomly. To mitigate catastrophic forgetting of the pretrained representation while allowing the pocket encoder to learn its initial features, we adopt a two-stage training strategy following \citet{chen2024pixart}. During the first 150 epochs, the weights of the pretrained base model are frozen, and only the newly added modules are trained to adapt to the pocket-conditional generation task. In this phase, both the monitored clash penalty and the training loss decrease, while molecular validity decreases. After this initial adaptation period, all model parameters are unfrozen and jointly optimized, enabling the full model to adapt to pocket-conditional generation. From this stage onward, we observe a steady increase in molecular validity on the validation set. In contrast, training NEAT-POCKET without the initial parameter-freezing stage led to unstable training behavior, underscoring the importance of our two-stage optimization procedure. 

\subsection{Bond prediction}\label{met:bond_prediction}
The NEAT base model infers bonds from the generated atom types and coordinates using RDKit’s \texttt{xyz2mol} method, which assumes a net molecular charge of zero by default. Although the target net charge can be specified, it must either be known \textit{a priori} or determined by iterating over possible charge states. This exhaustive search does not guarantee reliable convergence or optimal bond assignment, making the approach insufficient for charged molecules present in our training data. To better capture this chemical space, we trained a \textit{post hoc} bond-prediction model on the same dataset used for the generative model. Given atom types and coordinates, the model constructs a radius graph with a 2.5~Å cutoff and classifies each candidate edge as a single, double, triple, or non-bond. During training, we added small Gaussian perturbations to the atomic coordinates to improve robustness to imperfectly generated geometries. Following \citet{Dunn2025FlowMol}, bond prediction is performed on kekulized structures, and aromatic bonds are not modeled explicitly. At inference time, categorical bond probabilities are predicted for the radius graph of each generated molecule and converted into a chemically valid molecular graph using \ac{ilp} with atom-valency constraints. Single-commodity flow constraints are additionally used to enforce graph connectivity. Formal charges are then assigned after bond inference based on atom types and the resulting valencies. Because of its lightweight architecture, the bond predictor introduces negligible computational overhead during sampling while providing a robust alternative to \texttt{xyz2mol}. Further details are provided in the SI, Section \ref{si:implementation}.

\section{Methods}\label{sec:methods}

\subsection{Data acquisition and processing}\label{exp:data}
We used the GEOM-Drugs \cite{Axelrod2022GEOM} dataset for pre-training and the CrossDocked \cite{Francoeur2020CrossDocked} and SPINDR \cite{Cremer2026Flowr} datasets for fine-tuning. 

\paragraph*{GEOM-Drugs.}
GEOM-Drugs is a popular resource for unconditional 3D molecular generation. The dataset originally contains 304,313 drug-like molecules, each with up to 30 conformers. We obtained the data from the repository provided by \citet{raw_geom_data}. 
As in \citet{Vignac2023MiDi}, we retained the five lowest-energy conformers for each molecule, when available. 
The processed data was split into three subsets: 1,172,436 conformers across 243,451 unique molecules for training, 146,394 conformers across 30,430 unique molecules for validation, and 146,788 conformers across 30,432 unique molecules for testing. 

\paragraph*{CrossDocked.}
The CrossDocked dataset is the most widely used benchmark for pocket-conditioned molecular generation. It originally contains 22.5 million docked protein--ligand pairs at varying levels of quality. This dataset was first used by \citet{Ragoza2022} to train a 3D molecule generation model. In our work, we adopted the widely-used split proposed by \citet{Luo2021}, who refined the dataset into a higher-quality subset of 184,057 data points, and subsequently applied sequence alignment to cluster the data at 30\% sequence identity, randomly selecting 100,000 protein--ligand pairs for training and reserving 100 proteins from the remaining clusters for testing. To develop NEAT-POCKET, we randomly selected 5\% of the refined CrossDocked training data as a validation set. Since the structures provided by \citet{Luo2021} do not contain explicit hydrogen atoms, we reconstructed the missing ligand hydrogens with RDKit \cite{rdkit_2025} as part of preprocessing. A detailed description of the hydrogen-addition procedure, together with the associated challenges and limitations, is provided in the SI, Section~\ref{si:hydrogens}. 
After preprocessing, the training, validation, and test sets contain 94,981, 5,000, and 100 protein--ligand pairs, respectively.

\paragraph*{SPINDR.}
Although CrossDocked is valuable for benchmark comparisons, it is constructed using rigid-pocket cross-docking, resulting in an unrealistic distribution of ligand–-pocket interactions and thereby limited utility for drug-discovery applications.
The SPINDR dataset was introduced by \citet{Cremer2026Flowr}, as a refined derivative of PLINDER,\cite{Durairaj2024Plinder} designed to improve structure quality and reduce information leakage. Starting from the June 2024 release of PLINDER, \citet{Cremer2026Flowr} first removed systems containing multiple ligands or multiple protein chains in the binding pocket, as well as ligands annotated as oligos, ions, cofactors, artifacts, fragments, covalent binders, or other nonstandard categories. The remaining complexes were then processed with the Schroedinger Protein Preparation Wizard using the OPLS 2005 force field, which adds missing atoms, standardizes selected residue types, assigns protonation states, adds hydrogens to both proteins and ligands, infers bonds and formal charges, and performs local energy minimization of the protein--ligand complex. Finally, they retained only systems with RDKit-valid ligands, at least five residues in the pocket, and atom types in \{H, C, N, O, F, P, S, Cl, Se, Br\}; they also removed all systems contain N-Acetyl-D-glucosamine ligands because they found these to be highly over-represented. Since selenium atoms are not supported by our pre-trained NEAT model, we further excluded all protein--ligand pairs containing selenium (40 in the training, and 2 in the test set). After this filtering step, the SPINDR training, validation, and test sets contain 35,333, 68, and 223 protein--ligand pairs, respectively.

\subsection{Model training}\label{exp:training}
We initialized NEAT-POCKET from a NEAT model pre-trained on GEOM-Drugs for 1,000 epochs. We then fine-tuned the model separately on CrossDocked and SPINDR for 2,000 and 5,000 epochs, respectively, yielding two NEAT-POCKET models. System specifications, implementation details, and hyperparameters are provided in the SI, Section~\ref{si:implementation}. Our best-performing model has a 12-layer source-set transformer with 12 heads and a hidden dimension of 768, a 1-4-1-layer pocket transformer with 12 heads and a hidden dimension of 768, and an MLP projector with 6 AdaLN blocks and a hidden dimension of 1536 (242M parameters). We used a \ac{cfg} dropout factor of 0.2 and a clash penalty factor of 4.0.

\subsection{Evaluation metrics}\label{exp:metrics}

We evaluated the quality of the generated ligands using several metrics: the percentage of PoseBusters-valid molecules, the number of protein--ligand clashes, the ligand strain energy, docking scores before and after force-field minimization, and similarity of physicochemical descriptors to the training data distribution. We also report the sampling time required to generate 100 molecules for a given protein pocket. All metrics are reported as averages $\pm$ 95\% confidence intervals based on Student’s $t$ distribution, except strain energies, which are reported as median $\pm$ \ac{mad} due to the presence of a few extreme outliers for some baseline models (see SI, Section~\ref{si:strain_energies}). Additional details on the metrics and extended results are provided in the SI, Sections~\ref{si:metrics} and~\ref{si:results}. Finally, we provide examples of molecules generated by NEAT-POCKET and all baseline models in the SI, Section~\ref{si:examples}. 

\paragraph*{Pose-dependent metrics.}
A generated molecule is considered \textbf{PB valid} if it can be successfully parsed by RDKit and passes all 22 conditional PoseBusters checks.\cite{Buttenschoen2023PoseBusters}
The \textbf{number of clashes} between the ligand and the protein pocket were computed using the PoseCheck package.\cite{Harris2023PoseCheck} 
Docking scores were computed using the Gnina binary v1.1.\cite{McNutt2021}
We report the score of the generated pose without further optimization as the \textbf{Vina score}, and the score after local minimization as \textbf{Vina min}. 
Because Vina scores are strongly influenced by ligand size, we also report the average \textbf{molecular weight} of the generated molecules.

\paragraph*{Pose-independent metrics.}
The ligand \textbf{\ac{se}} was computed with the PoseCheck package \cite{Harris2023PoseCheck} and is defined as the difference in internal energy between the generated ligand conformation and its \ac{uff}-optimized conformation.
To assess how closely the generated molecules match the physicochemical properties of the training data, we compared a broad set of descriptors between each model and the CrossDocked dataset. These descriptors include molecular weight, number of heavy atoms, fraction of heteroatoms, fraction of halogen atoms, fraction of rotatable bonds, fraction of chiral centers, fractions of hydrogen bond acceptors and donors, computed logP, \ac{qed}, total number of rings, fractions of aromatic and aliphatic rings, fraction of bridgehead atoms, fraction of spiro atoms, fractions of $N$-membered rings with $N=\{3,4,5,6,7,8,>8\}$, and the distribution of atom types H, C, N, P, O, S, F, Cl, and Br. Additional details on how each descriptor was calculated are provided in the SI, Section~\ref{si:metrics}. Relative descriptor distributions with respect to CrossDocked are shown in Figures~\ref{fig:physchem_relative} and~\ref{fig:rings_and_atom_types_relative} and discussed in Section~\ref{exp:results-pocket-crossdocked}, while the corresponding absolute distributions are reported in the SI, Figures~\ref{fig:physchem_absolute} and~\ref{fig:rings_and_atom_types_absolute}. For conciseness, these descriptor-level comparisons are summarized in Table~\ref{tab:results_metrics_1} using a single aggregate metric, the \textbf{physicochemical rank}. This score was computed by ranking each model according to its similarity to CrossDocked for each descriptor, where lower ranks indicate closer agreement with the training distribution, and then averaging the ranks across all descriptors.

\paragraph*{Sampling time.}
Computational efficiency was measured as the average time required to generate 100 molecules for a protein pocket on a Rocky Linux 9.7 system equipped with an NVIDIA GeForce RTX 4090 GPU with 24 GB of GDDR6X VRAM and an AMD Ryzen 9 7950X 16-Core Processor.

\subsection{Baselines}\label{exp:baselines}
We compare NEAT-POCKET trained on CrossDocked against Pocket2Mol,\cite{Peng2022Pocket2Mol} TargetDiff,\cite{Guan2023TargetDiff} DiffSBDD,\cite{Schneuing2024DiffSBDD} and DrugFlow.\cite{Schneuing2025DrugFlow}
Because these models do not generate explicit hydrogen atoms, we added hydrogens to all generated molecules prior to evaluation using a consistent preprocessing protocol. This improves comparability across methods and avoids metric-dependent inconsistencies, since some commonly used evaluation tools add missing hydrogen atoms internally using their own procedures. We compare NEAT-POCKET trained on SPINDR with FLOWR.\cite{Cremer2026Flowr} FLOWR is available in two variants: one trained on molecules with explicit hydrogen atoms and one trained on molecules without explicit hydrogen atoms. Here, we use the explicit-hydrogen version of FLOWR. We did not retrain any of the baseline models and generated compounds using the authors' released code and pretrained model weights. Detailed descriptions of how each baseline was installed and run are provided in the SI, Section~\ref{si:baselines}.

\section{Results}\label{sec:results}
\begin{table*}[ht]
    \caption{Performance of NEAT-POCKET vs. other pocket-conditioned 3D molecule generators trained on CrossDocked or SPINDR (best in bold, second-best underlined; sampling times reported for the generation of 100 molecules)}
    \resizebox{\linewidth}{!}{
        \begin{tabular}{lcccccccc}
                        & \makecell[c]{PB valid}
                        & \makecell[c]{PC clashes}
                        & \makecell[c]{PC SE}
                        & \makecell[c]{Vina score}
                        & \makecell[c]{Vina min}
                        & \makecell[c]{Mol. weight}
                        & \makecell[c]{Rank}
                        & \makecell[c]{Sampling time} \\
                        & \makecell[c]{[\%] $\uparrow$}
                        & \makecell[c]{[1] $\downarrow$}
                        & \makecell[c]{[kcal/mol] $\downarrow$}
                        & \makecell[c]{[kcal/mol] $\downarrow$}
                        & \makecell[c]{[kcal/mol] $\downarrow$}
                        & \makecell[c]{[g/mol]}
                        & \makecell[c]{[1] $\downarrow$}
                        & \makecell[c]{[s] $\downarrow$} \\
            \midrule
            \midrule
            CrossDocked & 94.0 $\pm$ 4.7
                        & ~~6.9 $\pm$ 1.1
                        & ~~~~63 $\pm$ ~~23
                        & -7.1 $\pm$ 0.6
                        & -7.3 $\pm$ 0.5
                        & 332 $\pm$ 29
                        & -
                        & - \\
            \midrule
            Pocket2Mol  & \textbf{71.1 $\pm$ 3.7}
                        & ~~\underline{6.9 $\pm$ 0.9}
                        & ~~~~\textbf{65 $\pm$ ~~36}
                        & -5.4 $\pm$ 0.5
                        & \underline{-6.9 $\pm$ 0.5}
                        & 223 $\pm$ 21
                        & 3.00
                        & 845 $\pm$ 838 \\
            TargetDiff  & 46.4 $\pm$ 3.3
                        & 10.8 $\pm$ 1.4
                        & 1003 $\pm$ 378
                        & \textbf{-6.5 $\pm$ 0.4}
                        & \textbf{-7.5 $\pm$ 0.4}
                        & 324 $\pm$ 20
                        & 3.39
                        & 722 $\pm$ 201 \\
            DiffSBDD    & 49.6 $\pm$ 2.6
                        & 14.3 $\pm$ 2.0
                        & ~~891 $\pm$ 271
                        & -3.6 $\pm$ 0.8
                        & -6.3 $\pm$ 0.4
                        & 296 $\pm$ 14
                        & 3.32
                        & ~~\underline{88 $\pm$ ~~30} \\
            DrugFlow    & 61.7 $\pm$ 3.7
                        & ~~9.2 $\pm$ 1.4
                        & ~~\underline{255 $\pm$ 113}
                        & \underline{-5.8 $\pm$ 0.5}
                        & \underline{-6.9 $\pm$ 0.4}
                        & 309 $\pm$ 20
                        & \underline{2.81}
                        & 112 $\pm$ ~~35 \\
            NEAT-POCKET     & \underline{69.2 $\pm$ 3.6}
                        & ~~\textbf{6.6 $\pm$ 0.9}
                        & ~~265 $\pm$ 167 
                        & -5.1 $\pm$ 0.3
                        & -6.6 $\pm$ 0.4
                        & 289 $\pm$ 22
                        & \textbf{2.48}
                        & ~~~~\textbf{4 $\pm$ ~~~~1} \\
            \midrule
            \midrule
            SPINDR      & 99.0 $\pm$ 1.0
                        & ~~4.3 $\pm$ 0.5
                        & ~~~~16 $\pm$ ~~~~3
                        & -8.1 $\pm$ 0.3
                        & -8.5 $\pm$ 0.3
                        & 391 $\pm$ 16
                        & -
                        & - \\
            \midrule
            FLOWR       & 75.8 $\pm$ 1.0 
                        & ~~4.6 $\pm$ 0.4
                        & ~~~~\textbf{35 $\pm$ ~~~~9}
                        & \textbf{-7.0 $\pm$ 0.2}
                        & \textbf{-7.7 $\pm$ 0.2} 
                        & 395 $\pm$ ~~6 
                        & 1.56
                        & ~~36 $\pm$ ~~17 \\
            NEAT-POCKET     & \textbf{80.3 $\pm$ 1.6}
                        & ~~\textbf{3.7 $\pm$ 0.3}
                        & ~~~~62 $\pm$ ~~34 
                        & -5.5 $\pm$ 0.2 
                        & -6.8 $\pm$ 0.2 
                        & 307 $\pm$ 12
                        & \textbf{1.44}
                        & ~~~~\textbf{4 $\pm$ ~~~~1} \\
        \end{tabular}
    }
    \label{tab:results_metrics_1}
\end{table*}

\begin{figure*}
    \centering
    \includegraphics[width=1.0\linewidth]{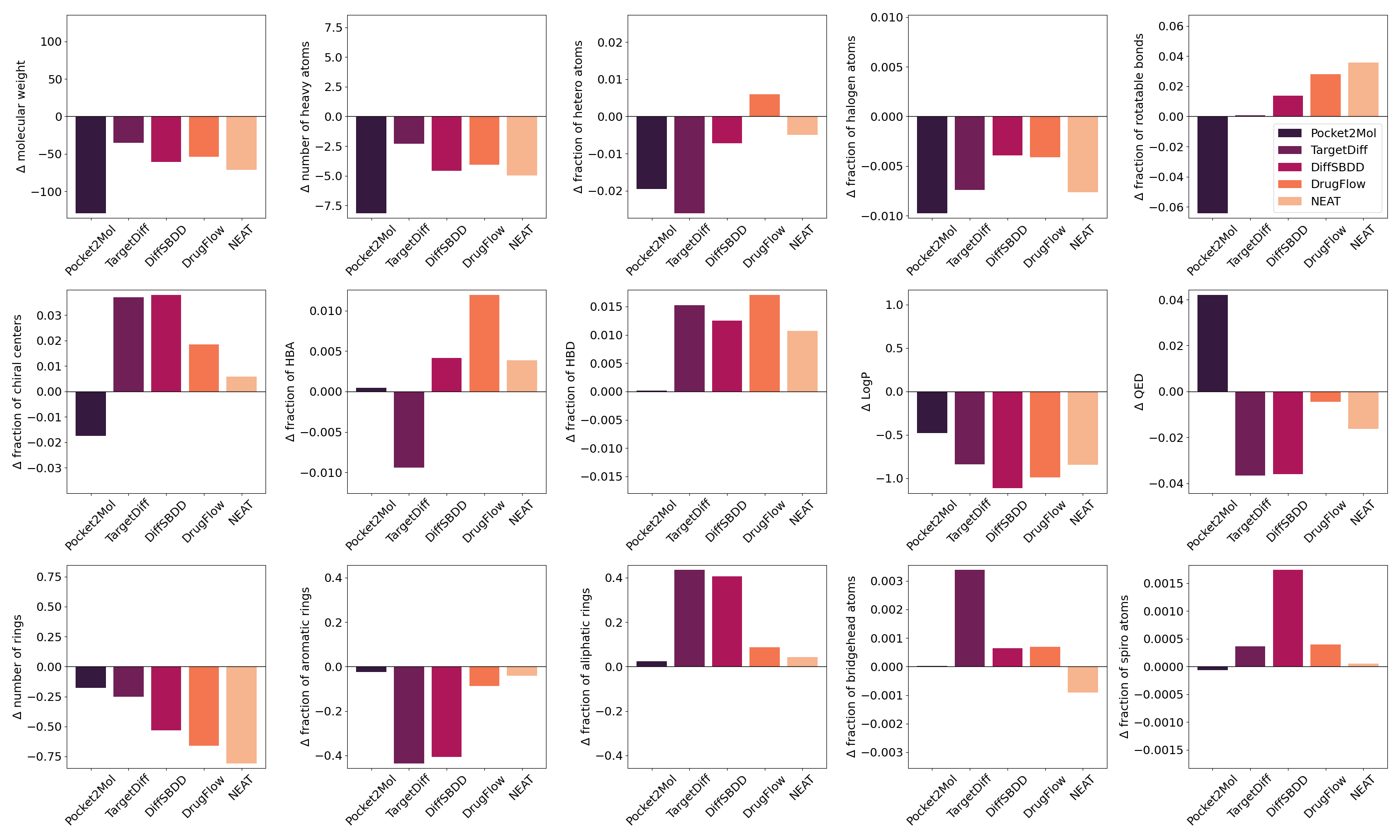}
    \caption{Average values of general physicochemical properties of molecules generated by Pocket2Mol, TargetDiff, DiffSBDD, DrugFlow and NEAT-POCKET trained on CrossDocked --- deviation from CrossDocked.}
    \label{fig:physchem_relative}
\end{figure*}

\begin{figure*}
    \centering
    \includegraphics[width=1.0\linewidth]{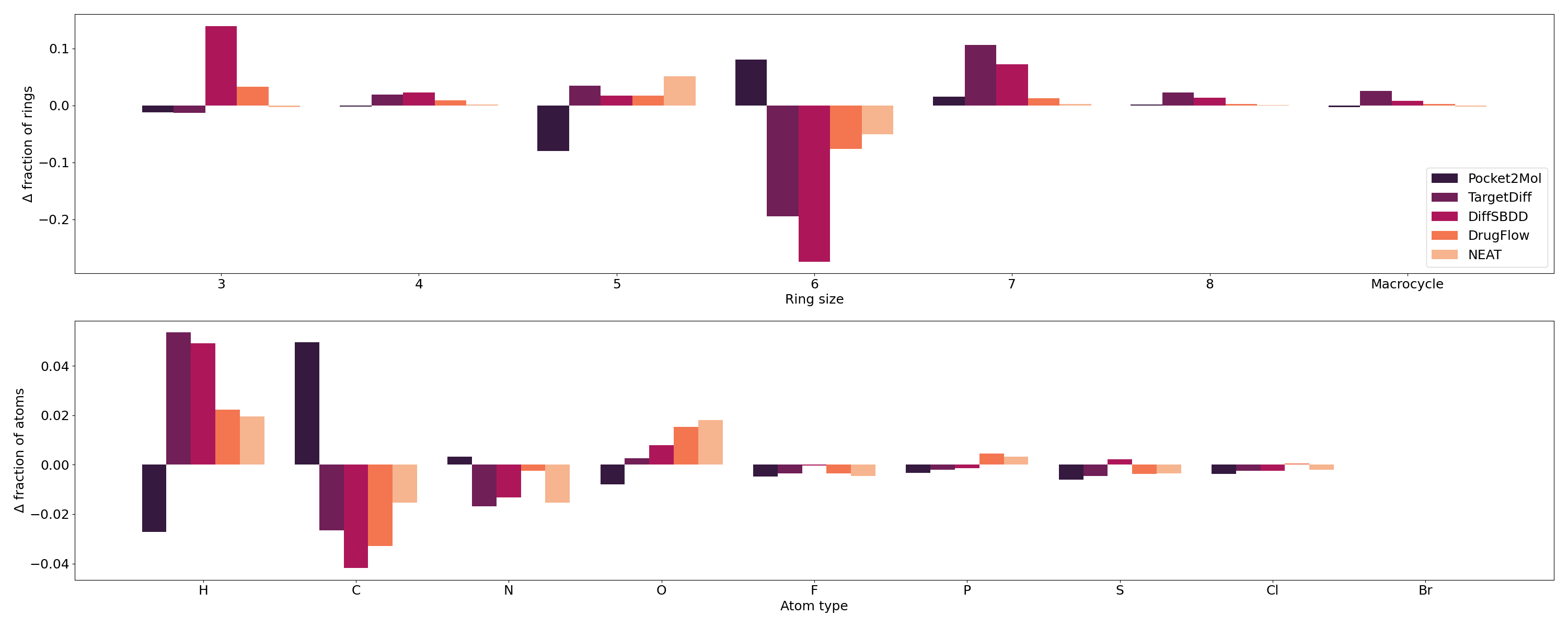}
    \caption{Distribution of ring sizes (top) and atom types (bottom) of molecules generated by Pocket2Mol, TargetDiff, DiffSBDD, DrugFlow and NEAT-POCKET trained on CrossDocked --- deviation from CrossDocked.}
    \label{fig:rings_and_atom_types_relative}
\end{figure*}

We benchmark NEAT-POCKET against established methods on the widely used CrossDocked benchmark, demonstrating competitive performance together with improved inference efficiency. Although CrossDocked is valuable for standardized comparisons, it is constructed via rigid-pocket cross-docking, which leads to an unrealistic distribution of ligand–-pocket interactions and limits its utility for drug-discovery applications. We therefore additionally evaluate NEAT-POCKET on the newly introduced SPINDR dataset, derived from crystal structures in the PLINDER resource, providing a setting more representative of practical use. Next, we analyze the \ac{cfg} scale as the key inference-time parameter controlling the trade-off between molecular validity and adherence to the conditioning context. Finally, we demonstrate NEAT's unique advantage over baseline models for prefix completion.

\subsection{Performance on CrossDocked}\label{exp:results-pocket-crossdocked}

Each of the 100 pockets in the CrossDocked test set is associated with a docked reference ligand. 
We first evaluated the quality of these reference ligands using the same metrics as for the generated molecules; the resulting values are reported in Table~\ref{tab:results_metrics_1} and serve as reference values. 
Overall, 94.0\% of the CrossDocked test-set ligands are PB valid. 
The failures are due to internal steric clashes, non-planarity of non-aromatic rings, and violations of the minimum-distance criterion to the protein. 
The reference ligands have an average of 6.9 protein--ligand clashes and an average strain energy of 63~kcal/mol. 
Their average Vina score is $-7.1$~kcal/mol, improving slightly to $-7.3$~kcal/mol after force-field minimization, and their average molecular weight is 332~Da.

\paragraph*{NEAT-POCKET generates valid, compact, low-clash ligands.}
Among the generative models, NEAT-POCKET achieved the second-highest PB validity, with 69.2\% PB valid molecules, and the best (lowest) number of protein--ligand clashes, with an average of 6.6 clashes. 
In terms of strain energy, NEAT-POCKET performed third-best among the generated models, with an average PC SE of 265~kcal/mol.
NEAT-POCKET shows a preference for compact ligands that fit the pocket while avoiding steric clashes with protein atoms. 
This behavior is consistent with its training objective, which explicitly penalizes ligand–-pocket clashes and thus discourages overfilling the pocket. 
As a result, NEAT-POCKET generates relatively small molecules, with an average molecular weight of 289~Da, compared with 324~Da for TargetDiff and 309~Da for DrugFlow. 
This is important when interpreting Vina scores, which are highly correlated with ligand size. 
NEAT-POCKET obtains an average Vina score of -5.1~kcal/mol and a minimized Vina score of -6.6~kcal/mol, placing it behind Pocket2Mol, DrugFlow, and TargetDiff on these metrics. 
However, its molecular weight range aligns well with early-stage drug discovery, where compounds in the 250–300~Da range are often desirable starting points because they leave room for subsequent optimization of pharmacokinetic or pharmacodynamic properties.

\paragraph*{NEAT-POCKET most closely reproduces the physicochemical properties of CrossDocked molecules.}
NEAT-POCKET achieves the lowest descriptor rank (2.48), indicating that its generated molecules exhibit the greatest average physicochemical similarity to the CrossDocked ligands.
Figures~\ref{fig:physchem_relative} and~\ref{fig:rings_and_atom_types_relative} show the distributions of individual physicochemical descriptors relative to the CrossDocked dataset.
NEAT-POCKET ranks first among all evaluated baselines in terms of the fractions of heteroatoms, chiral centers, spiro atoms, and hydrogen and carbon atoms.
Although NEAT-POCKET molecules contain fewer rings than those generated by most other models, it more faithfully reproduces the training-data distribution of ring sizes. 
In contrast, TargetDiff and DiffSBDD generate disproportionately many aliphatic rings relative to aromatic rings. TargetDiff also overproduces bridgehead atoms, whereas DiffSBDD overproduces spiro atoms. 
Both TargetDiff and DiffSBDD underproduce six-membered rings and overproduce seven-membered rings, and DiffSBDD additionally generates an excess of three-membered rings.
Moreover, TargetDiff and DiffSBDD oversample hydrogen atoms and undersample carbon atoms compared with the training data. 
Together with their high strain energies, these trends indicate that molecules generated by TargetDiff and DiffSBDD are often highly constrained and less consistent with typical drug-like structures. 
Pocket2Mol shows the opposite tendency, producing too many carbon atoms, too few hydrogen atoms, and too few rotatable bonds compared to the CrossDocked reference data. 
Combined with its low average molecular weight, this suggests that Pocket2Mol predominantly generates small, rigid, and highly saturated molecules.
Taken together, these analyses indicate that NEAT-POCKET generates molecules that are, on average, more consistent with the CrossDocked data distribution than those of the other baselines, while avoiding several structural artifacts observed in TargetDiff, DiffSBDD, and Pocket2Mol. 

\paragraph*{NEAT-POCKET is the fastest among evaluated models.}
NEAT-POCKET is substantially faster than all evaluated baselines, requiring only 4 seconds on average to generate 100 molecules for a pocket. 
This corresponds to an approximately 20-fold speed-up over the second-fastest model, DiffSBDD, which requires 88 seconds on average, and more than an order-of-magnitude improvement over the remaining baselines.

\subsection{Performance on SPINDR}

\begin{figure*}[ht]
    \centering
    \includegraphics[width=1.0\linewidth]{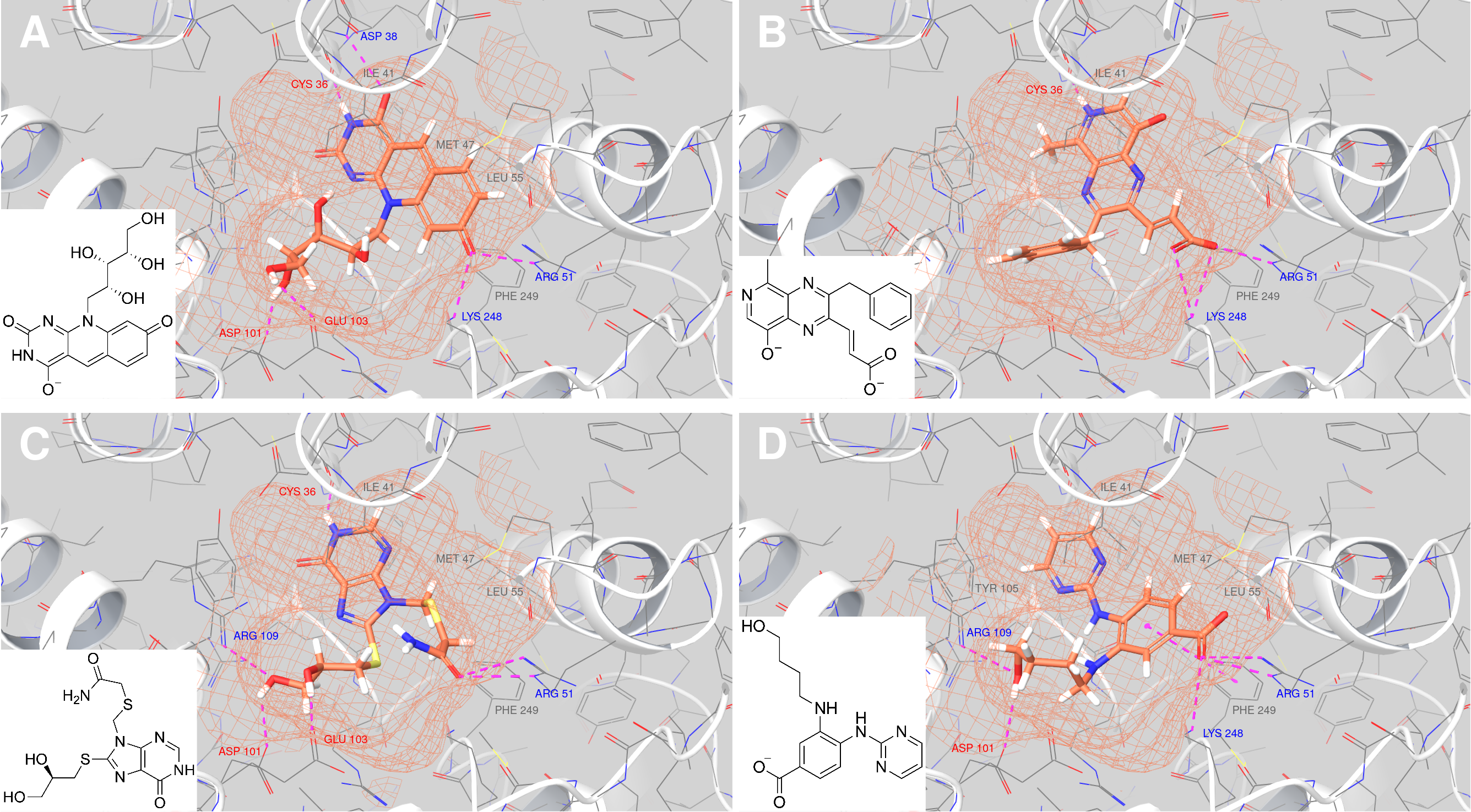}
    \caption{Representative examples of pocket-conditioned 3D molecule generation with NEAT-POCKET. Molecules were generated in the binding pocket of DNA photolyase from the cyanobacterium \textit{Anacystis nidulans} (PDB: 1QNF). (A) Experimentally resolved reference ligand (7,8-didemethyl-8-hydroxy-5-deazariboflavin). (B--D) Representative generated ligands.}
    \label{fig:1QNF}
\end{figure*}

Like for CrossDocked, we first evaluated the quality of the 223 reference ligands from the SPINDR test set using the same metrics as for the generated molecules; the resulting values are reported in Table~\ref{tab:results_metrics_1}. The SPINDR ligands are of higher quality than the CrossDocked ligands: 99.0\% are PB valid, with only 4.3 protein--ligand clashes on average and a low average strain energy of 16.1~kcal/mol. They also achieve substantially more favorable docking scores than the CrossDocked reference ligands, with an average Vina score of $-8.1$~kcal/mol and a minimized Vina score of $-8.5$~kcal/mol. Their average molecular weight is 391~Da.

\paragraph*{NEAT-POCKET prioritizes validity and clash-avoidance, while FLOWR favors larger, lower-energy ligands.}
Compared with FLOWR, which is currently the only other 3D molecular generation model trained on this dataset, NEAT-POCKET achieves higher PB validity, 80.3\% versus 75.8\%, and fewer protein--ligand clashes, 3.7 versus 4.6 on average. Notably, NEAT-POCKET also produces fewer clashes than the SPINDR reference ligands themselves, suggesting that the clash-penalized training objective effectively encourages poses that are geometrically compatible with the protein pocket.

In terms of strain energy and docking scores, FLOWR performed better than NEAT-POCKET. FLOWR's generated molecules had a lower average \ac{se}, 34~kcal/mol, compared with 62~kcal/mol for NEAT-POCKET, and more favorable Vina scores, with an average raw Vina score of $-7.0$~kcal/mol and a minimized score of $-7.7$~kcal/mol, compared with $-5.5$ and $-6.8$~kcal/mol for NEAT-POCKET. However, these differences should be interpreted in light of molecular size. FLOWR generated molecules with an average molecular weight of 395~Da, closely matching the SPINDR reference ligands, whereas NEAT-POCKET generated smaller molecules, with an average molecular weight of 307~Da. Since Vina scores generally become more favorable for larger ligands, the higher molecular weight of FLOWR molecules likely contributes to their stronger docking scores. By contrast, the smaller size of NEAT-POCKET molecules is consistent with its clash-aware objective and may be advantageous in early-stage lead generation, where lower molecular weight leaves room for subsequent optimization.

In terms of physicochemical similarity to the SPINDR dataset, NEAT-POCKET ranked higher than FLOWR, although by a small margin. This indicates that both models capture important aspects of the SPINDR molecular distribution, with NEAT-POCKET performing slightly better on average across the descriptor set.

\paragraph*{NEAT-POCKET is nine times faster than FLOWR.}
Finally, NEAT-POCKET was markedly more efficient than FLOWR. It required only 4 seconds on average to generate 100 molecules for a pocket, compared with 36 seconds for FLOWR, corresponding to a nine-fold speed-up. Taken together, the SPINDR results show that NEAT-POCKET compares favorably to FLOWR: while FLOWR generated molecules with lower Vina scores and strain energies, NEAT-POCKET generated more valid molecules with fewer protein--ligand clashes, achieved better overall physicochemical similarity to the dataset, and showed considerably higher speed.

\paragraph*{NEAT-POCKET generates diverse ligands with plausible pocket interactions.}
Figure~\ref{fig:1QNF} shows representative examples of pocket-conditioned generation with NEAT-POCKET for DNA photolyase from the cyanobacterium \textit{Anacystis nidulans} (PDB: 1QNF). The experimentally resolved ligand (7,8-didemethyl-8-hydroxy-5-deazariboflavin, panel~A) forms an extended interaction network in the crystal structure that includes (i) hydrogen bonds between the ribityl hydroxyl groups and Asp101 and Glu103, (ii) hydrogen bonds between the aromatic ligand core and Cys36, Asp38, Arg51, and Lys248, and (iii) hydrophobic interactions between the aromatic core and Met47, Ile41, Leu55, and Phe249.


The generated molecules in panels~B--D are well placed within the binding pocket and recover many of the interactions observed for the crystallographic ligand, while sampling distinct new chemotypes. In molecule~B, an acidic side chain forms hydrogen bonds and ionic interactions with Arg51 and Lys248, while the heteroaromatic core remains centered in the pocket, forming a hydrogen bond with Cys36 and hydrophobic contacts with Ile41 and Phe249. Molecule~C contains two polar substituents that recover the reference ligand's interactions with Asp101 and Glu103, and also forms an additional hydrogen bond with Arg51. Its heteroaromatic scaffold occupies the central pocket region and forms a hydrogen bond with Cys36, and hydrophobic interactions with Met47, Ile41, Leu55, and Phe249. Molecule~D recovers the hydrogen bonds with Arg51, Lys248, and Asp101, and establishes an additional polar contact with Arg109. It also forms a $\pi$--$\pi$ stacking interaction with Phe249, and hydrophobic interactions involving Met47, Ile41, Leu55, and Tyr105. Collectively, these examples show how NEAT-POCKET can generate chemically diverse ligands that occupy the correct pocket region while preserving plausible interaction motifs with key binding-site residues.

\subsection{Impact of classifier-free guidance}\label{exp:cfg}

Tables~\ref{tab:results_cfg_ablation_crossdocked} and~\ref{tab:results_cfg_ablation_spindr} report the effect of the \ac{cfg} scaling factor $\omega$ on molecules generated by NEAT-POCKET trained on CrossDocked and SPINDR, respectively. The ablation shows a consistent trade-off between pocket-specific optimization and molecular plausibility.

In the unguided setting, $\omega=0.0$, NEAT-POCKET achieves the highest PB validity on both datasets, with 71.0\% valid molecules on CrossDocked and 80.5\% on SPINDR. This setting also yields the lowest strain energies, 155~kcal/mol and 32~kcal/mol, respectively. However, the generated molecules are less well adapted to the target pocket: they exhibit the highest number of protein--ligand clashes and the lowest raw Vina scores among all guidance settings.

Increasing $\omega$ improves the pocket-level metrics. The number of clashes decreases monotonically from 8.3 to 4.5 on CrossDocked and from 4.8 to 2.3 on SPINDR. Similarly, the raw Vina score becomes more favorable, improving from $-4.4$ to $-5.6$~kcal/mol on CrossDocked and from $-4.8$ to $-6.0$~kcal/mol on SPINDR. This indicates that stronger \ac{cfg} increases the influence of the protein--pocket conditioning signal during sampling.

\begin{table}[ht]
    \caption{Impact of \ac{cfg} on the quality of molecules generated by NEAT-POCKET trained on the CrossDocked dataset (best in bold, second-best underlined)}
    \setlength{\tabcolsep}{2pt}
    \resizebox{\linewidth}{!}{
        \begin{tabular}{ccccccc}
            \makecell[c]{CFG} & \makecell[c]{PB valid} & \makecell[c]{PC clashes} & \makecell[c]{PC SE} & \makecell[c]{Vina score} & \makecell[c]{Vina min} & \makecell[c]{Mol. weight} \\
            \makecell[c]{} & \makecell[c]{[\%] $\uparrow$} & \makecell[c]{[1] $\downarrow$} & \makecell[c]{[kcal/mol] $\downarrow$} & \makecell[c]{[kcal/mol] $\downarrow$} & \makecell[c]{[kcal/mol] $\downarrow$} & \makecell[c]{[g/mol]} \\
            \midrule
            0.0  & \textbf{71.0 $\pm$ 3.2} & 8.3 $\pm$ 1.0 & \textbf{155 $\pm$ ~~90} & -4.4 $\pm$ 0.3 & -6.5 $\pm$ 0.4 & 273 $\pm$ 19 \\
            0.5  & \underline{69.2 $\pm$ 3.6} & 6.6 $\pm$ 0.9 & \underline{265 $\pm$ 167} & -5.1 $\pm$ 0.3 & -6.6 $\pm$ 0.4 & 289 $\pm$ 22 \\
            1.0  & 64.1 $\pm$ 3.9 & 5.8 $\pm$ 0.8 & 382 $\pm$ 250 & -5.3 $\pm$ 0.3 & \underline{-6.7 $\pm$ 0.4} & 300 $\pm$ 23 \\
            1.5  & 58.4 $\pm$ 4.3 & 5.1 $\pm$ 0.7 & 511 $\pm$ 308 & \underline{-5.5 $\pm$ 0.4} & \underline{-6.7 $\pm$ 0.4} & 308 $\pm$ 24 \\
            2.0  & 53.5 $\pm$ 4.5 & \underline{4.7 $\pm$ 0.7} & 544 $\pm$ 390 & \textbf{-5.6 $\pm$ 0.4} & \underline{-6.7 $\pm$ 0.4} & 316 $\pm$ 25 \\
            2.5  & 47.9 $\pm$ 4.6 & \textbf{4.5 $\pm$ 0.6} & 783 $\pm$ 574 & \textbf{-5.6 $\pm$ 0.5} & \textbf{-6.8 $\pm$ 0.4} & 323 $\pm$ 26 \\
        \end{tabular}
    }
    \label{tab:results_cfg_ablation_crossdocked}
\end{table}

This improvement comes at the cost of molecular quality. PB validity decreases steadily as $\omega$ increases, reaching 47.9\% on CrossDocked and 55.2\% on SPINDR at $\omega=2.5$. At the same time, strain energy rises substantially, indicating that excessive guidance can drive the sampler toward strained or implausible conformations. The minimized Vina score varies only weakly across guidance scales, suggesting that local minimization partially compensates for differences in the initially generated poses.

\begin{table}[ht]
    \caption{Impact of \ac{cfg} on the quality of molecules generated by NEAT-POCKET trained on the SPINDR dataset (best in bold, second-best underlined)}
    \setlength{\tabcolsep}{2pt}
    \resizebox{\linewidth}{!}{
        \begin{tabular}{ccccccc}
            \makecell[c]{CFG} & \makecell[c]{PB valid} & \makecell[c]{PC clashes} & \makecell[c]{PC SE} & \makecell[c]{Vina score} & \makecell[c]{Vina min} & \makecell[c]{Mol. weight} \\
            \makecell[c]{} & \makecell[c]{[\%] $\uparrow$} & \makecell[c]{[1] $\downarrow$} & \makecell[c]{[kcal/mol] $\downarrow$} & \makecell[c]{[kcal/mol] $\downarrow$} & \makecell[c]{[kcal/mol] $\downarrow$} & \makecell[c]{[g/mol]} \\
            \midrule
            0.0  & \textbf{80.5 $\pm$ 1.4} & 4.8 $\pm$ 0.3 & ~~\textbf{32 $\pm$ ~~15} & -4.8 $\pm$ 0.2 & -6.4 $\pm$ 0.2 & 281 $\pm$ 10 \\
            0.5  & \underline{80.3 $\pm$ 1.6} & 3.7 $\pm$ 0.3 & ~~\underline{62 $\pm$ ~~34} & -5.5 $\pm$ 0.2 & -6.8 $\pm$ 0.2 & 307 $\pm$ 12 \\
            1.0  & 75.2 $\pm$ 2.0 & 3.2 $\pm$ 0.2 & ~~92 $\pm$ ~~55 & -5.8 $\pm$ 0.2 & \underline{-6.9 $\pm$ 0.2} & 324 $\pm$ 13 \\
            1.5  & 68.6 $\pm$ 2.4 & 2.7 $\pm$ 0.2 & 138 $\pm$ ~~80 & \underline{-5.9 $\pm$ 0.2} & \textbf{-7.0 $\pm$ 0.2} & 338 $\pm$ 13 \\
            2.0  & 62.1 $\pm$ 2.8 & \underline{2.4 $\pm$ 0.2} & 181 $\pm$ 103 & \textbf{-6.0 $\pm$ 0.2} & \textbf{-7.0 $\pm$ 0.2} & 350 $\pm$ 14 \\
            2.5  & 55.2 $\pm$ 2.8 & \textbf{2.3 $\pm$ 0.2} & 235 $\pm$ 134 & \textbf{-6.0 $\pm$ 0.2} & \textbf{-7.0 $\pm$ 0.2} & 360 $\pm$ 14 \\
        \end{tabular}
    }
    \label{tab:results_cfg_ablation_spindr}
\end{table}

Based on this trade-off, we use $\omega=0.5$ as the default setting for NEAT-POCKET. Compared with unguided sampling, this value substantially reduces clashes and improves raw Vina scores on both datasets, while retaining most of the PB validity and avoiding the large increase in strain energy observed at higher guidance scales.

\subsection{Prefix-conditioned generation}\label{exp:results-prefix}

\begin{table*}[t]
    \caption{Performance of NEAT-POCKET vs. DiffSBDD on completing fragments of different sizes derived from the CrossDocked reference ligands (best in bold, second-best underlined)}
    \centering
        \begin{tabular}{lccccccc}
            Prefix 
            & \makecell[c]{PB valid} 
            & \makecell[c]{PC clashes} 
            & \makecell[c]{PC SE} 
            & \makecell[c]{Vina score} 
            & \makecell[c]{Vina min} 
            & \makecell[c]{Molecular weight} 
            & \makecell[c]{Unique} \\
            & \makecell[c]{[\%] $\uparrow$} 
            & \makecell[c]{[1] $\downarrow$} 
            & \makecell[c]{[kcal/mol] $\downarrow$} 
            & \makecell[c]{[kcal/mol] $\downarrow$} 
            & \makecell[c]{[kcal/mol] $\downarrow$} 
            & \makecell[c]{[g/mol]} 
            & \makecell[c]{[\%] $\uparrow$} \\
            \midrule
            \midrule
            \multicolumn{8}{l}{\textsc{DiffSBDD}}\\
            \midrule
            none 
            & \underline{49.6 $\pm$ 2.6} 
            & \underline{14.3 $\pm$ 2.0} 
            & 891 $\pm$ 271 
            & ~\textbf{-3.6 $\pm$ 0.8} 
            & \textbf{-6.3 $\pm$ 0.4} 
            & 296 $\pm$ 14 
            & \underline{97.8 $\pm$ 0.3} \\
            smallest 
            & 29.2 $\pm$ 5.0 
            & 36.5 $\pm$ 5.9 
            & \textbf{262 $\pm$ 160}
            & 14.2 $\pm$ 3.7 
            & -3.7 $\pm$ 0.5 
            & 280 $\pm$ 16 
            & \textbf{99.5 $\pm$ 0.5} \\
            second-largest 
            & 40.8 $\pm$ 4.8 
            & 22.3 $\pm$ 4.5 
            & \underline{275 $\pm$ 177}
            & ~~4.5 $\pm$ 2.6 
            & -4.7 $\pm$ 0.4 
            & 287 $\pm$ 16 
            & 98.3 $\pm$ 1.0 \\
            largest 
            & \textbf{51.8 $\pm$ 5.0} 
            & \textbf{14.1 $\pm$ 2.5} 
            & 299 $\pm$ 225
            & ~\underline{-2.2 $\pm$ 1.2} 
            & \underline{-6.1 $\pm$ 0.4} 
            & 306 $\pm$ 16 
            & 86.9 $\pm$ 4.8\\
            \midrule
            \midrule
            \multicolumn{8}{l}{\textsc{NEAT-POCKET (CrossDocked)}}\\
            \midrule
            none           
            & 69.2 $\pm$ 3.6 
            & \underline{6.6 $\pm$ 0.9} 
            & 265 $\pm$ 167 
            & -5.1 $\pm$ 0.3 
            & -6.6 $\pm$ 0.4 
            & 289 $\pm$ 22 
            & \textbf{92.8 $\pm$ 3.0} \\
            smallest       
            & 73.7 $\pm$ 4.1 
            & 7.0 $\pm$ 1.0 
            & 176 $\pm$ 118 
            & -5.4 $\pm$ 0.4 
            & \underline{-6.9 $\pm$ 0.4} 
            & 307 $\pm$ 23 
            & \underline{78.0 $\pm$ 5.9} \\
            second-largest 
            & \underline{79.4} $\pm$ 3.8 
            & \textbf{6.5 $\pm$ 0.9} 
            & \underline{105 $\pm$ ~~82} 
            & \underline{-5.7 $\pm$ 0.4} 
            & \underline{-6.9 $\pm$ 0.5} 
            & 307 $\pm$ 24 
            & 62.2 $\pm$ 7.1 \\
            largest        
            & \textbf{81.8 $\pm$ 5.2} 
            & 7.4 $\pm$ 1.2 
            & ~~\textbf{78 $\pm$ ~~51} 
            & \textbf{-6.1 $\pm$ 0.5} 
            & \textbf{-7.3 $\pm$ 0.4} 
            & 325 $\pm$ 26 
            & 42.4 $\pm$ 7.1 \\
        \end{tabular}
    \label{tab:results_prefix_completion_crossdocked}
\end{table*}

\begin{table*}[t]
    \caption{Performance of NEAT-POCKET vs. FLOWR on completing fragments of different sizes derived from the SPINDR reference ligands (best in bold, second-best underlined)}
    \centering
        \begin{tabular}{lccccccc}
            Prefix 
            & \makecell[c]{PB valid} 
            & \makecell[c]{PC clashes} 
            & \makecell[c]{PC SE} 
            & \makecell[c]{Vina score} 
            & \makecell[c]{Vina min} 
            & \makecell[c]{Molecular weight} 
            & \makecell[c]{Unique} \\
            & \makecell[c]{[\%] $\uparrow$} 
            & \makecell[c]{[1] $\downarrow$} 
            & \makecell[c]{[kcal/mol] $\downarrow$} 
            & \makecell[c]{[kcal/mol] $\downarrow$} 
            & \makecell[c]{[kcal/mol] $\downarrow$} 
            & \makecell[c]{[g/mol]} 
            & \makecell[c]{[\%] $\uparrow$} \\
            \midrule
            \midrule
            \multicolumn{8}{l}{\textsc{FLOWR}}\\
            \midrule
            none 
            & 75.8 $\pm$ 1.0 
            & 4.6 $\pm$ 0.4 
            & 35 $\pm$ ~~9
            & -7.0 $\pm$ 0.2 
            & -7.7 $\pm$ 0.2 
            & 395 $\pm$ ~~6 
            & \textbf{99.9 $\pm$ 0.1} \\
            smallest 
            & 57.7 $\pm$ 3.5 
            & 4.5 $\pm$ 0.4 
            & \underline{22 $\pm$ ~~9}
            & \textbf{-7.4 $\pm$ 0.3} 
            & \textbf{-8.0 $\pm$ 0.3} 
            & 381 $\pm$ 15 
            & \underline{95.6 $\pm$ 1.6} \\
            second-largest 
            & 41.8 $\pm$ 4.0 
            & \underline{4.4 $\pm$ 0.4} 
            & \underline{22 $\pm$ 10}
            & \underline{-7.2 $\pm$ 0.3} 
            & \underline{-7.8 $\pm$ 0.3} 
            & 383 $\pm$ 16 & 95.3 $\pm$ 1.7 \\
            largest 
            & 31.1 $\pm$ 3.7 
            & \textbf{4.2 $\pm$ 0.4} 
            & \textbf{19 $\pm$ ~~9}
            & \underline{-7.2 $\pm$ 0.3} 
            & \underline{-7.8 $\pm$ 0.3} 
            & 391 $\pm$ 16 
            & 88.6 $\pm$ 3.1 \\
            \midrule
            \midrule
            \multicolumn{8}{l}{\textsc{NEAT-POCKET (spindr)}}\\
            \midrule
            none           
            & 80.3 $\pm$ 1.6 
            & \textbf{3.7 $\pm$ 0.3} 
            & 62 $\pm$ 34 
            & -5.5 $\pm$ 0.2 
            & -6.8 $\pm$ 0.2 
            & 307 $\pm$ 12 
            & \textbf{97.6 $\pm$ 1.0} \\
            smallest       
            & 81.7 $\pm$ 2.0 
            & \underline{4.2 $\pm$ 0.3} 
            & 50 $\pm$ 32 
            & -6.1 $\pm$ 0.3 
            & -7.3 $\pm$ 0.3 
            & 328 $\pm$ 12 
            & \underline{86.1 $\pm$ 2.7} \\
            second-largest 
            & \underline{83.6 $\pm$ 2.4} 
            & \underline{4.2 $\pm$ 0.4} 
            & \underline{38 $\pm$ 24} 
            & \underline{-6.3 $\pm$ 0.2} 
            & \underline{-7.4 $\pm$ 0.2} 
            & 340 $\pm$ 13 
            & 74.1 $\pm$ 3.6 \\
            largest        
            & \textbf{88.8 $\pm$ 2.1} 
            & 4.4 $\pm$ 0.4 
            & \textbf{25 $\pm$ 17} 
            & \textbf{-7.0 $\pm$ 0.3} 
            & \textbf{-8.0 $\pm$ 0.3} 
            & 356 $\pm$ 13 
            &  52.2 $\pm$ 4.6 \\
        \end{tabular}
    \label{tab:results_prefix_spindr}
\end{table*}

\begin{figure*}[t]
\centering
\includegraphics[width=1.0\linewidth]{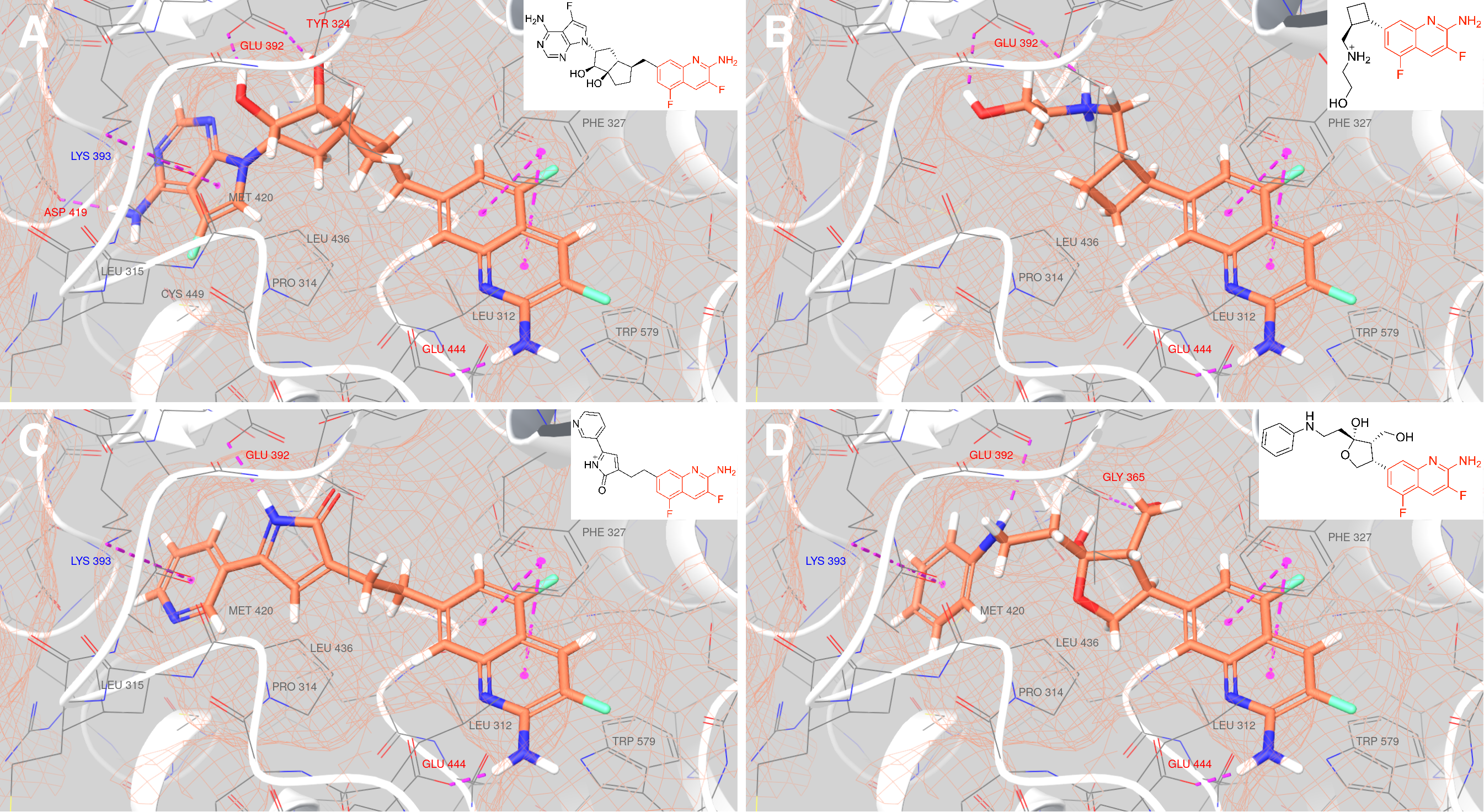}
\caption{Representative examples of pocket- and prefix-conditioned 3D molecule generation with NEAT-POCKET. Molecules were generated from a fixed prefix (highlighted in red) in the binding pocket of the human PRMT5:MEP50 (PDB: 7KID). (A) Experimentally resolved reference ligand (5,5-bicyclic inhibitor 72). (B--D) Representative generated ligands.}
\label{fig:7KID}
\end{figure*}

\paragraph*{NEAT-POCKET naturally supports fragment completion.}
A primary practical application of NEAT-POCKET is lead optimization, where fragments of a promising compound are modified or extended to improve pharmacodynamic or pharmacokinetic properties. The autoregressive architecture of NEAT-POCKET is well suited to this setting because it naturally supports conditional molecule completion: given a partial molecular prefix and a target protein pocket, the model generates the remaining atoms to complete the ligand.

\paragraph*{We evaluate prefix completion from BRICS-derived fragments.}
To benchmark NEAT-POCKET's prefix completion capabilities, we tasked NEAT-POCKET with generating complete molecules from fragments derived from the reference ligands in the CrossDocked and SPINDR test sets. 
Each ligand was decomposed using the BRICS fragmentation algorithm,\cite{Degen2008BRICS} and the resulting fragments were used as molecular prefixes. For each ligand, we considered three prefix sizes: the largest fragment, the second-largest fragment, and the smallest fragment. 
On CrossDocked, the largest fragments contained on average 10.4 heavy atoms, corresponding to 48\% of the original ligand. The second-largest and smallest fragments contained 5.1 and 1.2 heavy atoms, corresponding to 23\% and 7\%, respectively. 
On SPINDR, the corresponding values were 10.4 heavy atoms, or 41\%, for the largest fragments; 6.0 heavy atoms, or 23\%, for the second-largest fragments; and 1.2 heavy atoms, or 5\%, for the smallest fragments. 
Detailed distributions of fragment sizes, both in absolute terms, \textit{i.e.}, number of heavy atoms, and relative terms, \textit{i.e.}, fraction of the original ligand, are provided in the SI, Section~\ref{si:fragment_generation}.

\paragraph*{Larger prefixes improve validity, strain energies, and docking scores.}
The results of the prefix-completion experiments are summarized in Tables~\ref{tab:results_prefix_completion_crossdocked} and~\ref{tab:results_prefix_spindr}. 
Compared with \textit{de novo} generation without a prefix, fragment conditioning generally improves the quality of the generated molecules, with stronger effects for larger prefixes.
On CrossDocked, PB validity increases monotonically from 69.2\% without a prefix to 81.8\% when conditioning on the largest fragments. 
On SPINDR, PB validity increases from 80.3\% without a prefix to 88.8\% for the largest fragments. 
Strain energy decreases in parallel, from 265~kcal/mol to 78~kcal/mol on CrossDocked and from 62~kcal/mol to 25~kcal/mol on SPINDR. 
Thus, fragment conditioning guides the model toward more chemically and geometrically plausible molecules, especially when the supplied fragment accounts for a large fraction of the reference ligand.
Docking scores show the same trend. On CrossDocked, the raw Vina score improves from $-5.1$~kcal/mol without a prefix to $-6.1$~kcal/mol for the largest fragments. On SPINDR, it improves from $-5.5$ to $-7.0$~kcal/mol. 
Molecular weight increases with prefix size, suggesting that larger fragments provide a stronger structural prior and constrain generation toward larger, more reference-like ligands and binding modes.

\paragraph*{Larger prefixes reduce structural diversity.}
The improvements in validity, strain energies, and docking scores come at the cost of reduced molecular diversity. Structural uniqueness decreases steadily with prefix size, from 92.8\% for \textit{de novo} generation to 42.4\% for the largest prefixes on CrossDocked, and from 97.6\% to 52.2\% on SPINDR. This reflects the central trade-off in fragment-conditioned generation: larger prefixes provide more information and better constrain the model toward high-quality, pocket-compatible completions, but they also restrict the accessible chemical space.

\paragraph*{DiffSBDD's inpainting procedure does not yield adequate prefix-conditioned ligands.}
We next compared NEAT-POCKET trained on CrossDocked with DiffSBDD's inpainting procedure. 
To ensure a consistent input setup, we used the same prefixes but removed hydrogen atoms, since DiffSBDD uses a hydrogen-implicit representation. Additional details are provided in the SI, Section~\ref{si:baselines}.
Across all prefix-conditioned settings, DiffSBDD produces substantially fewer PB-valid molecules, more protein--ligand clashes, and less favorable Vina scores than NEAT-POCKET. Moreover, unlike NEAT-POCKET, DiffSBDD does not consistently improve when a prefix is provided. For several metrics, its best or second-best results are obtained without prefix conditioning, and Vina scores deteriorate substantially when small or medium prefixes are imposed. These results indicate that inpainting provides a general mechanism for imposing partial structural constraints, but does not reliably translate into chemically valid and pocket-compatible fragment growth.

\paragraph*{FLOWR often fails to preserve the supplied prefix.}
FLOWR is a stronger baseline because it includes fragment-conditioned examples during training. However, we observed that FLOWR often fails to preserve the supplied prefix exactly: the generation procedure can alter the prefix atom types or shift the prefix away from its original 3D placement in the binding pocket. This behavior is undesirable for lead optimization, where the goal is to retain a known binding motif while exploring alternative substituents or extensions. We therefore marked molecules as invalid if they did not preserve the prefix structure or if any prefix atom was displaced by more than 1~\AA. Under this criterion, FLOWR achieves favorable docking scores, but its PB validity drops sharply when a prefix is provided: from 75.8\% without a prefix to 57.7\%, 41.8\%, and 31.1\% for the smallest, second-largest, and largest fragments, respectively. By contrast, NEAT-POCKET is substantially more robust in the prefix-completion setting and improves with prefix size, reaching 88.8\% PB validity for the largest fragments. Since NEAT-POCKET uses the prefix as the initial state for autoregressive generation, it preserves both the fragment's chemical identity and its pose in the pocket by construction.

\paragraph*{NEAT-POCKET preserves binding motifs while proposing new extensions.}
Figure~\ref{fig:7KID} illustrates the prefix-completion capabilities of NEAT-POCKET in the binding pocket of the human PRMT5:MEP50 complex (PDB: 7KID). In this example, NEAT-POCKET was conditioned on the same molecular prefix for all generations. The prefix corresponds to the 2-amino-3,5-difluoroquinoline group of the crystallographic ligand and is highlighted in red in the 2D depictions. In the crystallographic complex, shown in panel~A, this prefix is anchored primarily by a $\pi$--$\pi$ stacking interaction with Phe327, hydrophobic contacts with Leu312 and Trp579, and a hydrogen-bonding interaction between the amino group and Glu444. In addition, the ligand forms hydrogen bonds with Tyr324, Glu392, Asp419, and Cys449, a $\pi$-cationic interaction with Lys393, and hydrophobic interactions with Pro314, Leu315, Leu436, and Met420. 

Across the generated molecules shown in panels~B--D, NEAT-POCKET preserves both the chemical identity of the prefix and its 3D placement, thereby retaining all its anchor interactions with Leu312, Pro314, Phe327, Glu444, and Trp579. The generated molecules further extend into the same region of the pocket as the reference ligand and recover several of the observed interactions, notably the hydrogen bonding to Glu392 (B--D), and hydrophobic interactions with Pro314 (B--D), and Leu315, Leu436, and Met420 (C and D), while exploring alternative chemistry. In molecule~B, a cyclobutane linker provides a compact, rigid bridge from the conserved prefix to a polar group that forms hydrogen bonds with Glu392. In molecules~C and D, NEAT-POCKET generated extensions that preserve the Lys393 cation--$\pi$ interaction and the hydrogen bond to Glu392. Molecule~D additionally introduces a new favorable hydrogen-bonding contact with Gly365.

\paragraph*{NEAT-POCKET enables reliable pocket-conditioned fragment completion.}
Overall, these results demonstrate that NEAT-POCKET can perform pocket-conditioned completion of partial ligand fragments. Prefix size acts as a practical control parameter: small prefixes allow greater diversity, whereas larger prefixes improve validity, strain energy, and docking performance by more strongly anchoring generation to the known ligand structure. The comparisons with DiffSBDD and FLOWR show that this behavior is not simply a generic consequence of using fragments. DiffSBDD does not generate sufficiently adequate molecules from prefixes, while FLOWR frequently violates the prefix-preservation requirements of the task. In contrast, NEAT-POCKET preserves the supplied fragment by construction, maintains key ligand--protein interaction motifs, and proposes alternative extensions that form plausible additional contacts with nearby residues.

\subsection{Limitations}

As an autoregressive model, NEAT-POCKET can suffer from error propagation during sequential generation, potentially affecting molecular quality, particularly for larger compounds. Its performance is also influenced by the quality and biases of the training data. CrossDocked is a large and useful benchmark for pocket-conditioned generation, but it contains approximations such as missing hydrogen atoms and cross-docked, rather than co-crystallized, ligand--receptor complexes. Consequently, some ligand--pocket interactions in the training data may be geometrically or chemically unrealistic, and models trained on this dataset may reproduce such artifacts. In addition, the current formulation treats protein pockets as rigid, although real proteins are dynamic and can undergo conformational changes upon ligand binding. NEAT-POCKET also does not explicitly model water molecules, despite their important role in mediating protein-ligand interactions. Finally, as with all other molecular generators, NEAT-POCKET should be viewed as an ideation tool: generated molecules require downstream assessment, including visual inspection by medicinal chemists, synthetic-feasibility analysis, and experimental validation.

\section{Conclusions}\label{sec:conclusions}
We present NEAT-POCKET, a pocket-conditioned extension of the autoregressive NEAT framework for generating three-dimensional molecules within protein-binding pockets. NEAT-POCKET encodes pocket information with a fine--coarse--fine transformer architecture and incorporates this information into the molecular generator through cross-attention and global adaptive layer normalization. Classifier-free guidance enables control over the strength of pocket conditioning, allowing users to balance pocket complementarity against the physical plausibility of the generated molecules.

On the CrossDocked benchmark, NEAT-POCKET achieved competitive performance relative to Pocket2Mol, TargetDiff, DiffSBDD, and DrugFlow. Among the generated molecules, it obtained the second-highest PoseBusters validity and the lowest average number of protein--ligand clashes. It also produced molecules with the closest aggregate physicochemical similarity to the CrossDocked training distribution. On the SPINDR benchmark, NEAT-POCKET showed higher PoseBusters validity than FLOWR, with fewer protein–ligand clashes, a more closely matched aggregate physicochemical similarity to the SPINDR training data, and faster sampling.

A central novelty that distinguishes NEAT-POCKET from existing baseline approaches is its ability to generate molecules from arbitrary fragments without architectural changes, retraining, or task-specific modifications. In fragment-completion experiments on CrossDocked and SPINDR, larger prefixes generally improved PoseBusters validity, strain energy, and docking performance, while reducing structural diversity. This behavior reflects the expected trade-off in fragment-conditioned generation: larger prefixes impose stronger structural constraints and better anchor the generated molecule within the pocket, but they also restrict the accessible chemical space. 

A key practical advantage of NEAT-POCKET is its sampling efficiency. NEAT-POCKET required only 4 seconds on average to generate 100 molecules for a pocket. This corresponds to an approximately nine-fold speed-up over the second-fastest baseline, FLOWR, and more than an order-of-magnitude improvement over the remaining baselines. Computational efficiency is important for structure-based design workflows, where many binding pockets, molecular prefixes, and optimization hypotheses may need to be explored.

Overall, NEAT-POCKET provides a fast, flexible, and competitive approach to pocket-conditioned 3D molecular generation. Its ability to combine \textit{de novo} generation, explicit hydrogen modeling, controllable pocket conditioning, and fragment-based completion makes it particularly suitable for early-stage hit generation and lead optimization workflows, where rapid exploration of chemically plausible design hypotheses is essential.

\section*{Author contributions}
Conceptualization: RAJ, DR.
Methodology: RAJ, DR.
Software: RAJ, DR.
Data curation: RAJ, DR.
Result analysis: RAJ, DR.
Visualizations: RAJ, DR.
Case studies: RAJ, DR, JK.
Writing (original draft): RAJ, DR.
Writing (review and editing): RAJ, DR, JK.
Project administration: TL, JK.
Funding acquisition: TL, JK.

\section*{Conflicts of interest}
There are no conflicts to declare.

\section*{Data availability}

The source code for NEAT-POCKET is available at \url{https://github.com/molinfo-vienna/NEAT-POCKET}. The model weights and generated molecules are available at \url{https://doi.org/10.6084/m9.figshare.33426877}. 
Supplementary Information 
includes additional information about system specifications and implementation details (S1), the use of baseline models (S2), the hydrogenation of CrossDocked data (S3), evaluation metrics (S4), additional evaluation results (S5), strain energies (S6), the generation of fragments (S7), and examples of generated molecules (S8).

\section*{Acknowledgments}

The financial support received for the Christian Doppler Laboratory for Molecular Informatics in the Biosciences by the Austrian Federal Ministry of Labour and Economy, the National Foundation for Research, Technology and Development, the Christian Doppler Research Association, Boehringer-Ingelheim RCV GmbH \& Co KG and BASF SE is gratefully acknowledged.


\balance

\bibliography{rsc}
\bibliographystyle{rsc}

\onecolumn

\section*{Supplementary information}

\setcounter{section}{0}
\renewcommand{\thesection}{S\arabic{section}}

\setcounter{figure}{0}
\renewcommand{\thefigure}{S\arabic{figure}}

\setcounter{table}{0}
\renewcommand{\thetable}{S\arabic{table}}

\section{Implementation details}\label{si:implementation}

\subsection{System and software specification}\label{si:hardsware_software}

Training was performed on an AlmaLinux 9.7 system with an NVIDIA RTX Pro 6000 Blackwell GPU with 96 GB of GDDR7 VRAM and an AMD EPYC 9555 64-Core Processor. Inference and data analysis was performed on a Rocky Linux 9.7 system with an NVIDIA GeForce RTX 4090 GPU with 24 GB of GDDR6X VRAM and an AMD Ryzen 9 7950X 16-Core Processor. We used CUDA 13.0 for GPU acceleration. Our model is implemented in Python (v3.11.15) with PyTorch (v2.11.0),\cite{Ansel2024PyTorch} PyTorch Geometric (v2.8.0),\cite{Fey2019PyG} and Pytorch Lightning (v2.5.5).\cite{Falcon2025Lightning} Model training was monitored with Tensorboard (v2.20.0). For chemical data processing, we employed RDKit (v2025.9.1).\cite{rdkit_2025} For the evaluation of generated molecules we used PoseBusters (v0.6.5) \cite{Buttenschoen2023PoseBusters} and PoseCheck (v1.3.1).\cite{Harris2023PoseCheck} For visualizations we used matplotlib (v3.10.9),\cite{Hunter2007MatPlotLib} seaborn (v0.13.2),\cite{Waskom2021Seaborn} and Schroedinger Maestro (v2025.3).

\subsection{Hyperparameters}\label{si:hyperparameters}

Table~\ref{tab:hyperparameters_crossdocked} summarizes the hyperparameters of our best-performing model. 
The model has 242M parameters. 
The loss function was minimized with the AdamW \citep{Kingma2017Adam} optimizer with parameters $\beta_1=0.9$ and $\beta_2=0.95$.
We further used gradient clipping to a maximum gradient norm of 1.0. 
We implemented a cosine annealing learning rate schedule with 10 linear warm-up epochs and set the minimum learning rate to 10\% of the initial learning rate. 
Pretraining NEAT on GEOM-Drugs took 40 hours, and finetuning NEAT-POCKET on CrossDocked and SPINDR took 37 and 40 hours, respectively, on the hardware specified above. We trained on CrossDocked for 2000 epochs with a batch size of 256, and on SPINDR with a batch size of 512. The remaining hyperparameters settings are identical.

\begin{table}[htb!]
    \caption{Hyperparameters of NEAT-POCKET pretrained on GEOM-Drugs and finetuned on CrossDocked.}
    \centering
    \resizebox{0.8\linewidth}{!}{
    \begin{tabular}{ll}
    \textbf{Parameter} & \textbf{Value} \\
    \midrule
        batch size & 256 \\
        bias & true \\
        CFG dropout & 0.2 \\
        clash penalty & 4.0 \\
        clash penalty margin & 0.0 \\
        dropout & 0.1 \\
        epochs & 2,000 \\
        flow matching head hidden dimension  & 1,536 \\
        flow matching head number of layers & 6 \\
        flow matching noise standard deviation $\sigma$ & 2.5 \\
        integrator & Euler-Maruyama \\
        learning rate, weight decay & $2.5 \times 10^{-5}$, $10^{-6}$ \\
        learning rate decay number of epochs (cosine annealing scheduling) & 2,000 \\
        learning rate minimum (cosine annealing scheduling) & $10\%$ \\
        learning rate warm-up number of epochs (cosine annealing scheduling) & 10 \\
        number of epochs with frozen pretrained weights & 150 \\
        pocket transformer heads & 12 \\
        pocket transformer hidden dimension & 768 \\
        pocket transformer number of atom-level layers & 1 \\
        pocket transformer number of residue-level layers & 4 \\
        source set noise fraction $\delta$ & 0.5 \\
        source set noise standard deviation & 0.5 \\
        source set transformer heads & 12 \\
        source set transformer hidden dimension & 768 \\
        source set transformer number of layers & 12 \\
        time step resampling & 4 \\
        time step sampling & logit normal \\
    \end{tabular}
    }
    \label{tab:hyperparameters_crossdocked}
\end{table}

\subsection{Bond prediction model}\label{si:bond_prediction}

As described in Section~\ref{met:bond_prediction}, we use a bond prediction model during inference to infer the molecular graph of the final molecule from the generated positions and atom types. The bond prediction model is a graph neural network with the following architecture. We first embed atom types using a learned embedding layer. These embeddings are then passed as node features, together with the atom positions, to a simple E(3)-invariant gated point network (v2103), as implemented in the e3nn library.\cite{geiger2022e3nn} We chose this model type because it is relatively lightweight and accounts for distances and angles between atoms.
We use three convolutional layers and a hidden layer with a dimension of 64. For each edge in the input radius graph, we sum the node embeddings of the source and target nodes and add a learned encoding of the Euclidean distance between the two atoms. We then use a three-layer MLP and a softmax layer to map this edge representation to a categorical distribution over the possible edge types. The model is supervised using a cross-entropy loss over the true bond types. It is trained with the AdamW optimizer \cite{Kingma2017Adam} using an initial learning rate of $10^{-3}$, and the validation loss is monitored for early stopping. Training terminates after 60 epochs on CrossDocked and after 17 epochs on SPINDR.

\section{Baselines}\label{si:baselines}

Although evaluation protocols for \textit{unconditional} 3D molecular generation have become relatively standardized, we found that the evaluation of \textit{pocket-conditional} 3D molecular generation varies substantially across the literature. These differences affect both the generation procedure itself and the downstream processing steps used to convert generated structures into finalized molecules. One important source of variation is the inference protocol used to obtain a fixed number of samples per protein pocket. In several publicly available baseline implementations, the sampling loop is continued until a predefined number of valid molecules, often 100 per pocket, has been obtained. We argue that this procedure can bias comparisons because invalid, duplicate, or otherwise unsuccessful generations are excluded from the effective sample count. For a fair comparison between generative models, each method should instead be evaluated under the same fixed sampling budget. In our evaluation, each model was allowed to generate exactly 100 molecules per pocket, enabling the computation of validity metrics in a manner analogous to the standard evaluation of unconditional molecular generative models.

Additional variation arises from the docking software used to score generated ligands. Common choices in the literature include the original AutoDock Vina implementation, QVina, and binaries distributed with Gnina. We found that different Vina releases yield widely varying docking scores. In this work, we standardized docking-based evaluation by using the Gnina 1.1 binary provided by \citet{McNutt2021}.

Another important difference concerns the treatment of hydrogen atoms. NEAT-POCKET explicitly generates hydrogen atoms, which is uncommon in structure-based three-dimensional molecular generation, and none of the considered baselines produce hydrogens directly. Therefore, hydrogens must be added to the output of baseline models as a post-processing step before downstream evaluation. Both RDKit and OpenBabel provide open-source functionality for hydrogen addition. We evaluated both options and found that RDKit produced higher-quality structures in our pipeline. Moreover, using RDKit is consistent with the PoseCheck \cite{Harris2023PoseCheck} and PoseBusters \cite{Buttenschoen2023PoseBusters} packages, which are also used in our evaluation workflow. Accordingly, hydrogen atoms were added for all baseline methods using \texttt{RDKit}'s \texttt{Chem.AddHs(addCoords=True)} functionality. More information on hydrogenation procedures can be found in Section~\ref{si:hydrogens}.

Finally, the evaluated methods differ in whether they explicitly generate molecular connectivity. Some approaches generate only atom types and three-dimensional coordinates, requiring a post hoc bond-inference step, whereas others directly predict the molecular graph in addition to atom positions. We did not alter this aspect of any baseline method. Instead, bond information was obtained according to the procedure specified by each original implementation. 

Overall, to harmonize the evaluation pipeline, we generated exactly 100 molecules per pocket for each method and added hydrogen atoms using RDKit with \texttt{Chem.AddHs(addCoords=True)} for all hydrogen-implicit baselines, and retained the bond-inference strategy defined by the respective method. In the following subsections, we describe the specific modifications made to the original baseline implementations to conform to this standardized evaluation protocol.

\subsection{Pocket2Mol}

Pocket2Mol \cite{Peng2022Pocket2Mol} was installed from the official repository provided by the authors \url{https://github.com/pengxingang/pocket2mol}. In the original inference procedure, molecules are generated sequentially until 100 valid, non-duplicate molecules have been obtained. Starting from the protein pocket, the model proposes a set of initial candidate atom positions and atom types, each of which is inserted into an active generation queue. At subsequent generation steps, the model samples a set of possible continuations for each active partial molecule, including the next atom placement, the corresponding atom type, and bond connectivity to previously placed atoms. As a result, the number of active partial molecules grows exponentially during the early stages of generation. To control the computational cost, the authors impose a maximum queue size of 500 partial molecules. When this limit is exceeded, only the 500 candidates with the highest model confidence are retained for further expansion, which typically occurs after approximately four generation steps. 

After each iteration, partial molecules are assessed to determine whether generation should continue. Terminated molecules are converted into RDKit molecule objects in the next step. A molecule is retained only if it can be converted to a SMILES string and subsequently reconstructed by RDKit, and if it is not a duplicate of a previously accepted molecule. Molecules failing this reconstruction check, as well as duplicate molecules, are discarded. The process is repeated until 100 molecules satisfying these criteria have been collected. 

For a fair comparison, we modified the Pocket2Mol sampling protocol to avoid selectively extending the generation process until 100 valid and unique molecules were obtained. In the modified procedure, generation is terminated after 100 molecular completion events, irrespective of whether the resulting structures can be reconstructed by RDKit. We also do not remove duplicate molecules. This modification ensures that the reported outcomes reflect the model's raw sampling behavior under a fixed generation budget, rather than a post-filtered set of successful, unique molecules.

\subsection{TargetDiff}

TargetDiff \cite{Guan2023TargetDiff} was obtained from the official repository provided by the authors at \url{https://github.com/guanjq/targetdiff}. The model generates three-dimensional atom positions and atom types for ligand structures, but does not directly predict bond connectivity or hydrogen atoms. Consequently, converting the generated point clouds into chemically valid molecular graphs requires a separate reconstruction step. This reconstruction is nontrivial because bond perception from heavy-atom coordinates alone is not uniquely defined, particularly in the absence of explicit hydrogens and when generated geometries deviate from idealized bond lengths and angles. In the original TargetDiff implementation, this issue is addressed using a modified version of the OpenBabel ``connect-the-dots'' procedure, with additional heuristics that allow, for example, deviations from standard bond-length criteria during connectivity assignment. The full reconstruction workflow is implemented in the repository's \texttt{reconstruct.py} file. For the evaluations reported here, we used the TargetDiff sampling and reconstruction pipeline as provided by the authors, without modifying the inference procedure.

\subsection{DiffSBDD}

DiffSBDD \cite{Schneuing2024DiffSBDD} was installed from the official repository available at \url{https://github.com/arneschneuing/DiffSBDD}. In the original inference procedure, generated molecules are explicitly filtered according to internal validity criteria before being returned. Specifically, a molecule is considered valid if its largest fragment, identified using RDKit's \texttt{Chem.GetMolFrags()} function, can be successfully sanitized by RDKit. The default sampling batch size is 120 molecules, and the model is allowed to perform up to 10 sampling iterations for each protein pocket. After each iteration, molecules passing the validity check are appended to an accumulated list of accepted samples. If this list contains more than 100 molecules, inference is terminated and the first 100 molecules in the list are returned. Thus, the original implementation effectively continues sampling until 100 valid molecules have been obtained.

To align DiffSBDD with our standardized evaluation protocol, we modified its inference procedure to impose a fixed generation budget. Specifically, we set the sampling batch size to exactly 100 molecules and restricted inference to a single sampling round per pocket. No additional sampling iterations were performed if fewer than 100 generated molecules passed the model’s internal processing steps. Thus, inference was terminated after one batch regardless of the number of molecules ultimately returned. This modification prevents the method from compensating for invalid generations through repeated sampling and ensures that DiffSBDD is evaluated under the same fixed-sample regime as the other baseline models.

For inpainting, the authors provide a dedicated inference script. We used this protocol with the default settings and supplied the same BRICS-derived fragments of the test-set ligands as those used to evaluate NEAT-POCKET. Since DiffSBDD uses a hydrogen-implicit molecular representation, we removed hydrogen atoms from all fragments. We then sanitized the generated molecules and filtered out fragmented structures.

\subsection{DrugFlow}

DrugFlow \cite{Schneuing2025DrugFlow} was installed from the official repository provided by the authors at \url{https://github.com/LPDI-EPFL/DrugFlow}. The DrugFlow inference procedure samples exactly the user-specified number of molecules and does not apply additional filtering steps during generation. Therefore, the model's raw sampling output already conforms to our fixed-budget evaluation protocol. Consequently, we used the generated DrugFlow samples directly in our evaluation pipeline without further modifications to the inference procedure.

\subsection{FLOWR}

FLOWR \cite{Cremer2026Flowr} was installed from the official repository provided by the authors at \url{https://github.com/jule-c/flowr}. The authors provide both hydrogen-implicit and hydrogen-explicit variants of the model. Since NEAT-POCKET uses a hydrogen-explicit representation, we used the hydrogen-explicit FLOWR model for all comparisons. As in DiffSBDD, the default FLOWR inference script samples molecules until 100 valid samples are obtained. For consistency with our evaluation protocol, we instead sampled a fixed batch of 100 molecules per pocket, irrespective of validity.
In its standard configuration, FLOWR uses the size of the test-set ligand when sampling molecules. Because this information would give FLOWR an advantage relative to other baselines, we instead sampled ligand sizes from the training-set size distribution, which is common practice for diffusion- and flow-matching-based molecular generation models. 

For the prefix-completion experiments, we enabled the scaffold-inpainting option in the FLOWR inference script. Since this option does not allow molecular size sampling, we used the default setting to provide the test-set ligand size during inference in this case. We then supplied the same BRICS-derived fragments of the test-set ligands as those used to evaluate NEAT-POCKET. We sanitized the generated molecules and filtered out fragmented structures. Since FLOWR’s inpainting procedure does not guarantee preservation of the input fragment or its 3D placement, we also filtered out molecules that did not contain the prefix or in which any prefix atom was displaced by more than 1 \AA.

\section{Post hoc hydrogenation of CrossDocked data}\label{si:hydrogens}

The widely used split of the CrossDocked dataset provided by \citet{Luo2021} contains no explicit hydrogen atoms. Across the combined training and test sets, which comprise 100,100 ligands, only 4,320 hydrogen atoms remain, corresponding to approximately 0.2\% of all ligand atoms. This is consequential because most previous pocket-conditioned 3D molecular generation models have been trained and evaluated on this dataset, and therefore also on molecular structures without explicit hydrogens.

The omission of hydrogen atoms is not merely a representational simplification. Hydrogen atoms affect the molecular volume and shape of ligands and are directly involved in important noncovalent interactions, most notably hydrogen bonds. Moreover, protonation states determine formal charges, influence tautomeric forms, and affect donor and acceptor assignments. These properties are central to molecular recognition in protein binding sites. We therefore argue that models for protein-conditioned \textit{de novo} 3D molecular design should ultimately be trained with explicit hydrogen atoms.

A common counterargument is that hydrogens can be added after generation as a post-processing step. However, this is only straightforward when the heavy-atom structure has chemically consistent bond orders, aromaticity assignments, tautomeric states, valences, and formal charges. These quantities are not always fully specified in training data, nor are they always predicted by 3D molecular generators. Formal charges, in particular, are often omitted from generated molecular outputs. In the widely used version of CrossDocked, we found only 12 ligands with formal charges, all of which corresponded to zwitterionic N-oxide annotations. This number is unexpectedly low, suggesting that charge information may have been lost or altered during dataset preparation. Reliable hydrogen reconstruction is further complicated by the fact that protonation and charge states depend on experimental and environmental conditions, especially pH. Although CrossDocked structures originate from the PDB and are therefore assumed to approximate physiological conditions, this assumption is not guaranteed, and it is unclear to what extent protonation states were curated in the original structures.

The omission of hydrogens in previous work is likely motivated, at least in part, by computational considerations. Many recent pocket-conditioned generative models are based on diffusion or flow-matching architectures with equivariant message passing over dense graphs, for which computational and memory costs scale quadratically with the number of atoms. Since hydrogens can constitute a substantial fraction of all atoms in an all-atom molecular representation, omitting them substantially reduces the cost of training and inference. In addition, models trained on hydrogen-free data may appear to perform better under common evaluation metrics because they are required to predict fewer atoms and fewer atom types. Conversely, adding explicit hydrogens increases the number of predicted coordinates, thereby increasing the number of opportunities for geometric or chemical errors. This computational and statistical convenience, however, does not remove the underlying chemical ambiguity introduced by training on hydrogen-depleted structures.

Adding hydrogens post hoc requires determining both the correct number of hydrogens and plausible 3D coordinates, ideally without modifying the coordinates of the existing heavy atoms. For CrossDocked ligands, this is non-trivial because the heavy-atom structures can contain inconsistent bond orders, problematic aromaticity assignments, strained geometries, or missing charge information. In this work, we evaluated three approaches for reconstructing ligand hydrogen atoms. The first was RDKit’s built-in \texttt{Chem.AddHs(addCoords=True)} functionality. The second was Open Babel’s \texttt{AddHydrogens()} method. The third was a two-step RDKit procedure in which hydrogens are first added without coordinates using \texttt{Chem.AddHs(addCoords=False)} and then embedded using \texttt{AllChem.ConstrainedEmbed()}.

\begin{figure}
    \centering
    \includegraphics[width=1.0\linewidth]{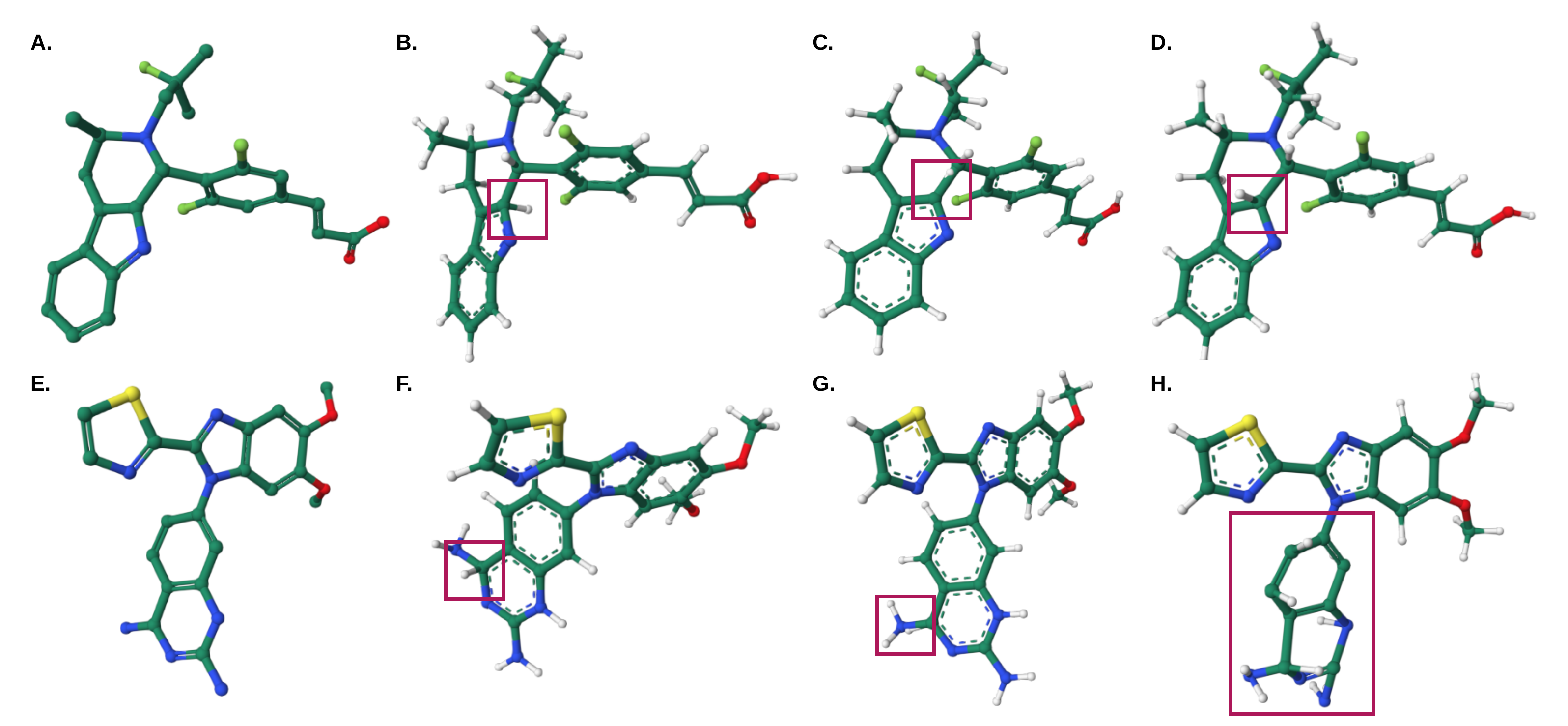}
    \caption{Two representative CrossDocked ligands are shown before and after hydrogen addition. Panels A--D show the first ligand: (A) raw input structure, (B) RDKit default hydrogen addition, (C) Open Babel hydrogen addition, and (D) RDKit constrained force-field-based hydrogen placement. Panels E--H show the same sequence for a second ligand. The raw molecules are displayed as kekulized structures, as provided in the input data, to illustrate the representation errors that can lead to incorrect perceptions of aromaticity, protonation assignments, and hydrogen placement.}
    \label{fig:hydrogens}
\end{figure}

The first approach, \texttt{Chem.AddHs(addCoords=True)}, is fast and simple, but it can fail or produce implausible hydrogen positions when the input heavy-atom structure is chemically inconsistent. Such failures can arise when the hydrogen-depleted molecule has an implausible kekulized representation, incorrect valence assignments, or problematic aromaticity perception. N-containing heteroaromatic rings are a common source of errors: if the input representation prevents the correct recovery of aromaticity or the correct assignment of protonation state, hydrogen placement may become chemically unreasonable. Examples of such failures are shown in Figure~\ref{fig:hydrogens}.

The Open Babel-based approach exhibits similar limitations. Although its hydrogen-placement heuristics differ from those used by RDKit, it also fails on chemically inconsistent inputs and can produce incorrect hydrogen positions for many of the same problematic structures. Representative examples are shown in Figure~\ref{fig:hydrogens}.

The third approach was motivated by the expectation that force-field-based placement might produce more realistic hydrogen geometries than purely heuristic coordinate assignment. In this procedure, hydrogens are added using RDKit, and their coordinates are generated via constrained embedding with the \ac{uff}, while attempting to preserve the original heavy-atom coordinates. In practice, however, the constrained embedding procedure can still perturb heavy-atom coordinates in some cases. This method is also substantially slower than direct hydrogen addition and failed to produce embeddings for 3.8\% of CrossDocked ligands. Moreover, it does not resolve the underlying problems caused by inconsistent valence, aromaticity, or charge assignments in the input structures, and it often produces strained or otherwise implausible geometries. Examples are shown in Figure~\ref{fig:hydrogens}.

Overall, we found RDKit’s \texttt{Chem.AddHs(addCoords=True)} to provide the most practical compromise in our pipeline. It is fast, widely used, and it produced the most satisfactory structures among the tested options, despite the limitations described above. Its use is also consistent with the hydrogen-addition procedures used by the PoseCheck and PoseBusters evaluation packages when input molecules do not already contain hydrogen atoms. Nevertheless, some incorrect hydrogen placements propagate into NEAT-POCKET’s outputs, illustrating that post hoc hydrogenation is only a pragmatic workaround for the limitations of CrossDocked, not a principled solution. We use CrossDocked in this work primarily to enable direct comparison with existing baselines, most of which were trained and evaluated on the same hydrogen-depleted data. The appropriate long-term solution is to train and evaluate structure-based generative models on datasets that natively contain explicit hydrogen atoms, together with curated bond orders, formal charges, and protonation states. In such a setting, hydrogen atoms should be part of the molecular representation and should be predicted directly by the model, rather than inferred during post-processing.

\section{Evaluation metrics}\label{si:metrics}

For each metric, we first compute the mean over generated molecules within each pocket. Let $x_{i}$ denote this per-pocket mean for pocket $i = 1,...,n$. We then report the mean across pockets,

\begin{equation}
    \bar{x} = \frac{1}{n} \sum_{i=1}^{n} x_{i}
\end{equation}

together with a 95\% confidence interval for the mean based on Student’s $t$ distribution,

\begin{equation}
    \bar{x} = t^{*}_{n-1,0.975} \frac{s}{\sqrt{n}}
\end{equation}

where $s$ is the sample standard deviation of $\{x_{i}\}$ with $n-1$ degrees of freedom, and $t^{*}_{n-1,0.975}$ is the 97.5th percentile of the $t$ distribution with $n-1$ degrees of freedom. Unless stated otherwise, values are reported as $\bar{x}$ $\pm$ the corresponding interval's half-width.

A generated molecule is \textbf{PB valid} if it can be successfully parsed by RDKit and passes all 22 conditional PoseBusters checks.\cite{Buttenschoen2023PoseBusters}

The \textbf{PoseCheck} package \cite{Harris2023PoseCheck} was used to evaluate protein–ligand binding poses by quantifying steric clashes, intermolecular interactions, and ligand \ac{se}. A clash occurs when the distance between two atoms is less than the sum of their van der Waals radii, with a tolerance of 0.5 $\mathring{\text{A}}$. PoseCheck was also used to compute the number of interactions (\ac{hba}, \ac{hbd}, hydrophobic interactions, and Van der Waals contacts) between ligand and pocket atoms.  PoseCheck uses ProLIF \cite{Bouysset2021ProLIF} internally to compute interactions. Interaction counts for the molecules generated by NEAT-POCKET, Pocket2Mol, TargetDiff, DiffSBDD, and DrugFlow are shown in Table~\ref{tab:results_metrics_2}. Finally, ligand \ac{se} is computed as the difference in internal energy between the generated ligand conformation and the geometry-optimized conformation, where optimization is performed with the Universal Force Field (UFF) \cite{Rappe1992UFF} as implemented in RDKit.
    
The Vina score was computed using the Gnina binary v1.1.\cite{McNutt2021} We report the score without any further optimization as \textbf{Vina score} and after minimization as \textbf{Vina min}. Since Vina scores are heavily dependent on the molecular weight of the docked ligands, we also report the average \textbf{molecular weight} of the generated molecules.

The \textbf{physchem rank} quantifies how closely the generated molecules align with the CrossDocked dataset based on a set of physicochemical descriptors. This rank is derived by evaluating each model’s proximity to the CrossDocked data across 31 physicochemical descriptors (15 general, 7 ring-size-related, and 9 element-type-related). The final rank for each model is the average of its individual descriptor ranks, with lower ranks indicating better performance. Below, we list all descriptors and detail their computation methods:

\begin{enumerate}
    \item The \textbf{molecular weight} is computed with RDKit's \texttt{Chem.rdMolDescriptors.CalcExactMolWt()} method.
    \item The \textbf{number of heavy atoms} is computed with \texttt{RDKit's Chem.rdMolDescriptors.CalcNumHeavyAtoms()} method.
    \item The \textbf{fraction of hetero atoms} is computed as the ratio between the outputs of RDKit's \texttt{Chem.rdMolDescriptors.CalcNumHeteroatoms()} and \texttt{Chem.rdMolDescriptors.CalcNumAtoms()} methods.
    \item The \textbf{fraction of halogen atoms} is computed as the ratio between the number of halogen atoms (F, Cl, Br, I) and the output of RDKit's \texttt{Chem.rdMolDescriptors.CalcNumAtoms()} method.
    \item The \textbf{fraction of rotatable bonds} is computed as the ratio between the outputs of \texttt{RDKit's Chem.rdMolDescriptors.CalcNumRotatableBonds()} and \texttt{Chem.rdChem.GetNumBonds()} methods.
    \item The \textbf{fraction of chiral centers} is computed as the ratio between the output of RDKit's \texttt{Chem.FindMolChiralCenters()} and \texttt{Chem.rdMolDescriptors.CalcNumAtoms()} methods.
    \item The \textbf{fraction of hydrogen bond acceptors} is computed as the ratio between the output of RDKit's \texttt{Chem.rdMolDescriptors.CalcNumHBA()} and \texttt{Chem.rdMolDescriptors.CalcNumAtoms()} methods.
    \item The \textbf{fraction of hydrogen bond donors} is computed as the ratio between the output of RDKit's \texttt{Chem.rdMolDescriptors.CalcNumHBD()} and \texttt{Chem.rdMolDescriptors.CalcNumAtoms()} methods.
    \item The \textbf{logP} is computed with RDKit's \texttt{Chem.rdMolDescriptors.CalcCrippenDescriptors()} method.
    \item The \textbf{quantitative estimate of drug-likeness (QED)} is computed with RDKit's \texttt{Chem.QED.qed()} method.
    \item The \textbf{number of rings} is computed with RDKit's \texttt{Chem.rdMolDescriptors.CalcNumRings()} method.
    \item The \textbf{fraction of aromatic rings} is computed as the ratio between RDKit's \texttt{Chem.rdMolDescriptors.CalcNumAromaticRings()} \texttt{Chem.rdMolDescriptors.CalcNumRings()} methods.
    \item The \textbf{fraction of aliphatic rings} is computed as the ratio between RDKit's \texttt{Chem.rdMolDescriptors.CalcNumAliphaticRings()} \texttt{Chem.rdMolDescriptors.CalcNumRings()} methods.
    \item The \textbf{fraction of bridgehead atoms} is computed as the ratio between RDKit's \texttt{Chem.rdMolDescriptors.CalcNumBridgeheadAtoms()} \texttt{Chem.rdMolDescriptors.CalcNumAtoms()} methods.
    \item The \textbf{fraction of spiro atoms} is computed as the ratio between RDKit's \texttt{Chem.rdMolDescriptors.CalcNumSpiroAtoms()} \texttt{Chem.rdMolDescriptors.CalcNumAtoms()} methods.
    \item The \textbf{ring size distribution} is calculated as the proportion of rings with sizes between 3 and 8 (inclusive) and those larger than 8, relative to the total number of rings in the molecule. Ring data was extracted using RDKit’s \texttt{Chem.rdchem.GetRingInfo().AtomRings()} method. Molecules without any rings are excluded from the average calculations.
    \item The \textbf{atom type distribution} is calculated as the ratio between the number of atom of each type and the output of RDKit's \texttt{Chem.rdMolDescriptors.CalcNumAtoms()} method.
\end{enumerate}

\begin{figure}
    \centering
    \includegraphics[width=1.0\linewidth]{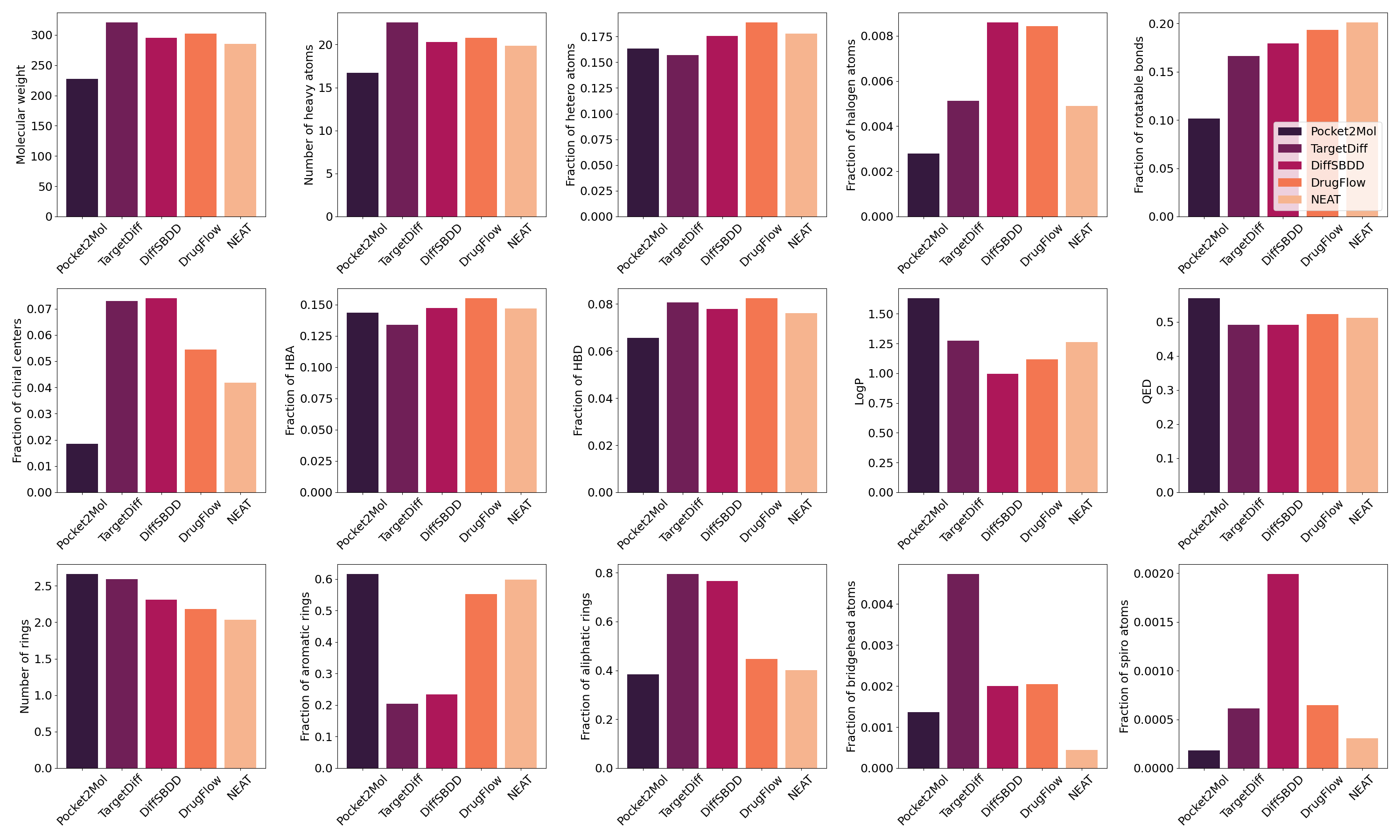}
    \caption{Average values of general physicochemical properties of molecules generated by Pocket2Mol, TargetDiff, DiffSBDD, DrugFlow and NEAT-POCKET trained on CrossDocked.}
    \label{fig:physchem_absolute}
\end{figure}

\begin{figure}
    \centering
    \includegraphics[width=1.0\linewidth]{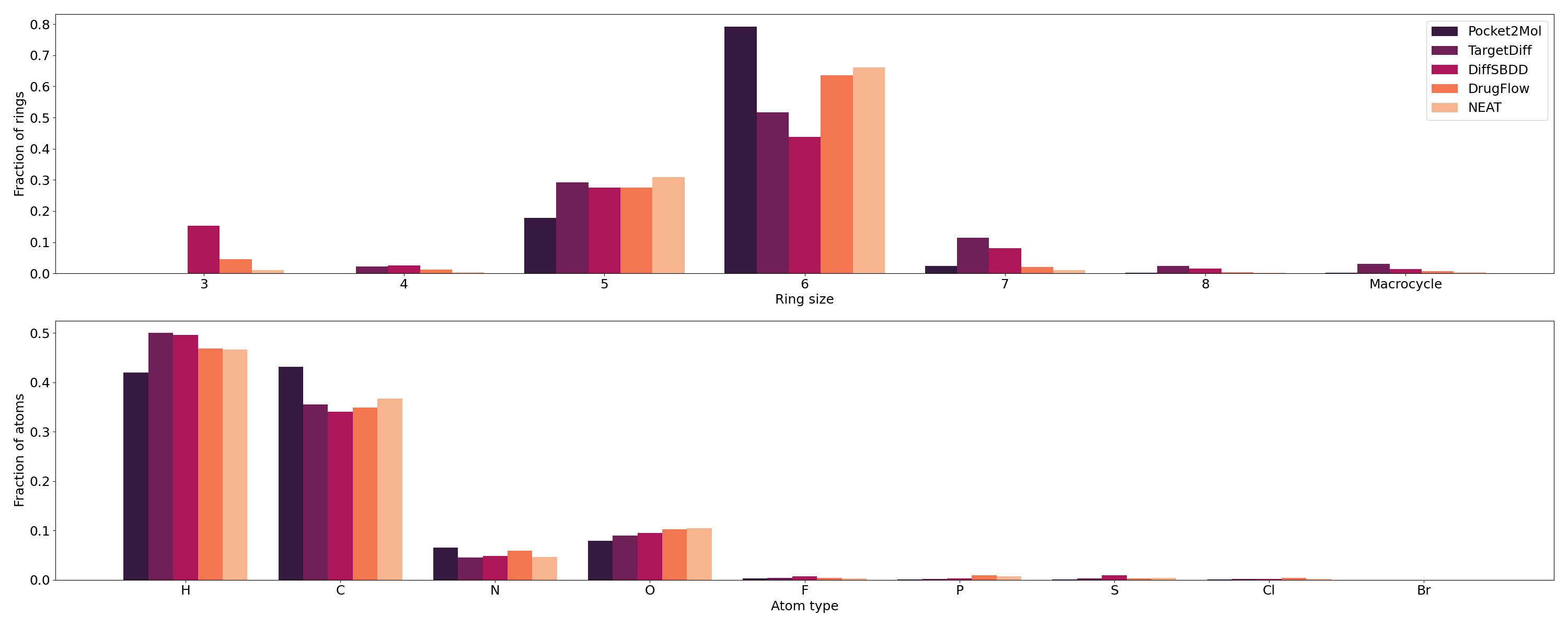}
    \caption{Distribution of ring sizes (top) and atom types (bottom) of molecules generated by Pocket2Mol, TargetDiff, DiffSBDD, DrugFlow and NEAT-POCKET trained on CrossDocked.}
    \label{fig:rings_and_atom_types_absolute}
\end{figure}

Comparative distributions of physicochemical descriptors relative to the CrossDocked dataset are shown in the main text (Figures~\ref{fig:physchem_relative} and \ref{fig:rings_and_atom_types_relative}), while absolute values are provided below (Figures~\ref{fig:physchem_absolute} and \ref{fig:rings_and_atom_types_absolute}).

The \textbf{\ac{sa} score} was calculated using RDKit's \texttt{SA\_Score module}, which estimates the ease of synthesizing a molecule based on its structural complexity. The score aggregates three terms: (1) an empirical fragment score based on the frequency of ECFP4 fingerprint bits, (2) a complexity penalty aggregating the number of atoms, chiral centers, spiro atoms, bridgehead atoms, and macrocyles, and (3) a symmetry correction term to rank highly symmetrical molecules as easier to synthesize. All contributions are summed, then linearly rescaled to 1–10 (with lower values being easier to synthesize).
    
Adherence to \textbf{Lipinski's \ac{ro5}} is reported as the percentage of generated molecules meeting the following criteria: molecular weight < 500 Da, logP < 5, number of HBD < 5, and number of HBA < 10.

\section{Additional results}\label{si:results}

In Table~\ref{tab:results_metrics_2} we report additional metrics used to compare the quality of molecules generated by NEAT-POCKET with those generated by other protein-pocket-conditioned 3D molecular generators, namely Pocket2Mol,\cite{Peng2022Pocket2Mol} TargetDiff,\cite{Guan2023TargetDiff} DiffSBDD,\cite{Schneuing2024DiffSBDD} DrugFlow,\cite{Schneuing2025DrugFlow} and FLOWR.\cite{Cremer2026Flowr} These metrics include the number of favorable protein-ligand interactions computed with the PoseCheck package,\cite{Harris2023PoseCheck} specifically \ac{hba}, \ac{hbd}, hydrophobic interactions, and \ac{vdw} contacts. We also report the \ac{sa} score, which estimates the ease of synthesizing a molecule based on its structural complexity, and the percentage of molecules that adhere to Lipinski's \ac{ro5}.

\begin{table}[t!]
    \caption{Performance of NEAT-POCKET vs. other 3D molecule generators trained on CrossDocked and SPINDR -- extended results}
    \centering
        \begin{tabular}{lcccccc}
                        & \makecell[c]{PC HBA}
                        & \makecell[c]{PC HBD}
                        & \makecell[c]{PC hydrophobic}
                        & \makecell[c]{PC vdW}
                        & \makecell[c]{SA score}
                        & \makecell[c]{Lipinski RO5} \\
                        & \makecell[c]{[1] $\uparrow$}
                        & \makecell[c]{[1] $\uparrow$}
                        & \makecell[c]{[1] $\uparrow$}
                        & \makecell[c]{[1] $\uparrow$}
                        & \makecell[c]{[1] $\downarrow$}
                        & \makecell[c]{[\%] $\uparrow$} \\
            \midrule
            \midrule
            CrossDocked & 2.13 $\pm$ 0.41
                        & 0.96 $\pm$ 0.21
                        & 1.49 $\pm$ 0.41
                        & 8.55 $\pm$ 0.71
                        & 7.50 $\pm$ 0.21
                        & 62.0 $\pm$ 9.7 \\
            \midrule
            Pocket2Mol  & 0.94 $\pm$ 0.16
                        & 0.35 $\pm$ 0.05
                        & \underline{1.63 $\pm$ 0.28}
                        & 6.60 $\pm$ 0.54
                        & \textbf{7.12 $\pm$ 0.07}
                        & \textbf{89.3 $\pm$ 4.2} \\
            TargetDiff  & \textbf{1.68 $\pm$ 0.27}
                        & \textbf{0.72 $\pm$ 0.09}
                        & 1.51 $\pm$ 0.24
                        & \textbf{9.44 $\pm$ 0.64}
                        & 8.04 $\pm$ 0.10
                        & 66.3 $\pm$ 4.7 \\
            DiffSBDD    & 1.37 $\pm$ 0.19
                        & 0.55 $\pm$ 0.06
                        & 1.59 $\pm$ 0.23
                        & \underline{8.53 $\pm$ 0.57}
                        & 7.84 $\pm$ 0.08
                        & \underline{77.8 $\pm$ 3.0} \\
            DrugFlow    & \underline{1.61 $\pm$ 0.29}
                        & \underline{0.69 $\pm$ 0.10}
                        & \textbf{1.60 $\pm$ 0.32}
                        & 8.31 $\pm$ 0.61
                        & 7.54 $\pm$ 0.09
                        & 69.0 $\pm$ 6.5 \\
            NEAT-POCKET     & 0.98 $\pm$ 0.18
                        & 0.33 $\pm$ 0.05
                        & \textbf{1.60 $\pm$ 0.30}
                        & 7.20 $\pm$ 0.52
                        & \underline{7.48 $\pm$ 0.09}
                        & 76.4 $\pm$ 5.6 \\
            \midrule
            \midrule
            SPINDR      & 2.39 $\pm$ 0.30 
                        & 1.27 $\pm$ 0.15
                        & 2.20 $\pm$ 0.26
                        & 8.52 $\pm$ 0.47
                        & 7.59 $\pm$ 0.12
                        & 63.2 $\pm$ 6.4 \\
            \midrule
            FLOWR       & \textbf{1.47 $\pm$ 0.19}
                        & \textbf{0.87 $\pm$ 0.09}
                        & \textbf{2.27 $\pm$ 0.19}
                        & \textbf{9.61 $\pm$ 0.26}
                        & 7.77 $\pm$ 0.04
                        & 62.9 $\pm$ 2.7 \\
            NEAT-POCKET     & 0.95 $\pm$ 0.12
                        & 0.40 $\pm$ 0.04
                        & 1.86 $\pm$ 0.17
                        & 7.02 $\pm$ 0.28
                        & \textbf{7.46 $\pm$ 0.06}
                        & \textbf{78.3 $\pm$ 3.4} \\
        \end{tabular}
    \label{tab:results_metrics_2}
\end{table}

\section{Strain energies}\label{si:strain_energies}

Strain energies were computed using PoseCheck.\cite{Harris2023PoseCheck} For each protein pocket, we first aggregated the strain energies of the 100 generated ligands by taking their mean, yielding one average strain-energy value per pocket. These per-pocket values were then aggregated across all pockets in the test set. Our original intention was to use the mean for this second aggregation step as well, in line with the other metrics reported in this work. However, we observed that the distribution of per-pocket strain energies can be highly skewed for some models, with a small number of pockets producing extremely large values. Therefore, in the main text, we report the median strain energy, which is more robust to such extreme cases. In this section, we further examine the effect of different aggregation strategies to provide a more complete picture of the strain-energy distributions.

Table~\ref{tab:strain_energies} reports, for each model, mean and median strain energies across pockets, the largest per-pocket average strain energy, and the mean and median after removing this largest value. In addition, we report the number of detected outlier pockets, distinguishing between upper-tail and lower-tail outliers, as well as the mean and median strain energies after removing all detected outliers.

Outliers were identified using the modified $z$-score computed on $\log_{10}$-transformed strain energies. Specifically, for each pocket $i$, the score is defined as

\begin{equation}
    z_i = 0.6745 \cdot \frac{\log_{10}(e_i) - \mathtt{median}(\log_{10}(e))}{\mathtt{MAD}},
\end{equation}

where $e_i$ denotes the average strain energy for pocket $i$, and $\mathtt{MAD}$ is the median absolute deviation of the log-transformed strain energies. A pocket is classified as an outlier if $|z_i| > 3.5$.

\begin{figure}
    \centering
    \includegraphics[width=1.0\linewidth]{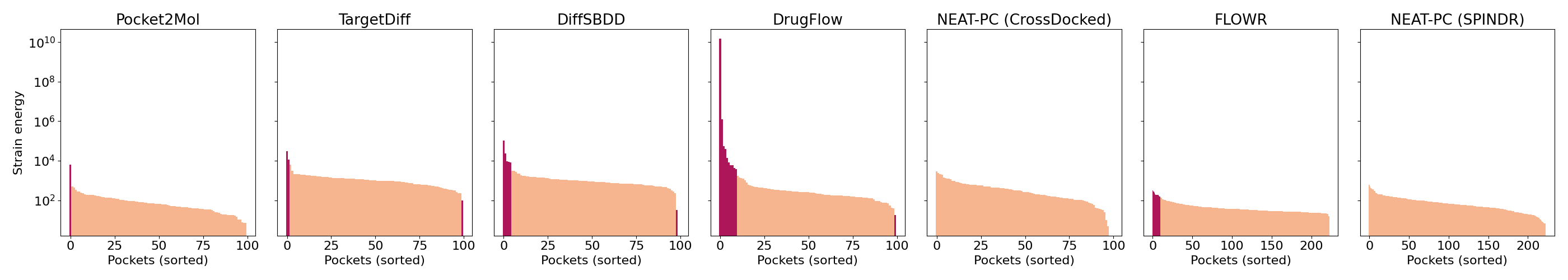}
    \caption{Average strain energies of generated molecules per CrossDocked and SPINDR test pockets.}
    \label{fig:strain_energies}
\end{figure}

The results in Table~\ref{tab:strain_energies} and Figure~\ref{fig:strain_energies} show that, for some models, the mean strain energy is strongly affected by a small number of extreme pockets. This effect is most pronounced for DrugFlow, for which the mean strain energy is $1.5 \times 10^8$, whereas the median is only 255. The largest per-pocket average strain energy for DrugFlow reaches $1.5 \times 10^{10}$, indicating that the mean is dominated by an extreme upper-tail value. Removing only this top-1 value reduces the mean to 14,422. A similar, although less extreme, pattern is observed for DiffSBDD: its mean strain energy is 2,485, but after removing the top-1 value, it decreases to 1,430. TargetDiff also exhibits several high-strain pockets, leading to a noticeable gap between its mean and median values. In contrast, Pocket2Mol, FLOWR, and NEAT-POCKET show substantially more stable strain-energy distributions. Their mean and median values are comparatively close, and removing either the largest value or the detected outliers has only a modest effect.

Overall, the analysis of strain energies obtained from models trained on CrossDocked highlights a qualitative difference between the evaluated model classes: the diffusion- and flow-based models considered here (TargetDiff, DiffSBDD, and DrugFlow) are more prone to producing pockets with extremely large average strain energies, whereas the autoregressive approaches (Pocket2Mol and NEAT-POCKET) yield more stable strain-energy profiles across the test set.

Interestingly, we observe a different trend on the SPINDR dataset. The median strain energies of NEAT‑PC and FLOWR are of the same order of magnitude, 62 vs. 35~kcal/mol, whereas the corresponding means differ more strongly, 848 vs. 48~kcal/mol. However, no NEAT‑PC pockets are flagged as outliers by the modified $z$-score on $\log_{10}$-transformed strain energies, and removing the single largest pocket only modestly changes the mean from 848 to 825~kcal/mol while leaving the median unchanged. This suggests that the elevated mean does not arise from isolated extreme failures, but rather from a broader upper tail in the strain-energy distribution. This indicates that strain energies remain comparable to those in FLOWR for most pockets, while highlighting a subset of higher-strain cases as targets for future improvement.

\begin{table}[t!]
    \caption{Aggregated strain energies in kcal/mol across CrossDocked and SPINDR test pockets under different aggregation and outlier-removal strategies}
    \centering
    \resizebox{\linewidth}{!}{
        \begin{tabular}{lrrrrrrrrr}
                        & \makecell[c]{Mean\\~}
                        & \makecell[c]{Median\\~}
                        & \makecell[c]{Top-1\\~}
                        & \makecell[c]{Mean\\w/o top-1}
                        & \makecell[c]{Median\\w/o top-1}
                        & \makecell[c]{Num.outliers\\upper tail}
                        & \makecell[c]{Num.outliers\\lower tail}
                        & \makecell[c]{Mean\\w/o outliers}
                        & \makecell[c]{Median\\w/o outliers} \\
            \midrule
            \midrule
            \multicolumn{10}{l}{\textsc{CrossDocked}} \\
            \midrule
            Pocket2Mol  & 156
                        & 65
                        & 6,538
                        & 91
                        & 65
                        & 1
                        & 0
                        & 91
                        & 65 \\
            TargetDiff  & 1,516
                        & 1,003
                        & 30,257
                        & 1,225
                        & 989
                        & 2
                        & 1
                        & 1,129
                        & 989 \\
            DiffSBDD    & 2,485
                        & 891
                        & 106,920
                        & 1,430
                        & 882
                        & 5
                        & 2
                        & 1,004
                        & 869 \\
            DrugFlow    & $1.5 \times 10^8$
                        & 255
                        & $1.5 \times 10^{10}$
                        & 14,422
                        & 254
                        & 10
                        & 1
                        & 306
                        & 226 \\
            NEAT-POCKET     & 411
                        & 265
                        & 2960
                        & 385
                        & 264
                        & 0
                        & 2
                        & 445
                        & 279 \\
            \midrule
            \midrule
            \multicolumn{10}{l}{\textsc{SPINDR}} \\
            \midrule
            FLOWR       & 48
                        & 35
                        & 314
                        & 48
                        & 35
                        & 10
                        & 0
                        & 41
                        & 34 \\
            NEAT-POCKET     & 848
                        & 62
                        & 610
                        & 825
                        & 62
                        & 0
                        & 0
                        & 848
                        & 62 \\
        \end{tabular}
    }
    \label{tab:strain_energies}
\end{table}

\section{Fragment generation}\label{si:fragment_generation}

\begin{figure}
    \centering
    \includegraphics[width=1.0\linewidth]{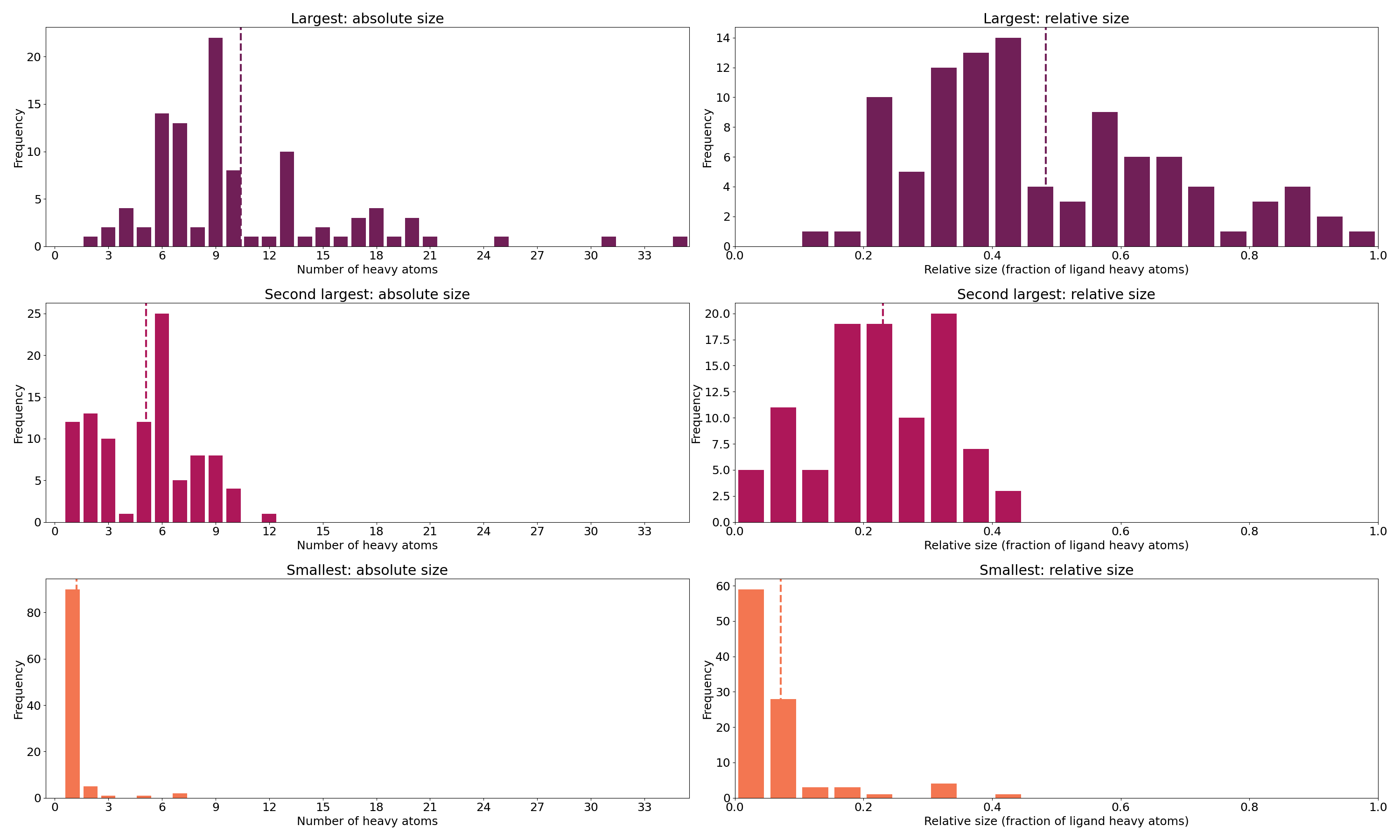}
    \caption{Distribution of absolute and relative fragment sizes obtained from the CrossDocked reference ligands.}
    \label{fig:fragment_stats_crossdocked}
\end{figure}

\begin{figure}
    \centering
    \includegraphics[width=1.0\linewidth]{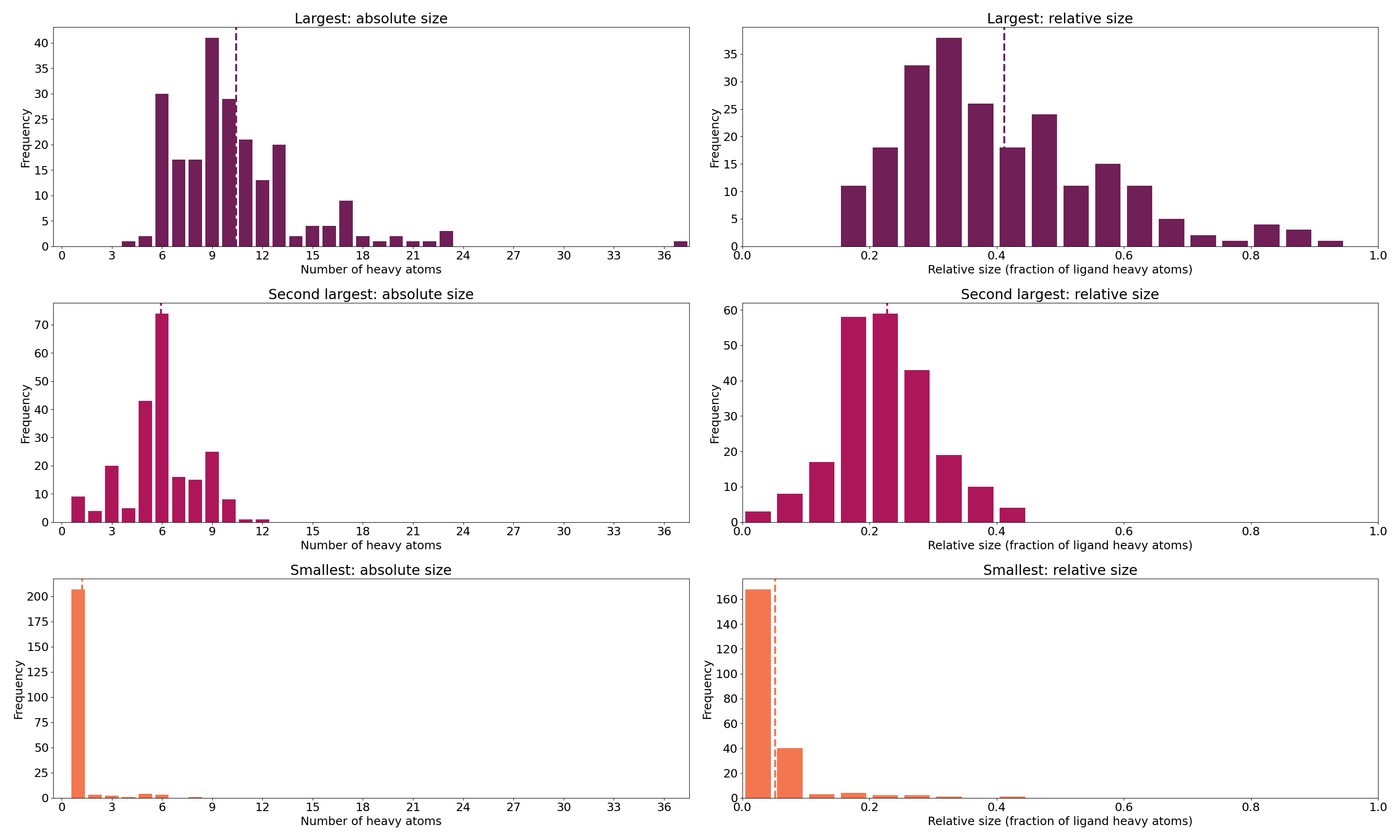}
    \caption{Distribution of absolute and relative fragment sizes obtained from the SPINDR reference ligands.}
    \label{fig:fragment_stats_spindr}
\end{figure}

To evaluate NEAT-POCKET's capabilities to generate complete molecules from fragments of different sizes, we generated a diverse set of fragments from the reference ligands in the CrossDocked and SPINDR test set. We decomposed each ligand using the BRICS fragmentation algorithm and used the resulting fragments as molecular prefixes. For each ligand, we considered three prefix sizes: the largest fragment, the second-largest fragment, and the smallest fragment. NEAT-POCKET was then tasked with completing each prefix in the corresponding protein pocket. On CrossDocked, the largest fragments contained on average 10.4 heavy atoms, corresponding to 48\% of the original ligand; the second-largest fragments contained 5.1 heavy atoms, corresponding to 23\%; and the smallest fragments contained 1.2 heavy atoms, corresponding to 7\%. The corresponding values on SPINDR were 10.4 heavy atoms, or 41\%, for the largest fragments; 6.0 heavy atoms, or 23\%, for the second-largest fragments; and 1.2 heavy atoms, or 5\%, for the smallest fragments. Figure~\ref{fig:fragment_stats_crossdocked} and \ref{fig:fragment_stats_spindr} show the distribution of absolute (\textit{i.e.}, number of heavy atoms) and relative (\textit{i.e.}, fraction of the reference ligand) fragment sizes sourced from CrossDocked and SPINDR, respectively.

\section{Examples of generated molecules}\label{si:examples}

\begin{figure}
    \centering
    \includegraphics[width=1.0\linewidth]{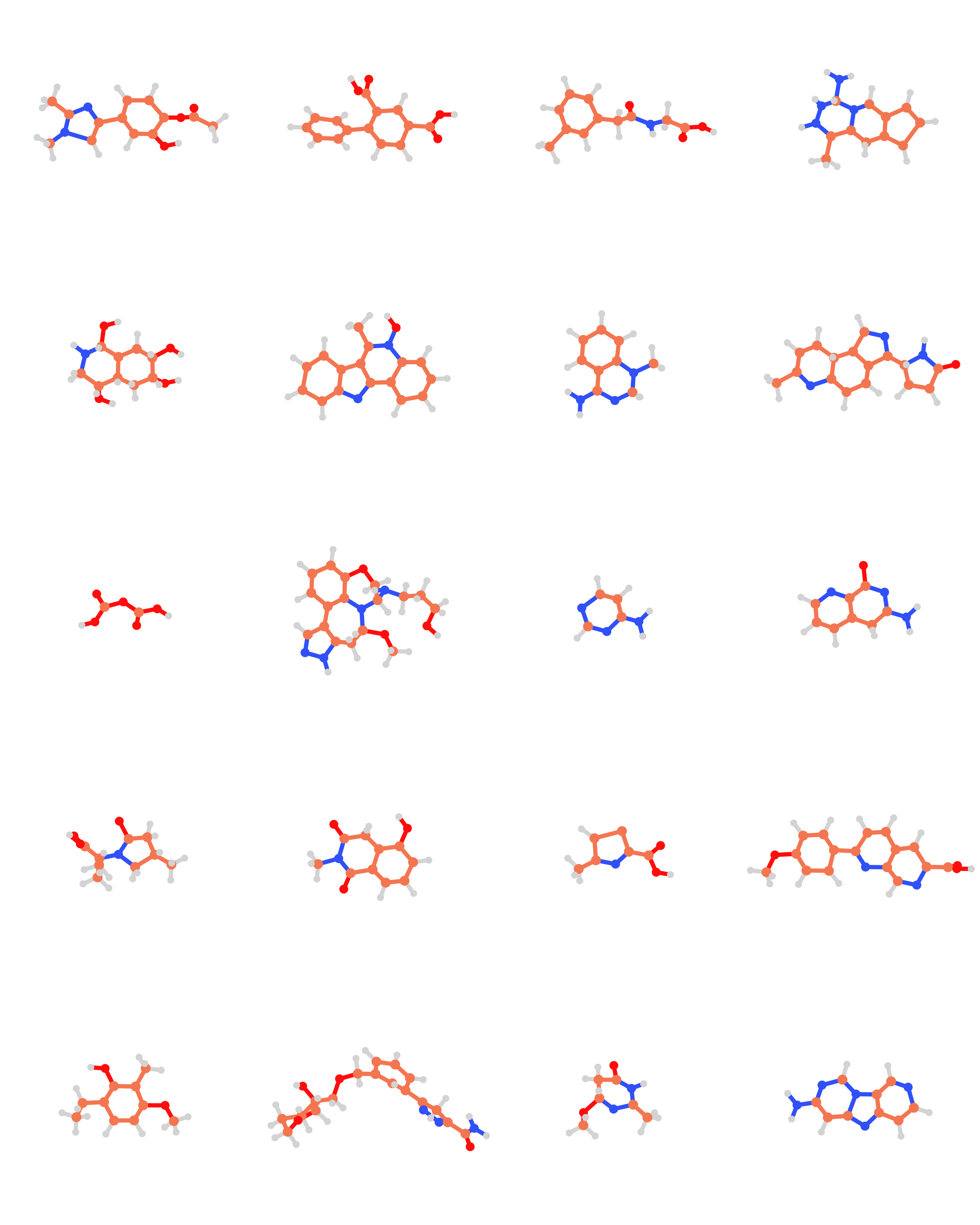}
    \caption{Examples of molecules generated by Pocket2Mol (3D).}
    \label{fig:pocket2mol}
\end{figure}

\begin{figure}
    \centering
    \includegraphics[width=1.0\linewidth]{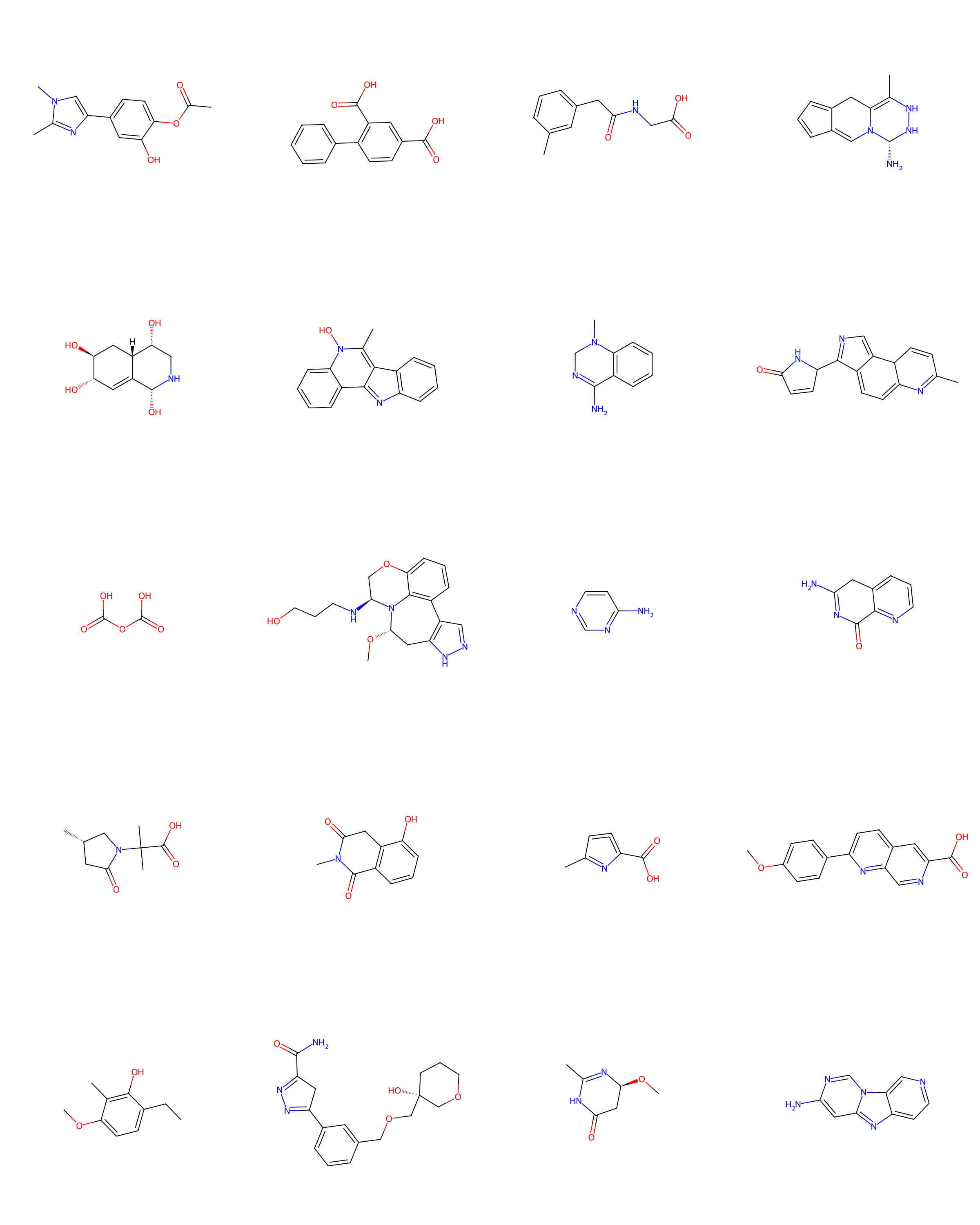}
    \caption{Examples of molecules generated by Pocket2Mol (2D).}
    \label{fig:pocket2mol_2D}
\end{figure}

\begin{figure}
    \centering
    \includegraphics[width=1.0\linewidth]{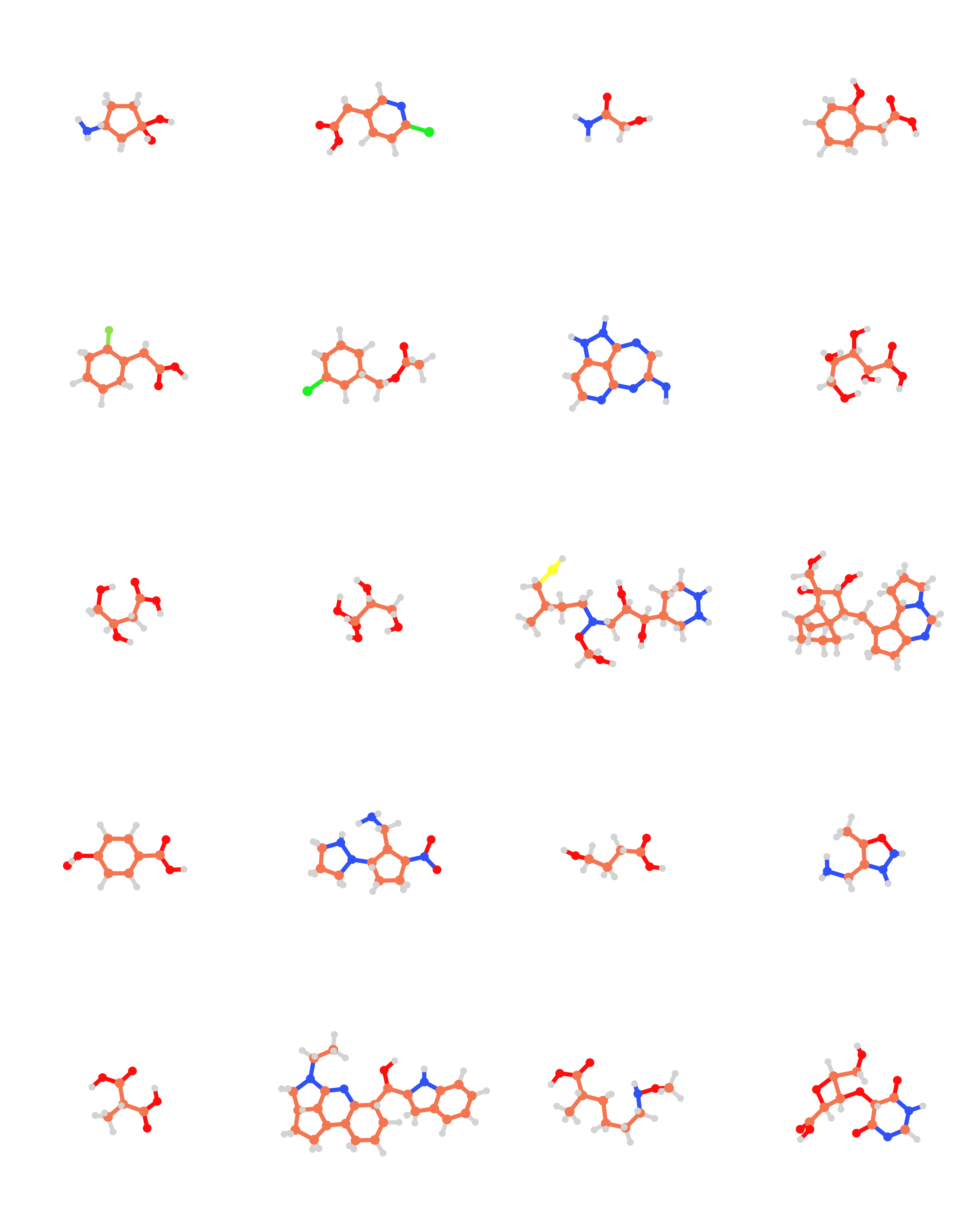}
    \caption{Examples of molecules generated by TargetDiff (3D).}
    \label{fig:targetdiff}
\end{figure}

\begin{figure}
    \centering
    \includegraphics[width=1.0\linewidth]{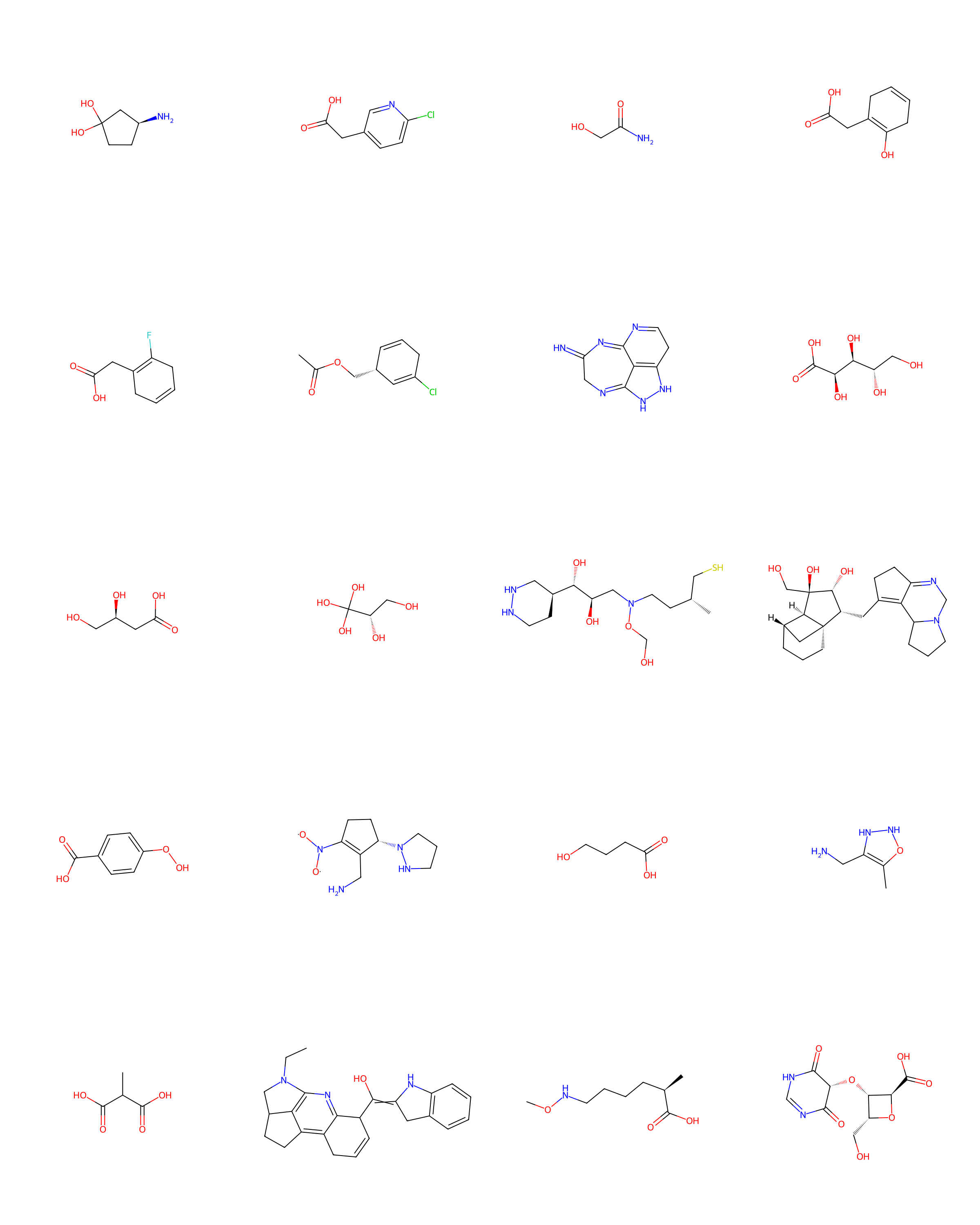}
    \caption{Examples of molecules generated by TargetDiff (2D).}
    \label{fig:targetdiff_2D}
\end{figure}

\begin{figure}
    \centering
    \includegraphics[width=1.0\linewidth]{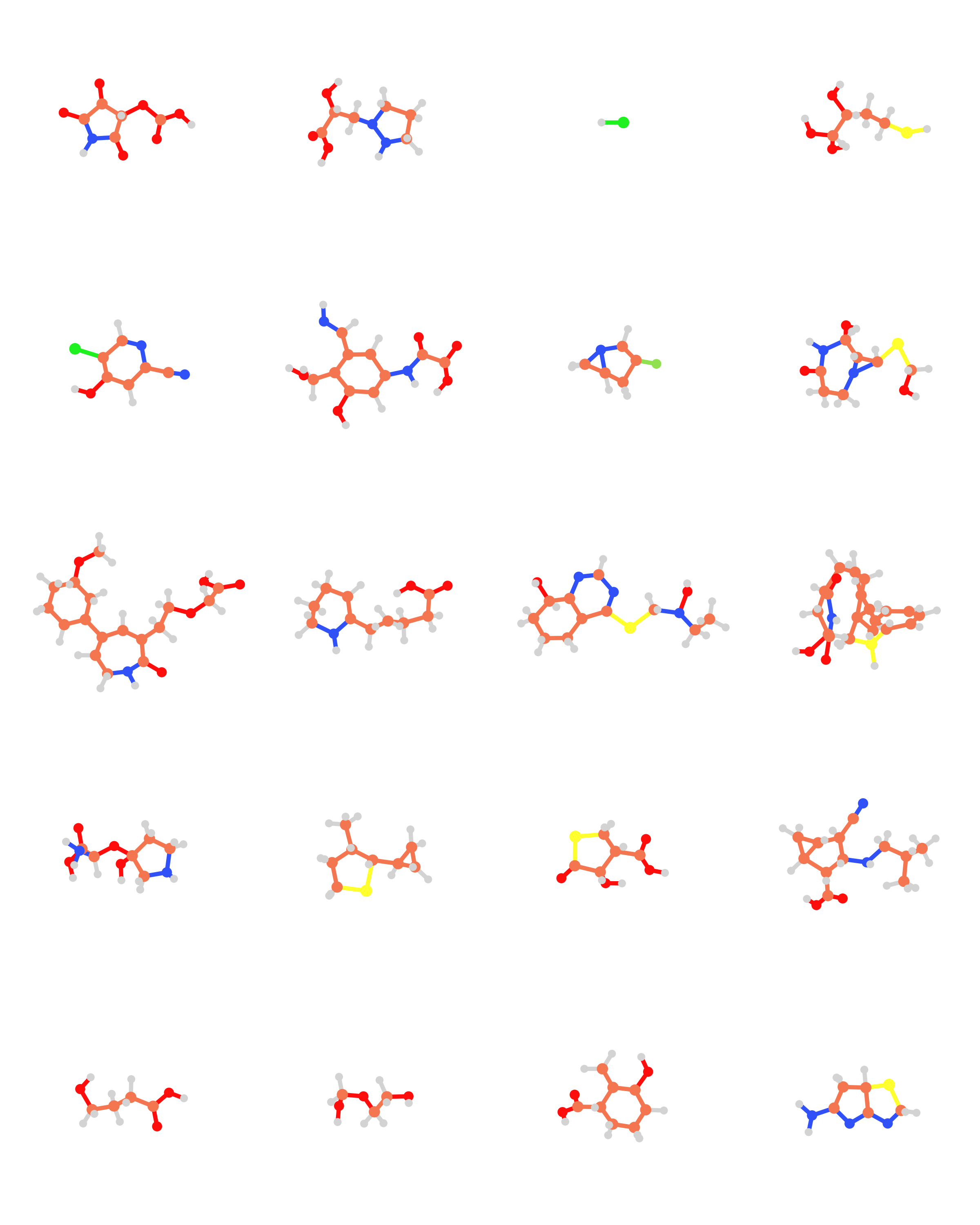}
    \caption{Examples of molecules generated by DiffSBDD (3D).}
    \label{fig:diffsbdd}
\end{figure}

\begin{figure}
    \centering
    \includegraphics[width=1.0\linewidth]{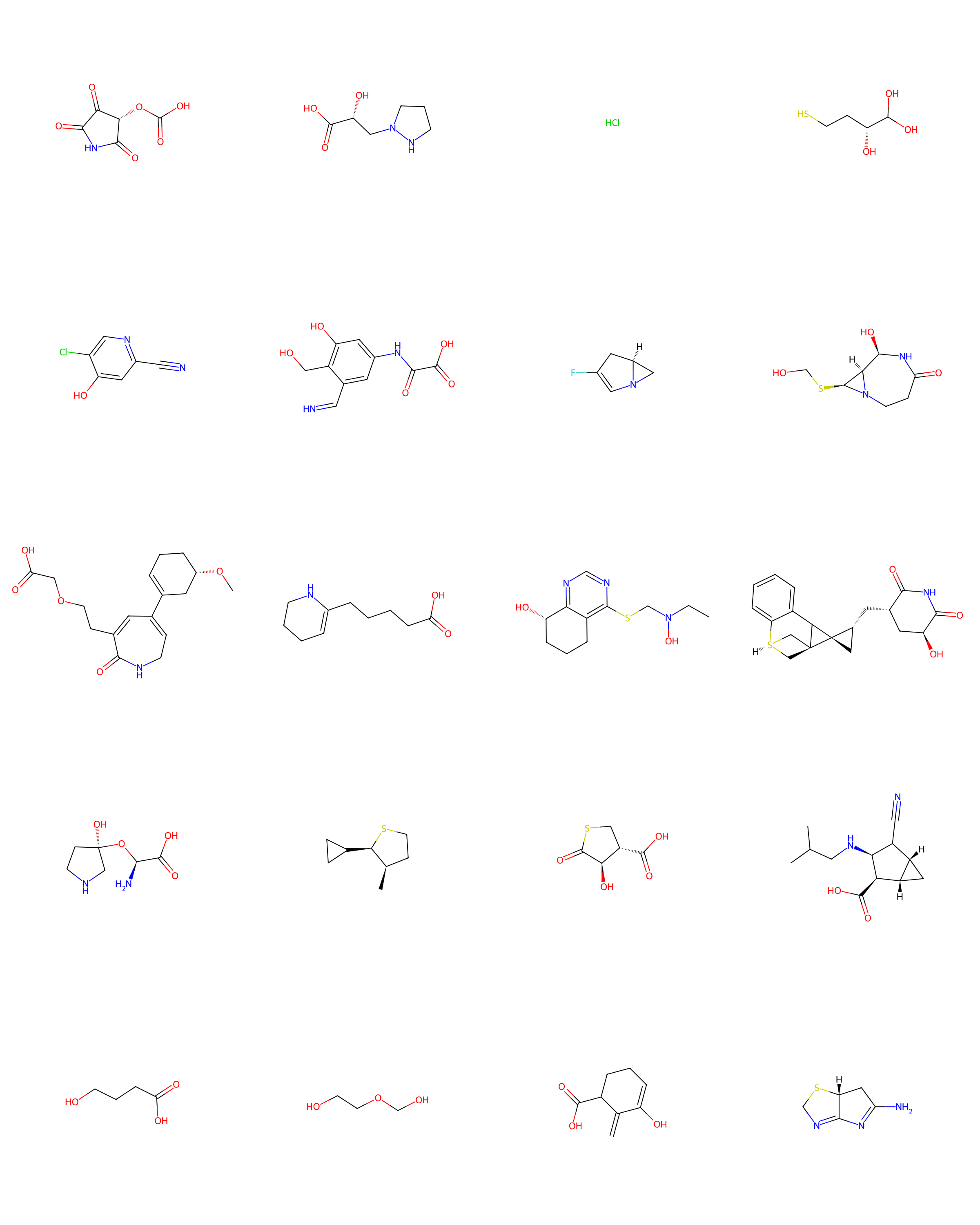}
    \caption{Examples of molecules generated by DiffSBDD (2D).}
    \label{fig:diffsbdd_2D}
\end{figure}

\begin{figure}
    \centering
    \includegraphics[width=1.0\linewidth]{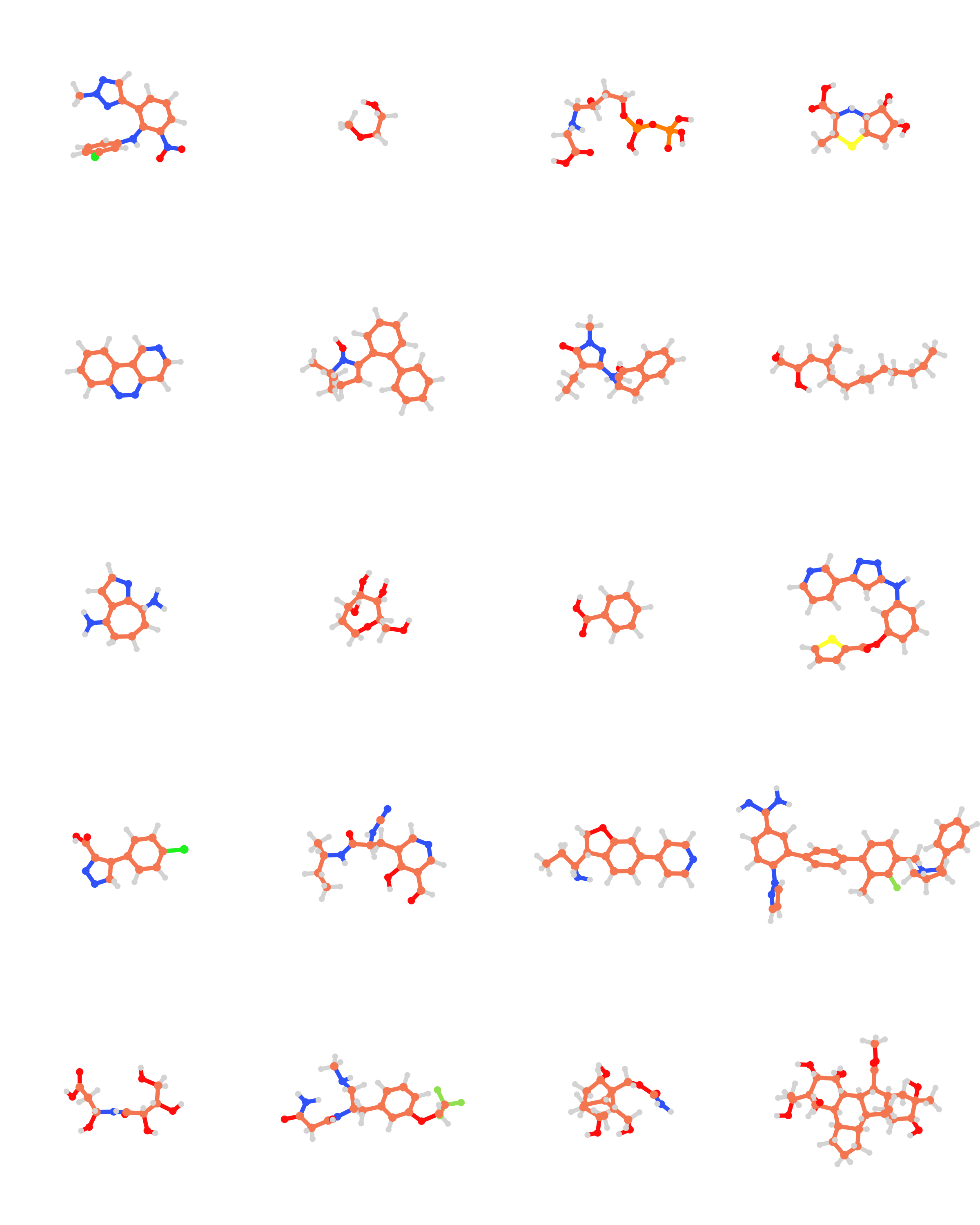}
    \caption{Examples of molecules generated by DrugFlow (3D).}
    \label{fig:drugflow}
\end{figure}

\begin{figure}
    \centering
    \includegraphics[width=1.0\linewidth]{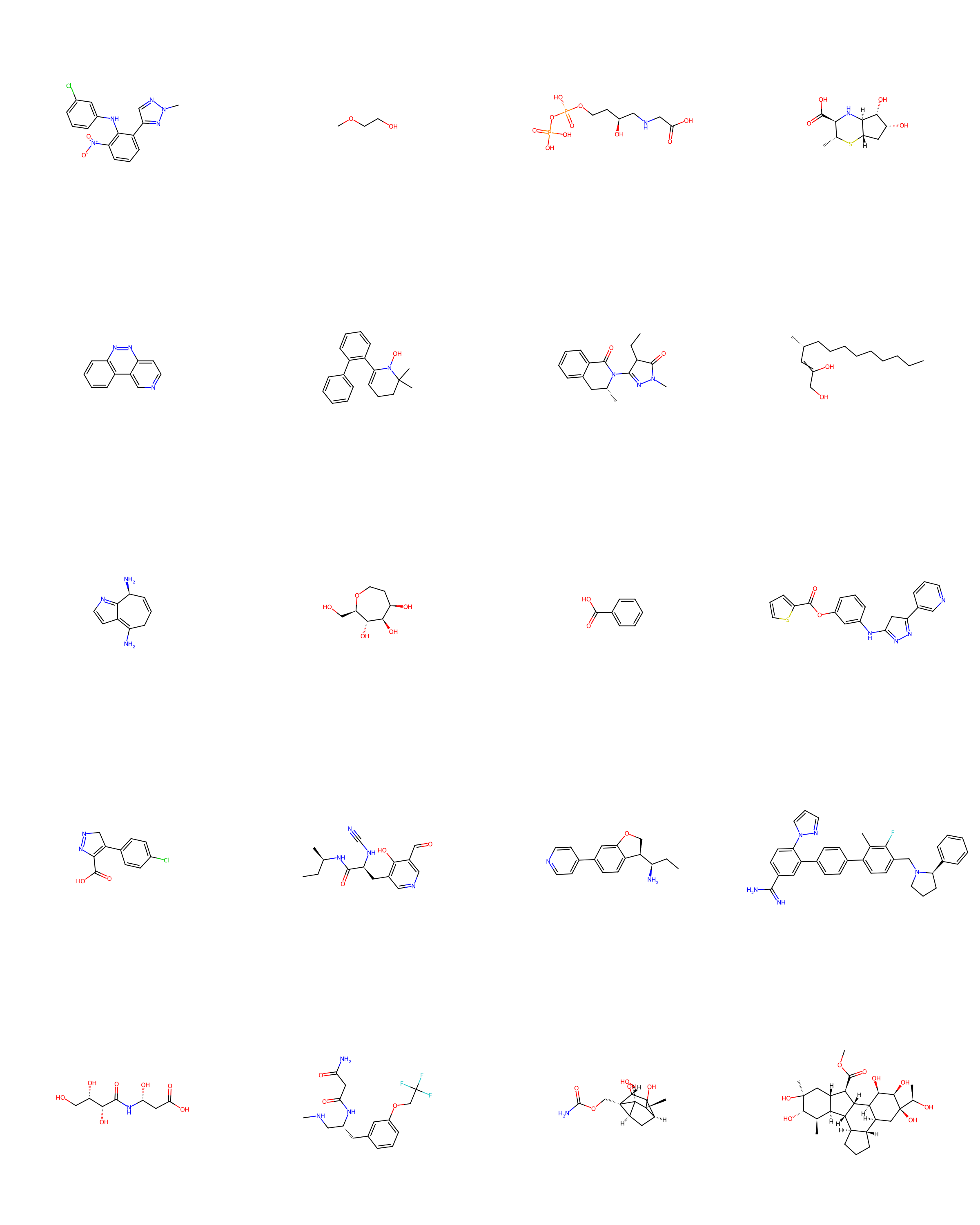}
    \caption{Examples of molecules generated by DrugFlow (2D).}
    \label{fig:drugflow_2D}
\end{figure}

\begin{figure}
    \centering
    \includegraphics[width=1.0\linewidth]{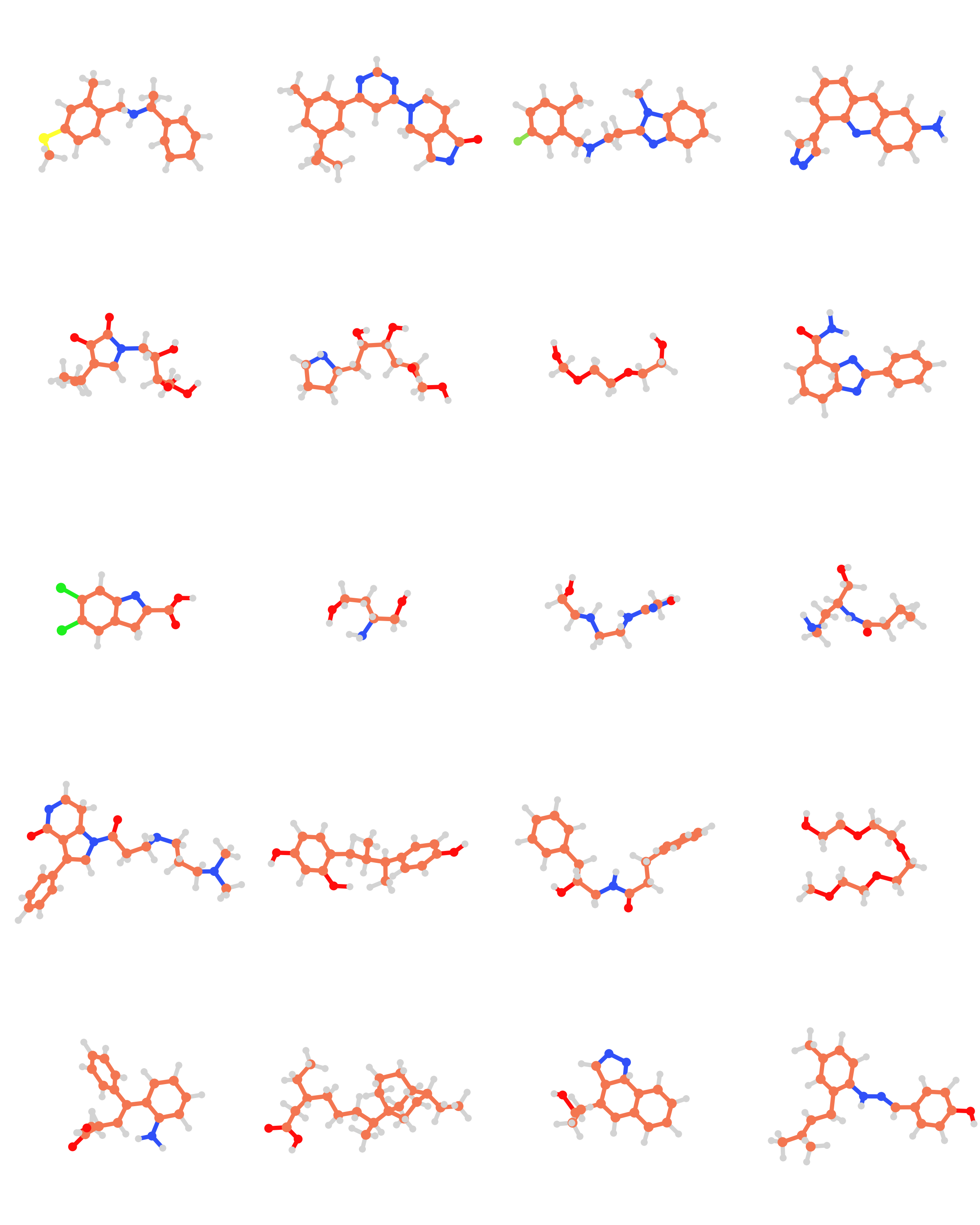}
    \caption{Examples of molecules generated by NEAT-POCKET trained on CrossDocked (3D).}
    \label{fig:neat_crossdocked}
\end{figure}

\begin{figure}
    \centering
    \includegraphics[width=1.0\linewidth]{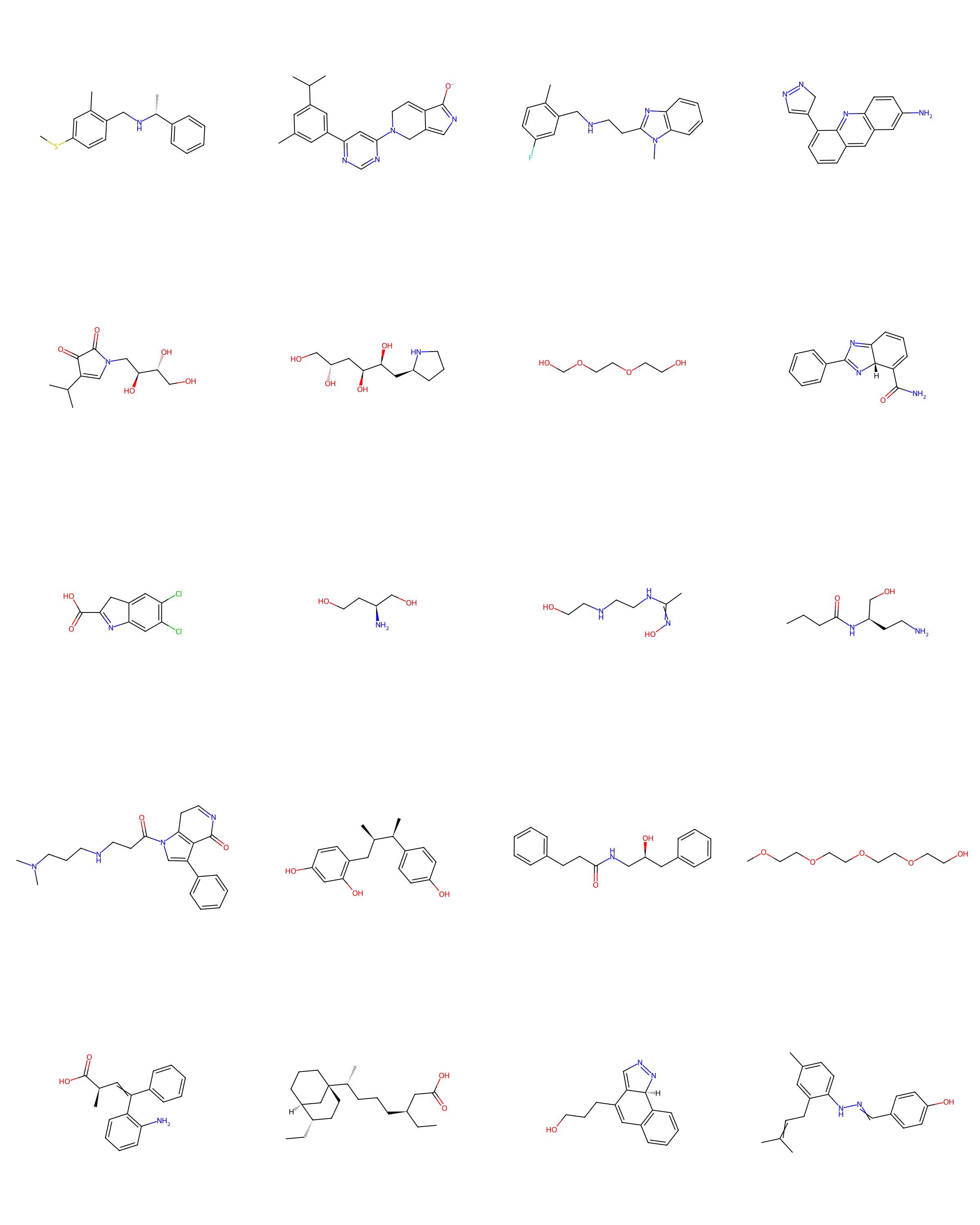}
    \caption{Examples of molecules generated by  NEAT-POCKET trained on CrossDocked (2D).}
    \label{fig:neat_crossdocked_2D}
\end{figure}

\begin{figure}
    \centering
    \includegraphics[width=1.0\linewidth]{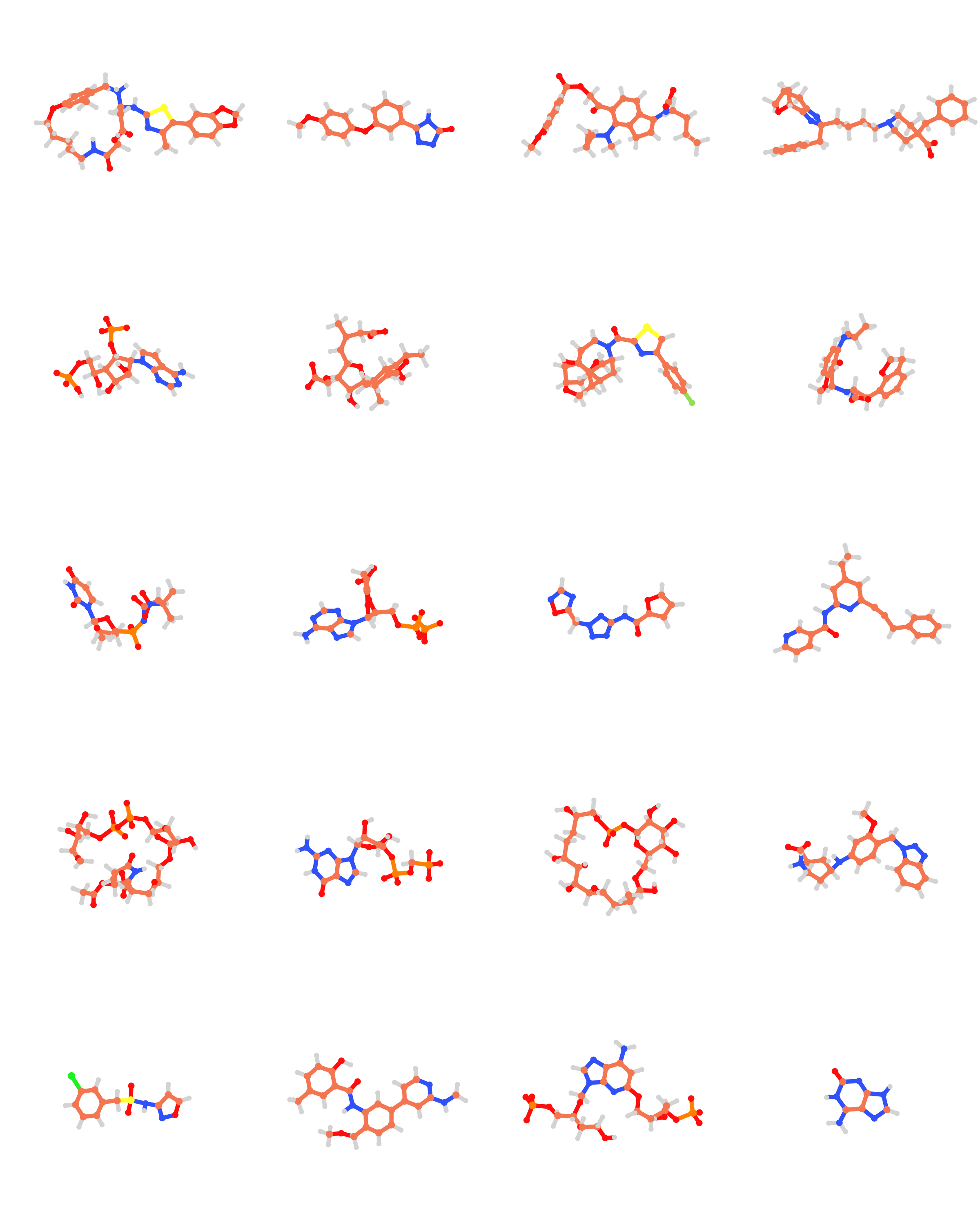}
    \caption{Examples of molecules generated by FLOWR (3D).}
    \label{fig:flowr}
\end{figure}

\begin{figure}
    \centering
    \includegraphics[width=1.0\linewidth]{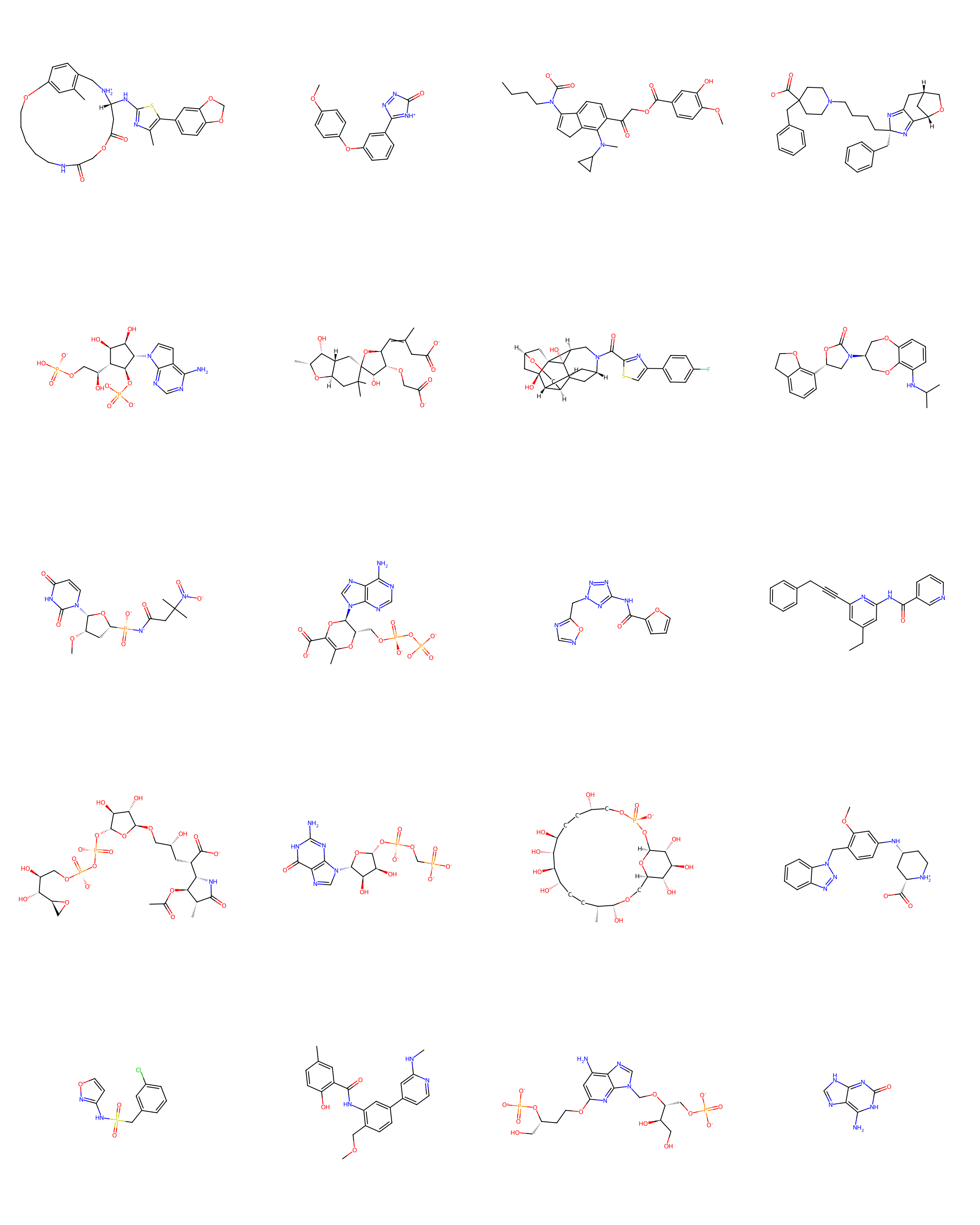}
    \caption{Examples of molecules generated by FLOWR (2D).}
    \label{fig:flowr_2D}
\end{figure}

\begin{figure}
    \centering
    \includegraphics[width=1.0\linewidth]{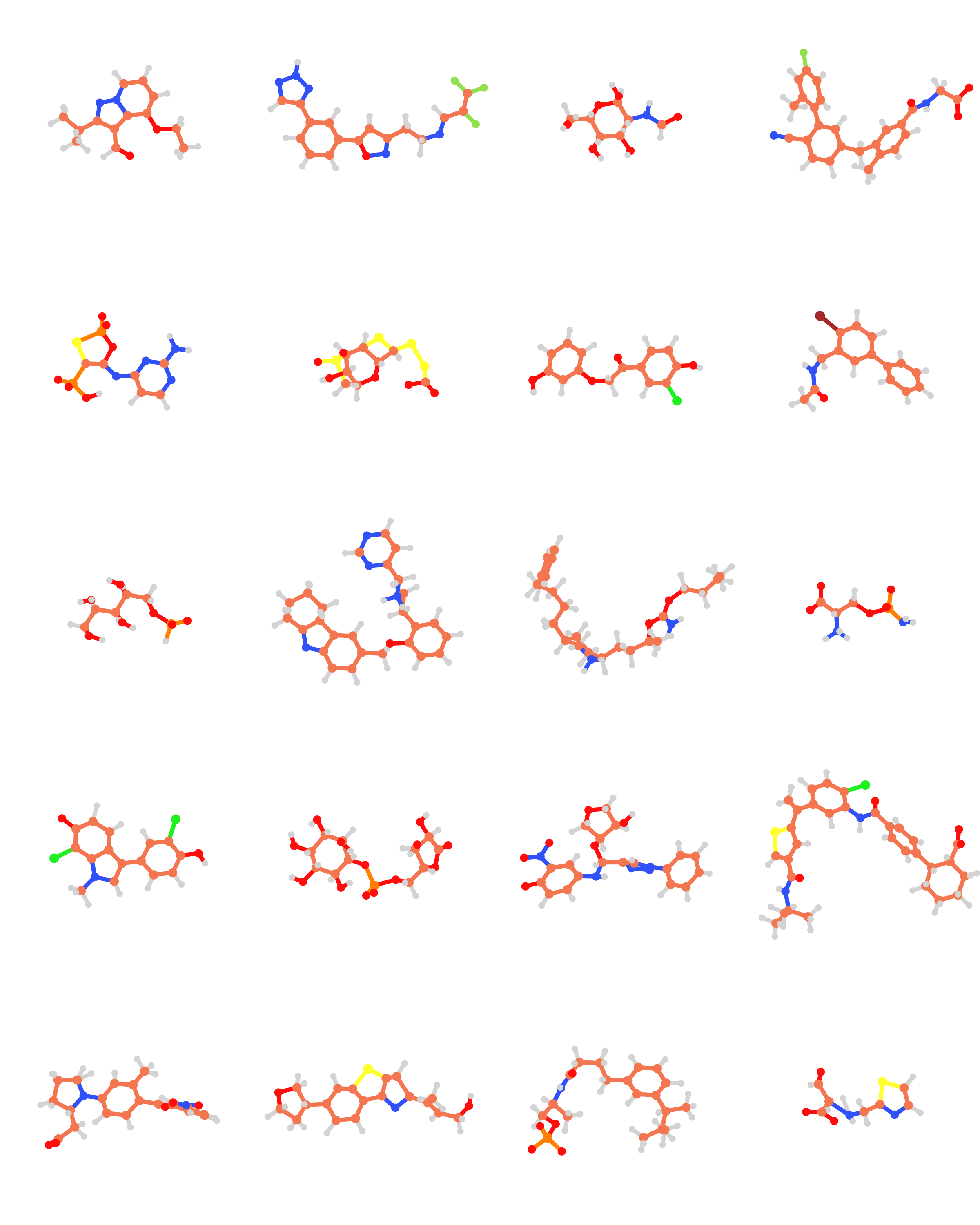}
    \caption{Examples of molecules generated by NEAT-POCKET trained on SPINDR (3D).}
    \label{fig:neat_spindr}
\end{figure}

\begin{figure}
    \centering
    \includegraphics[width=1.0\linewidth]{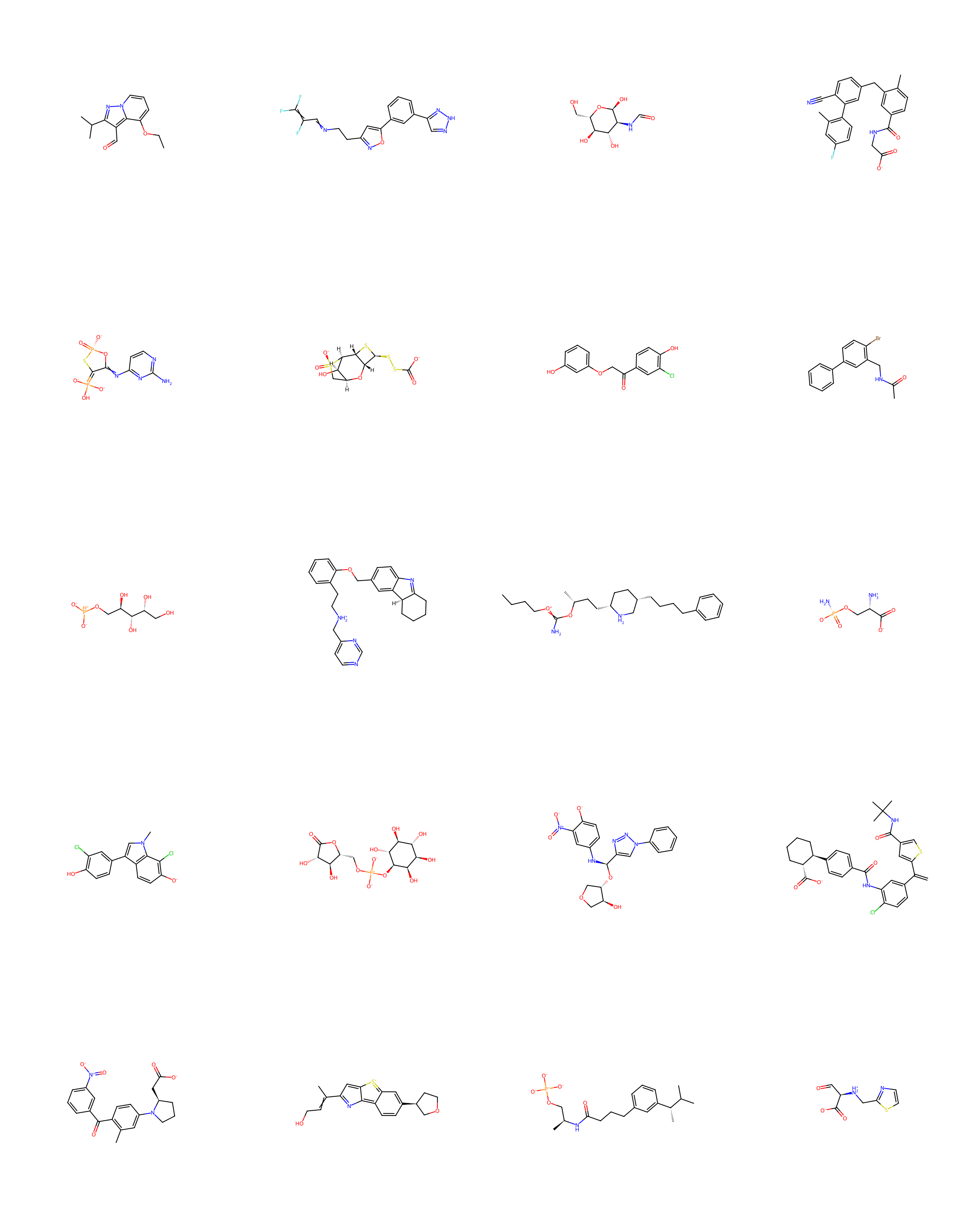}
    \caption{Examples of molecules generated by NEAT-POCKET trained on SPINDR (2D).}
    \label{fig:neat_spindr_2D}
\end{figure}

In this section, we show examples of molecules generated by each model evaluated in this study: Pocket2Mol (Figure~\ref{fig:pocket2mol} and \ref{fig:pocket2mol_2D}), TargetDiff (Figure~\ref{fig:targetdiff} and \ref{fig:targetdiff_2D}), DiffSBDD (Figure~\ref{fig:diffsbdd} and \ref{fig:diffsbdd_2D}), DrugFlow (Figure~\ref{fig:drugflow} and \ref{fig:drugflow_2D}), NEAT-POCKET trained on CrossDocked (Figure~\ref{fig:neat_crossdocked} and \ref{fig:neat_crossdocked_2D}), FLOWR (Figure~\ref{fig:flowr} and \ref{fig:flowr_2D}), and NEAT-POCKET trained on SPINDR (Figure~\ref{fig:neat_spindr} and \ref{fig:neat_spindr_2D}). For each model, we plotted 20 generated molecules in both 3D and 2D. Whenever possible, these molecules were randomly sampled from 20 distinct pockets. To improve readability, we restricted sampling to pockets with an average PB validity exceeding 70\%. For models with fewer than 20 pockets satisfying this criterion, namely DiffSBDD with 3 pockets and TargetDiff with 17, we sampled additional molecules from the qualifying pockets until 20 molecules were obtained.

\end{document}